\documentclass{article}
\usepackage[main, final]{neurips_2026}
\setcitestyle{numbers,square,comma}

\usepackage{amsmath,amsfonts,bm}

\def\eqref#1{equation~\ref{#1}}

\def\1{\bm{1}}

\DeclareMathAlphabet{\mathsfit}{\encodingdefault}{\sfdefault}{m}{sl}
\SetMathAlphabet{\mathsfit}{bold}{\encodingdefault}{\sfdefault}{bx}{n}

\usepackage{url}
\usepackage{graphicx}
\usepackage{microtype}
\usepackage{booktabs}
\usepackage{amsmath, amssymb}
\usepackage{xcolor}
\usepackage{algorithm}
\usepackage{algorithmic}
\usepackage{tcolorbox}
\tcbuselibrary{breakable}
\usepackage{enumitem}
\usepackage{titlesec}
\usepackage{tikz}
\usetikzlibrary{arrows.meta}
\usepackage{wrapfig}
\usepackage{authblk}
\usepackage{hyperref}

\setlist{topsep=2pt,itemsep=1pt,partopsep=0pt,parsep=0pt}

\title{The Alien Space of Science: Sampling Coherent but Cognitively Unavailable Research Directions}

\author[1,2,3]{Alejandro H.~Artiles}
\author[1]{Martin Weiss}
\author[2]{Levin Brinkmann}
\author[2]{Iyad Rahwan}
\author[3,4]{Bernhard Sch\"olkopf}
\author[5,6,7]{Christopher Pal}
\author[6,7,8]{Hugo Larochelle}
\author[ ]{Anirudh Goyal}
\author[1]{Nasim Rahaman}

\affil[1]{\small{Tiptree Systems}}
\affil[2]{\small{Max Planck Institute for Human Development, Berlin}}
\affil[3]{\small{Max Planck Institute for Intelligent Systems, T\"ubingen}}
\affil[4]{{\small ELLIS Institute T\"ubingen}\quad\textsuperscript{5}{\small Polytechnique de Montréal}}
\affil[6]{{\small Canada CIFAR AI Chair}\quad\textsuperscript{7}{\small Mila -- Quebec AI Institute}}
\affil[8]{{\small Université de Montréal}}

\begin{document}

\maketitle

\setlength{\parskip}{3pt plus 1pt minus 1pt}

\begin{abstract}
Scientific discovery is constrained not only by what is true, but by what is cognitively available to the researchers currently exploring a field. Many directions are coherent in light of the literature yet unlikely to be proposed because no existing community occupies the right combination of concepts, methods, and intuitions. Modern language models inherit this bias, recombining high-density regions of the literature when prompted for novel ideas. We introduce a framework that targets the complementary region, which we call the \textbf{alien space of science}, where directions are plausible under the structure of existing knowledge but unlikely under the distribution of existing researchers. Our method first decomposes papers into granular conceptual units and clusters them into a shared vocabulary of \emph{idea atoms}. It then learns two complementary models over this vocabulary. A \emph{coherence model} scores whether a combination of atoms forms a viable research direction, and an \emph{availability model} scores whether any existing author community is positioned to produce a given combination. Sampling alien directions then reduces to ranking atom combinations that maximize coherence while minimizing availability. On a corpus of 16{,}068 peer-reviewed LLM papers from NeurIPS, ICLR, ICML, and major NLP venues, the resulting sampler produces a more diverse portfolio of research ideas than frontier LLM ideation baselines, including batch-prompted variants that see their entire output, while matching or exceeding their per-idea quality under blind human, LLM, and downstream experimental evaluation. A stronger test is whether the learned scores can recover combinations that later materialize as real papers. With every component built exclusively from pre-2025 data, selecting less than 1\% of the candidate space recovers 2025 combinations at $35.9\times$ chance; weighting unavailability increases recovery of strictly novel combinations to $3.24\times$ chance. By separating scientific plausibility from community availability, our framework points toward AI ideation that complements rather than merely accelerates human science, expanding exploration into coherent directions that the current community may overlook.
\end{abstract}

\begin{center}
\small Code: \url{https://github.com/alejandrohdez00/alien-space-of-science}
\end{center}

\section{Introduction}

Scientific discovery is often described as a search through the space of possible ideas. But the space visible to a scientific community is only a small part of the space that may be scientifically coherent. Researchers inherit concepts, methods, collaborators, datasets, institutions, and disciplinary intuitions that make some directions easy to imagine and others effectively invisible. Two ideas may be equally plausible in light of the literature, yet differ dramatically in whether any existing researcher or community is likely to propose them.

We call the latter region the \emph{alien space of science}: directions that are coherent under the structure of existing knowledge, but that do not naturally arise from the conceptual trajectories of existing researchers within a community. In hindsight such ideas may look obvious; before they appear, they sit outside prevailing taste and require expertise beyond what the field has already organized. They are not alien because they are speculative or unscientific; they are alien because they are cognitively unavailable to the community currently organized around the literature.

This distinction matters increasingly as large language models (LLMs) are used for scientific ideation. LLMs excel at synthesizing existing knowledge, but synthesis is not the same as escaping a field's prior. When prompted for novel ideas, they tend to interpolate between fashionable concepts, salient mechanisms, and recurring research tropes already present in the literature. Rather than reliably helping researchers escape the community prior, they often approximate and amplify it \citep{zhang2025language, si2024can, schopf2026is, si2026theideaexecutiongap, jiang2025artificial}.

The central question of this paper is whether we can explicitly model and search beyond this prior. Rather than asking whether an idea is novel in an absolute sense, we ask whether it is \emph{cognitively available} to the current scientific community \citep{tversky1973availability, evans2023accelerating}. An idea is highly available if many researchers, given their past work, are naturally positioned to propose it; it is cognitively unavailable if it requires an unusual combination of concepts, methods, or intuitions not well represented by existing research trajectories.

This gives a two-axis view of scientific ideation. \emph{Coherence} asks whether the components of an idea fit together in a way that could plausibly support a research contribution. \emph{Availability} asks whether this combination is likely to be generated by the researchers and communities currently active in the field. Standard LLM ideation is often coherent but available. Random recombination can be unavailable but incoherent. The region we seek is the high-coherence, low-availability frontier: ideas that are simultaneously plausible and non-obvious.

\begin{figure}[t!]
\centering
\includegraphics[width=1\textwidth]{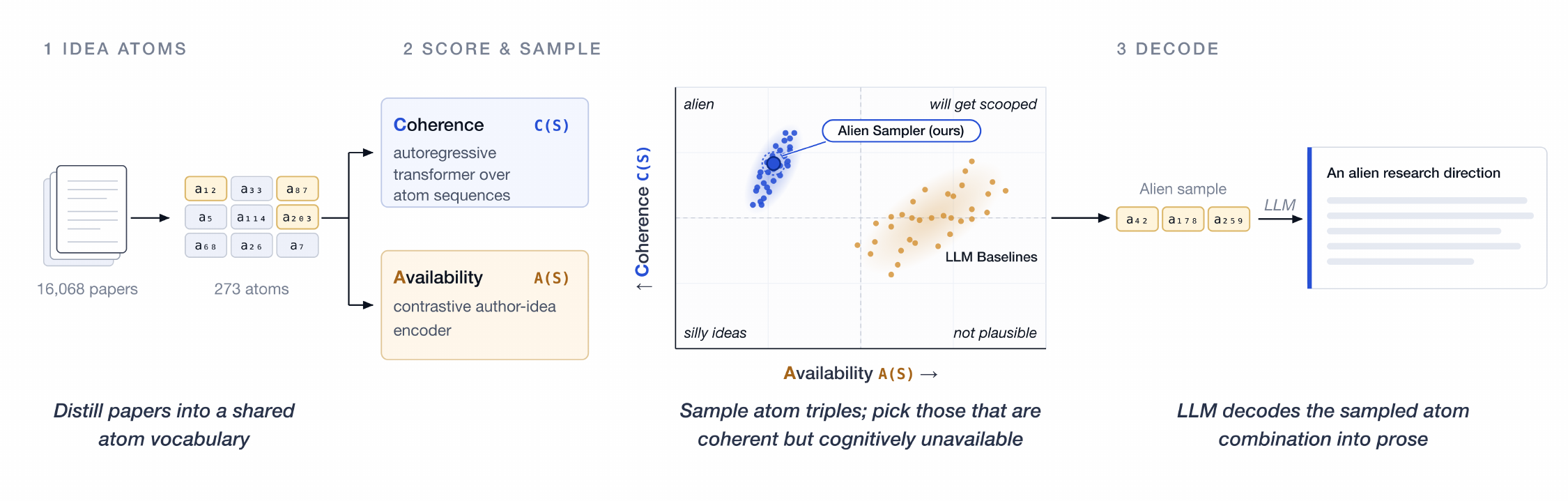}
\vspace{-14pt}
\caption{\looseness=-1 Overview of the Alien Science Sampling pipeline. Papers are distilled into conceptual units, which are clustered into a shared vocabulary of idea atoms. A coherence model learns which atom combinations form viable research directions, while an availability model estimates which combinations existing author communities are positioned to produce. Alien directions are sampled by maximizing coherence while minimizing availability.}
\label{fig:architecture}
\end{figure}

To operationalize this view, we proceed in three steps. \textbf{First}, we decompose papers into short \emph{conceptual units}\footnote{An example conceptual unit: ``\emph{Polysemanticity in neural networks occurs when individual neurons respond to multiple, unrelated features (like 'text' and 'dog faces') due to superposition, which makes internal representations difficult for humans to interpret directly.}'' Examples of \emph{idea atoms} are in Appendix~\ref{app:example_atoms}.} and cluster recurring units into a vocabulary of \emph{idea atoms} (Section~\ref{sec:units}); each paper is represented as a sparse combination of such atoms. \textbf{Second}, we learn two complementary models over this discrete idea space: a \emph{coherence model} (Section~\ref{sec:coherence}), trained on atom sets from papers, that scores whether a combination resembles a viable research direction; and an \emph{availability model} (Section~\ref{sec:availability}) that learns an author--idea compatibility function, scoring whether a community of existing researchers is positioned to produce it. \textbf{Third}, we sample combinations that maximize coherence while minimizing availability (Section~\ref{sec:sampling}). We call this process \emph{alien sampling}.

\textbf{Contributions.} Our contributions are three-fold. 

\begin{itemize}[beginpenalty=10000]
    \item \textbf{Idea atoms.} We introduce a compositional representation of scientific directions, in which papers are mapped to sparse combinations over a shared, LLM-distilled vocabulary of recombinable idea atoms.
    \item \textbf{Cognitive availability as a search objective.} We operationalize \emph{cognitive availability} \citep{tversky1973availability, evans2023accelerating} as a learnable, community-level author--idea compatibility score, and pair it with a coherence model to search the alien--science frontier.
    \item \textbf{Empirical evidence for search beyond the community prior.} On a curated corpus of 16{,}068 LLM papers, we show that (i) Alien sampling yields a more diverse idea portfolio than frontier LLM baselines, including batch-prompted variants, with independently evaluated ideas matching the strongest baseline's mean human rating and leading the coherent-originality and report comparisons, ranking first on a joint quality--diversity score; (ii) the availability model generalizes to researchers it has never seen, identifying unpublished combinations within their reach; and (iii) with vocabulary and models built only from pre-2025 data, the scores recover combinations first appearing in 2025 papers far above chance, with the availability term steering recovery toward work far from existing author communities.
\end{itemize}

\section{Related Work}

\paragraph{AI-assisted scientific discovery.}
Automated discovery has a long history in mathematics, chemistry, physics or scientific modeling \citep{lenat1977auto,buchanan1981dendral,langley1987scientific, krenn16,langley2024integrated}. Recent LLM-based systems such as AI-Scientist \citep{lu2024ai} and AlphaEvolve \citep{novikov2025alphaevolve} extend this agenda to broader scientific and engineering settings, using language models to propose, implement, evaluate, and refine candidate discoveries \citep{romera2024mathematical,hayes2025simulating}. Our focus is the ideation step inside this broader loop: how should a system choose which directions to investigate?

\paragraph{The ideation bottleneck.}
LLMs are increasingly competent at literature synthesis, coding, experimental design, and scientific writing \citep{xu2022systematic,lu2024discovering,wang2024autosurvey}. Yet open-ended generation exposes a distinctive failure mode: models repeatedly return to a narrow region of familiar concepts, templates, and motifs, both within a model and across models \citep{si2024can,doshi2024generative,jiang2025artificial,zhang2025noveltybench}, also flagged by \citet{lu2024ai}. We interpret this as a community-prior problem: plausibility is often achieved by staying close to the distribution of existing researchers.

\paragraph{Retrieval- and feedback-augmented ideation.}
Systems such as SCIMON \citep{wang2024scimon}, ResearchAgent \citep{baek2025researchagent}, and AI Co-Scientist \citep{gottweis2025towards} augment LLMs with literature retrieval, knowledge graphs, review loops, and refinement procedures. These systems improve grounding and execution, but they often begin from a human-supplied seed concept, core paper, or problem formulation. Since human seeds tend to come from cognitively available regions, the downstream system can inherit the same bias.

\paragraph{Combinatorial creativity and concept recombination.}
Treating new ideas as recombinations of existing ones has deep roots in the study of creativity, from Einstein's ``combinatory play'' \citep{hadamard1945psychology} to Koestler's bisociation of previously unconnected frames \citep{koestler1964act}, Boden's combinational creativity \citep{boden2004creative}, and conceptual blending \citep{fauconnier2002way}. A recent line treats the LLM as a decoder over externally supplied concept combinations \citep{zhao2025ramon,sternlicht2025chimera,radensky2024scideator, gu2026generatinginterestingscientificideas}. Along with the aforementioned papers, we share the view that novelty can be induced by placing the model in a conceptual state it would not naturally reach, but differ in two ways: prior systems operate over coarse keyword-level concepts (\emph{LLM}, \emph{RAG}, \emph{MCTS}) \citep{zhao2025ramon, gu2026generatinginterestingscientificideas}, whereas we build a content-bearing vocabulary bottom-up from methodological statements; and they sample by random choice, human choice, or distance heuristics, whereas we learn an objective that separates coherence from availability. Our atoms also play a role analogous to skills in compositional accounts of LLM capabilities \citep{arora2023theory,yu2023skillmix,didolkar2024metacognitive,kaur2025instructskillmix}, but are methodological units distilled from papers and are searched with learned objectives rather than mixed at random.

\paragraph{Cognitive availability and scientific surprise.}
Surprising combinations of research content can predict outsized scientific impact, especially when they bridge distant intellectual regions \citep{shi2023surprising}. Other work operationalizes surprise, novelty, or unexpectedness for active discovery \citep{agarwal2025autodiscovery,kargupta2026sparking,si2024can,schopf2026is, artiles2025cultural, zhang2025language}. We target a related but distinct construct: \emph{cognitive availability} \citep{tversky1973availability}. \citet{evans2023accelerating} model collective scientific attention through random walks on a researcher--concept hypergraph. We adapt the availability intuition to a generative setting, replacing graph distance with a learned dual-encoder model that scores whether author communities are positioned to produce a full research direction. Walk-based distances capture meaningful structure in domains with narrow author specialization, but ML researchers publish broadly across subareas, making nearly every atom pair reachable within two hops. As a result, direct hypergraph approaches largely collapse in this setting (Appendix~\ref{app:availability_hypergraph_comparison}).

\section{Method}

The method follows directly from the thesis. To search coherent directions outside the community prior, we need: a representation of research directions, a model of scientific coherence, a model of community availability, and a search objective that trades them off.

\subsection{Representing Research Directions as Idea Atoms}
\label{sec:units}

\paragraph{Corpus.}
We collect 16{,}068 papers on large language models from NeurIPS, ICLR, ICML, and major NLP and AI conferences (2017--2025; Appendix~\ref{app:data_collection}). The domain is topically dense and methodologically diverse, yet bounded enough to approximate broad coverage. This matters for availability modeling: when a direction is estimated as unavailable, the estimate should reflect a real absence from the author community rather than missing data. Availability is therefore measured relative to the community this corpus covers, which is also the community that LLM-ideation tools would serve; extending to the broader ML or AI literature requires substantially greater data coverage.

\paragraph{Conceptual units.}
Raw papers are too entangled for recombination. They include citations, repeated motivation, formatting, experimental detail, and result-specific context. We therefore first compress each paper into a methodological summary focused on mechanisms, objectives, architectures, training procedures, evaluation setups, and technical insights. We then prompt an LLM to extract \emph{conceptual units}: short, self-contained statements describing a technique, mechanism, objective, architectural choice, or evaluation procedure. Each unit states both what is done and why it matters, and must be understandable without the original paper. This focus on methodology is deliberate: methods contain the recombinable conceptual building blocks we wish to sample over, whereas experimental results are environment-dependent outcomes that cannot be meaningfully recombined.

\paragraph{Idea atoms.}
Many conceptual units express the same reusable idea in different language. We embed all units and cluster them with HDBSCAN \citep{campello2013density,mcinnes2017hdbscan}; each cluster is summarized into a canonical description using an LLM. The result is an \emph{idea atom}: a recurring conceptual building block shared across papers. The final representation gives a vocabulary of 273 atoms and a sparse mapping from each paper to the atoms it expresses. Idea atoms sit between keywords and full papers: keywords are too coarse to encode mechanism; full text is too specific to recombine.

\subsection{Modeling Scientific Coherence}
\label{sec:coherence}

Coherence asks whether a set of atoms may plausibly co-occur in one paper. We model coherence with an autoregressive Transformer \citep{DBLP:journals/corr/VaswaniSPUJGKP17}, treating each atom as a discrete token. At train time, each paper's atoms are presented as multiple random serializations, so the model captures paper-level co-occurrence structure without committing to a canonical ordering.\footnote{At inference time the model is robust to ordering: it distinguishes real papers, random combinations, and disjoint examples regardless of order (Figure~\ref{fig:coherence-order-sensitivity}).} For an atom set $S=\{a_1,\ldots,a_n\}$ serialized as $(a_1,\ldots,a_n)$, the model optimizes next-atom prediction $p_\theta(a_t\mid a_{<t})$. We score coherence by length-normalized log likelihood,
$C(S)=\frac{1}{|S|}\sum_{t=1}^{|S|}\log p_\theta(a_t\mid a_{<t})$.
High $C(S)$ indicates that the atoms in $S$ plausibly co-occur within a single paper.

\subsection{Modeling Cognitive Availability}
\label{sec:availability}

\paragraph{What availability measures.}
Cognitive availability originates in the availability heuristic \citep{tversky1973availability}: people judge how likely something is by how easily instances come to mind. \citet{evans2023accelerating} lift it to the population level: a community explores the combinations of ideas that are reachable within its researchers' experience, and proxies of this reachability predict which directions a field takes next. We adopt their construct and change only its measurement. Rather than inferring a researcher's exposure from proximity in a collaboration graph, which saturates in ML (Appendix~\ref{app:availability_hypergraph_comparison}), we infer it from the researcher's realized output: atoms a researcher has used are direct evidence that those atoms are available to them. The resulting score is a lower bound on availability, since researchers can conceive more than they publish, and it is relative to the community covered by the corpus.

Availability is relational: an idea is available \emph{to someone}. We model this by learning an author--idea compatibility function. For each author $i$, let $R_i$ be the author's repertoire: the union of atoms across all their papers. We randomly split $R_i$ into a query subset $Q_i$ and a complement $R_i\setminus Q_i$. These are two disjoint views of the same research trajectory.

We train two encoders, $\phi$ and $\psi$, that share an atom embedding table. The set encoder $\phi$ embeds the query subset with a bidirectional Transformer without positional encodings, mean-pools the output, projects it, and L2-normalizes it. The author encoder $\psi$ embeds the complement as a larger bag of atoms, mean-pools, applies an MLP, and L2-normalizes. This asymmetry reflects the problem: a candidate idea is a small structured combination, while an author repertoire is a broader distribution over interests.

For a batch of $B$ triples $(i,Q_i,R_i\setminus Q_i)$, we form scaled cosine logits
\[
\ell_{ij}=\tau\,\psi(R_i\setminus Q_i)^\top\phi(Q_j),
\]
with learned temperature $\tau$, and minimize the symmetric InfoNCE loss
\[
\mathcal{L}=\frac{1}{2B}\sum_{i=1}^{B}\left[\mathrm{CE}(\ell_{i,:},i)+\mathrm{CE}(\ell_{:,i},i)\right].
\]

\begin{wrapfigure}{r}{0.40\textwidth}
\vspace{-12pt}
\centering
\begin{tikzpicture}[
  font=\scriptsize,
  >=stealth,
  thick,
  enc/.style={draw, rounded corners=1.5pt, minimum width=5.5mm, minimum height=5mm,
              fill=gray!12, inner sep=1pt},
  vec/.style={circle, draw, fill=white, minimum size=2.8mm, inner sep=0pt}
]
  \draw[fill=red!8, draw=red!50, rounded corners=2pt] (-0.05, 0.45) rectangle (0.65, 0.85);
  \fill[red!75] (0.10, 0.65) circle (1.1pt);
  \fill[red!75] (0.30, 0.65) circle (1.1pt);
  \fill[red!75] (0.50, 0.65) circle (1.1pt);
  \node[font=\tiny, red!70!black, anchor=east] at (-0.07, 0.65) {$Q_i$};
  \coordinate (Q) at (0.65, 0.65);

  \draw[fill=blue!8, draw=blue!50, rounded corners=2pt] (-0.05, -0.20) rectangle (1.05, 0.20);
  \foreach \x in {0.10, 0.30, 0.50, 0.70, 0.90} { \fill[blue!75] (\x, 0) circle (1.1pt); }
  \node[font=\tiny, blue!70!black, anchor=east, align=right] at (-0.07, 0) {$R_i\!\setminus\!Q_i$};
  \coordinate (Ri) at (1.05, 0);

  \draw[fill=black!6, draw=black!35, rounded corners=2pt] (-0.05, -0.85) rectangle (0.85, -0.45);
  \foreach \x in {0.10, 0.30, 0.50, 0.70} { \fill[black!55] (\x, -0.65) circle (1.1pt); }
  \node[font=\tiny, black!70, anchor=east, align=right] at (-0.07, -0.65) {$R_j\!\setminus\!Q_j$};
  \coordinate (Rj) at (0.85, -0.65);

  \node[enc] (phi)  at (1.85, 0.65)  {$\phi$};
  \node[enc] (psii) at (1.85, 0)     {$\psi$};
  \node[enc] (psij) at (1.85, -0.65) {$\psi$};

  \draw[->] (Q)  -- (phi.west);
  \draw[->] (Ri) -- (psii.west);
  \draw[->] (Rj) -- (psij.west);

  \node[vec, fill=red!30]   (vphi)  at (2.85, 0.65)  {};
  \node[vec, fill=blue!25]  (vpsii) at (2.85, 0)     {};
  \node[vec, fill=black!18] (vpsij) at (2.85, -0.65) {};

  \draw[->] (phi)  -- (vphi);
  \draw[->] (psii) -- (vpsii);
  \draw[->] (psij) -- (vpsij);

  \draw[<->, green!50!black] (vphi.south) -- (vpsii.north);
  \node[font=\tiny, green!45!black, anchor=west, inner sep=1pt]
    at (2.99, 0.325) {pull};

  \draw[<->, red!65!black, dashed] (vphi.east)
    .. controls (3.70, 0.55) and (3.70, -0.55) .. (vpsij.east);
  \node[font=\tiny, red!55!black, anchor=west, inner sep=1pt]
    at (3.60, 0) {push};
\end{tikzpicture}
\vspace{-2pt}
\caption{Two-encoder contrastive training for cognitive availability. A held-out query subset $Q_i$ is encoded by $\phi$; the complement $R_i\!\setminus\!Q_i$ and other authors' complements are encoded by $\psi$. Disjoint views of the same author are pulled together, while views from different authors are pushed apart.}
\label{fig:phi-psi-arch}
\vspace{-10pt}
\end{wrapfigure}

Rows ask each author representation to identify its held-out query; columns ask each query to identify its author. We mask off-diagonal entries that share an author, since different subsets from the same author are unlabeled positives rather than true negatives. The disjoint split is crucial: if the author encoder could see $Q_i$, the model could solve the task through raw atom overlap. Holding $Q_i$ out forces the model to learn which atoms tend to belong together within a researcher's trajectory. This mechanism is depicted in Figure \ref{fig:phi-psi-arch}.

At inference, we precompute $\psi(R_i)$ for the 39,791 training authors using their full repertoires. For a candidate atom set $S$, we score $\phi(S)$ against every author by cosine similarity and aggregate the top-$m$ author similarities by their median, $A(S)=\operatorname{median}_{i\in\mathrm{Top}_m(S)}\,\psi(R_i)^\top\phi(S)$. High $A(S)$ means that a community of authors is close to the research direction; we define unavailability as $U(S)=-A(S)$. Taking the median over the top-$m$ rather than the top-$1$ enforces this community reading: in AI/ML, a few authors publish across many subfields and can cover atom combinations that no coherent group supports, so requiring the next-best authors to also match prevents a single generalist from single-handedly marking an idea as available (Appendix~\ref{app:availability_architecture_comparison}).

\subsection{Sampling the Alien Space}
\label{sec:sampling}

Given coherence $C(S)$ and unavailability $U(S)$, sampling alien directions becomes ranking candidate atom sets. For a candidate pool, we compute within-pool $z$-scores $z_C(S)$ and $z_U(S)$, then rank candidates by
\[
F_\beta(S)=(1-\beta)z_C(S)+\beta z_U(S),\qquad \beta\in[0,1].
\]
The parameter $\beta$ controls how aggressively the sampler moves away from the community prior. At $\beta=0$, the sampler selects the most paper-like combinations. At $\beta=1$, it selects the least available combinations, often sacrificing plausibility. The intended regime is intermediate: far enough from existing author communities to be cognitively unavailable, but coherent enough to support investigation. A selected atom set is not yet a readable proposal; we pass the atoms and their canonical descriptions to a decoder LLM, prompted to synthesize a short research-idea sketch that uses every atom (examples in Appendix~\ref{app:example_ideas}).

\section{Experimental Setup}
\label{sec:setup}

\paragraph{Methods compared.}
Every method produces 300 three-atom combinations ($k=3$, close to the corpus mean of 3.38 atoms per paper). \textbf{Alien} scores all $\binom{273}{3}=3{,}353{,}896$ triples and keeps the top 300 under $F_\beta$ with $\beta=0.7$, selected by an originality--coherence sweep (Appendix~\ref{app:beta_selection}). The \textbf{LLM baselines}, Claude Opus 4.7 and Gemini 3.1 Pro, receive the full atom catalogue and are prompted to select novel yet feasible combinations in two regimes: \emph{independent}, with 300 separate calls and the atom order shuffled per call; and \emph{batch}, where all 300 combinations are generated in a single context, so that the model sees, and can diversify, its entire output (Appendix~\ref{app:sampler_setting}). \textbf{Uniform random} draws 300 triples.

\paragraph{Decoding.}
For downstream evaluation, every method's combinations are decoded by the same LLM (Claude Opus 4.7) with the same prompt, so that differences reflect which atoms were selected rather than which model wrote the prose. The method is agnostic to this choice, including open-weight decoders (Appendix~\ref{app:decoder_ablation}).

\paragraph{Evaluation.}
We first validate the representation and learned scores (Section~\ref{sec:validity}) and test whether frozen pre-2025 scores recover 2025 research directions (Section~\ref{sec:future_validation}). We then evaluate portfolios of 300 generated ideas along two axes: \emph{portfolio diversity} and \emph{per-idea quality}.

\emph{Portfolio diversity} measures the variety of research directions explored, capturing the tendency of LLMs to generate homogeneous ideas \citep{jiang2025artificial,zhang2025noveltybench}. Our primary metric is the Vendi score \citep{friedman2023vendi}, which estimates the effective number of distinct ideas in a portfolio. Since each idea consists of three atoms, we compute pairwise Jaccard similarity, taking values of 0, 0.2, 0.5, or 1 when two ideas share zero, one, two, or three atoms, respectively. We construct a $300 \times 300$ similarity matrix and compute its normalized eigenvalues $\{\lambda_i\}$. The Vendi score is defined as $\exp(-\sum_i \lambda_i \log \lambda_i)$, the exponentiated entropy of the eigenvalue distribution. It ranges from 1 (all ideas identical) to 300 (all ideas mutually dissimilar), providing an interpretable measure of portfolio diversity. We complement it with three metrics: \emph{effective atoms}, $\exp(H(p))$, where $p$ is the empirical atom frequency distribution; \emph{repeated pairs}, the number of atom-pair occurrences beyond their first appearance (out of 900); and \emph{near-duplicates}, the fraction of ideas sharing at least two atoms with another idea in the portfolio.

\emph{Per-idea quality} is assessed through expert human ratings, a blind LLM tournament, and blind evaluations of research reports produced by an autoresearch pipeline (Section~\ref{sec:quality}). We also combine portfolio diversity and per-idea quality into a composite score.

\section{Results}
\label{sec:results}

We first validate the representation and learned scores, then test their recovery of held-out 2025 research directions. We then evaluate the generated ideas, showing that Alien sampling produces more diverse portfolios than LLM baselines with competitive per-idea quality.

\subsection{Validating the Representation and Learned Scores}
\label{sec:validity}
\label{sec:score_model_validation}

\paragraph{Idea atoms preserve paper content.}
Reconstructing each paper's methodological summary from its conceptual units alone yields a full mechanism-level match for 97\% of evaluated papers under an LLM judge. Clustering is selected with two LLM-judged tests over the original unit text, an intruder task and a two-cluster partition task, which choose \texttt{bge-large-en-v1.5} embeddings \citep{bge_embedding}, UMAP, and HDBSCAN, giving 273 atoms; an 80\% coverage point raises the mean from 1.65 to 3.38 atoms per paper. This point preserves semantic separability while producing enough co-occurrence signal for downstream models (Appendix~\ref{app:atoms_validation}).

\paragraph{The two scores are distinct and order controlled pools correctly.}
The coherence model ranks exact paper atom sets above random sets, and random sets above pairwise-disjoint sets that never co-occur in training; the availability model ranks sets with high author support above random and zero-support sets, for both $k=3$ and $k=4$ (Figure~\ref{fig:score-model-diagnostics}). The availability model also rejects ``prolific-only'' combinations covered by a single broad author profile but by no community (Appendix~\ref{app:availability_architecture_comparison}).

\paragraph{Availability generalizes to researchers the model has never seen.}
If availability measures what is reachable from a researcher's repertoire, rather than properties of the published record such as coverage or productivity, it should predict the behavior of researchers absent from training. Using models trained only on pre-2025 papers, under the frozen protocol detailed in Section~\ref{sec:future_validation}, we evaluate on the 3{,}186 authors who first appear in our corpus in 2025 with at least two papers, none of whom is in the training set. For each author, we pick three atoms they use across different papers but never combine in one paper, remove these atoms from the author's visible work, and ask the model to match the author to this never-published combination among those of the other unseen authors (coauthors excluded). The model succeeds far above chance (AUC 0.656 author-to-combination and 0.673 combination-to-author; reciprocal rank $4.2$--$4.7\times$ random; $p<10^{-149}$, $n=2{,}779$), and it also matches authors to their actual withheld 2025 papers (AUC 0.551/0.563, $n=1{,}983$). Because the author's visible work shares no atoms with the target, atom-overlap baselines fall below chance (AUC 0.45) and context size alone is at chance (Appendix~\ref{app:heldout_authors}). The score thus captures which combinations are within a researcher's reach beyond their published record, while remaining, by construction, a corpus-relative lower bound.

\subsection{Frozen Pre-2025 Scores Recover 2025 Research Directions}
\label{sec:future_validation}

\begin{figure}[t]
\centering
\includegraphics[width=0.9\textwidth]{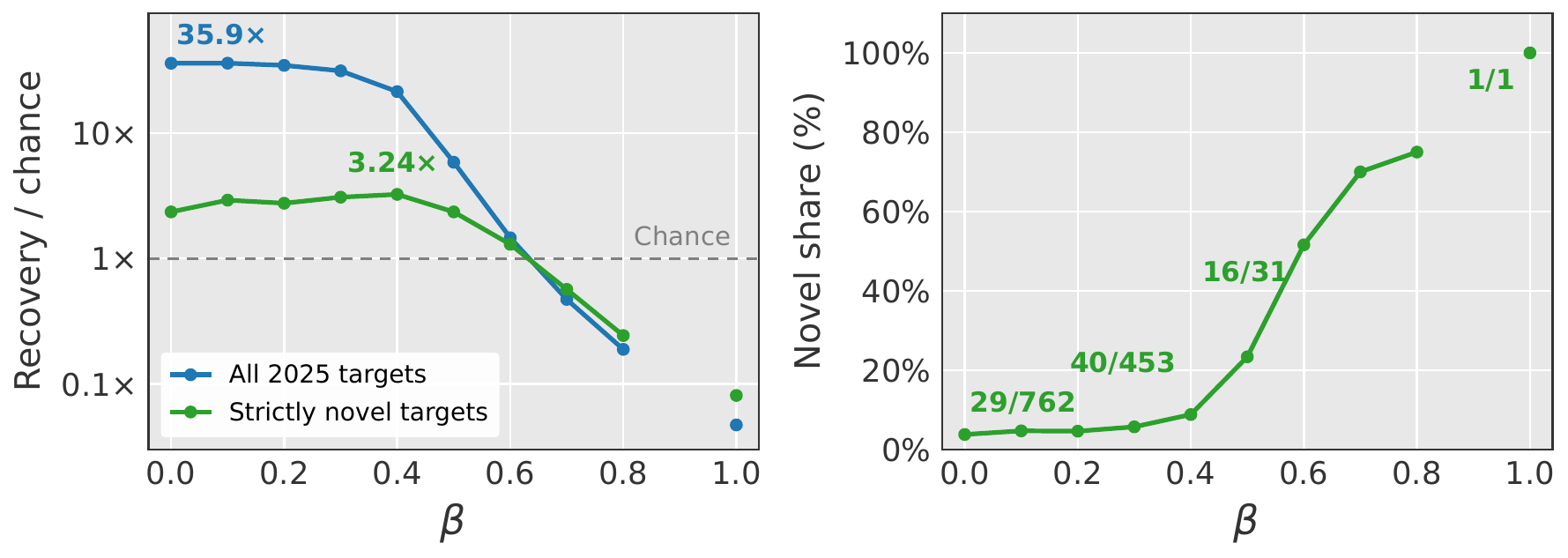}
\vspace{-6pt}
\caption{Frozen temporal validation at $K=10{,}000$ (0.875\% of 1{,}143{,}135 candidates). \textbf{Left:} Recovery relative to uniform random ranking for all 2{,}424 targets and the 1{,}411 strictly novel targets (log scale; dashed line: chance). A triple is strictly novel if its atoms never co-occurred in any pre-2025 corpus paper. \textbf{Right:} Strictly novel targets as a percentage of all recovered targets; labels give novel/total hits. At $\beta=0.9$, no targets are recovered, so zero enrichment and undefined novelty share are omitted.}
\label{fig:future-validation}
\vspace{-8pt}
\end{figure}

We test whether the scores recover research directions published after training. To prevent temporal leakage, we build the vocabulary from the 41{,}321 conceptual units of pre-2025 papers (191 atoms) and train both scorers on papers through 2024. Units from held-out 2025 papers are then mapped to the frozen centroids under the original similarity threshold; they do not change the vocabulary or models (Appendix~\ref{app:frozen_temporal}). Prediction ranks all $N=1{,}143{,}135$ three-atom combinations using only the trained scorers, without LLM calls. The 2025 papers are used solely for evaluation: we compare the top $K=10{,}000$ predictions against their 2{,}424 distinct three-atom sets. Of these, 1{,}411 are \emph{strictly novel}: their three atoms never co-occurred in any pre-2025 paper, including papers with additional atoms.

\paragraph{A small fraction of the space recovers many future combinations.}
With coherence alone ($\beta=0$), selecting just 0.875\% of the candidate space recovers 762 of the 2{,}424 targets (31.4\%), versus 21.2 expected under uniform random ranking: $35.9\times$ chance (Figure~\ref{fig:future-validation}, left). Recovery includes 29 strictly novel targets, versus 12.3 expected ($2.35\times$ chance). Enrichment is the observed number of hits divided by $K n/N$, where $n$ is the size of the target group. Thus the scores concentrate future combinations within a small shortlist, including combinations absent from the training corpus.

\paragraph{Availability increases novelty among recovered targets.}
Giving more weight to unavailability trades overall recovery for a greater proportion of strictly novel hits (Figure~\ref{fig:future-validation}, right). At $\beta=0.4$, the shortlist recovers 453 targets ($21.4\times$ chance), including 40 strictly novel targets ($3.24\times$ chance). Relative to coherence alone, this increases both the number of novel hits, from 29 to 40, and their share of all recovered targets, from 3.8\% to 8.8\%. At $\beta=0.6$, the novel share reaches 51.6\% (16 of 31 hits), although fewer novel targets are recovered in absolute terms. Individual papers illustrate how availability changes the ranking. \emph{Judge Decoding} \citep{bachmann2025judge}, an ICLR 2025 oral that a reviewer called ``a novel approach to speculative sampling'', rises from rank 1{,}489 at $\beta=0$ to rank 1 of 1.14M at $\beta=0.5$--$0.6$, and is still ranked 21st at the sampler's $\beta=0.7$. \emph{Knowledge Entropy Decay} \citep{kim2025knowledge}, also an ICLR 2025 oral, rises from rank 297 at $\beta=0$ to 18 at $\beta=0.4$ (Appendix~\ref{app:frozen_temporal}).

Across the eleven tested $\beta$ values, the union of the top-10{,}000 lists recovers 831 distinct 2025 targets in total, including 80 strictly novel combinations; 51 of the 69 additional targets beyond coherence alone are strictly novel. 

The benefit of moderate availability weighting is therefore selective: it improves recovery of previously unseen combinations while reducing recovery overall. This is consistent with \citet{evans2023accelerating} results. Our model predicts more available combinations than unavailable ones. Estimates of unavailable directions are likely conservative, as they may require a longer evaluation horizon, an effect also observed in \citet{evans2023accelerating}. Given that 2024-2025 account for most LLM-related research in our corpus, we focus on one-year-ahead predictions.

\begin{table}[t]
\centering
\footnotesize
\setlength{\tabcolsep}{3.6pt}
\begin{tabular}{@{}lccccc@{\hspace{8pt}}ccc@{}}
\toprule
& \multicolumn{5}{c}{Portfolio diversity (300 ideas)} & \multicolumn{3}{c}{Diversity--quality composite $\uparrow$} \\
\cmidrule(lr){2-6}\cmidrule(l){7-9}
Method & Atoms & Eff.\ atoms $\uparrow$ & Vendi $\uparrow$ & Rep.\ pairs $\downarrow$ & Near-dup.\ $\downarrow$ & Human & Tournament & Autoresearch \\
\midrule
\textbf{Alien (ours)} & 251 & \textbf{198.0} & \textbf{225.2} & \textbf{106} & \textbf{51.3\%} & \textbf{0.906} & \textbf{0.917} & \textbf{0.977} \\
Claude, batch$^*$ & 235 & 191.7 & 197.9 & 244 & 81.7\% & 0.708 & 0.049 & 0.506 \\
Gemini, batch$^*$ & 120 & 118.1 & 182.7 & 314 & 94.0\% & $-0.467$ & 0.489 & 0.030 \\
Gemini, independent & 118 & 56.2 & 97.4 & 482 & 94.0\% & $-1.084$ & $-0.129$ & $-0.588$ \\
Claude, independent & 85 & 28.7 & 56.5 & 614 & 96.3\% & $-0.316$ & $-0.975$ & $-0.519$ \\
\midrule
\textcolor{gray}{Uniform random} & \textcolor{gray}{266} & \textcolor{gray}{236.3} & \textcolor{gray}{250.4} & \textcolor{gray}{12} & \textcolor{gray}{8.0\%} & \textcolor{gray}{0.252} & \textcolor{gray}{$-0.351$} & \textcolor{gray}{$-0.406$} \\
\bottomrule
\end{tabular}
\caption{Portfolio diversity of each method's 300 ideas, and the equal-weight composite $\tfrac12[z(\text{Vendi})+z(\text{quality})]$ using three evaluation measures (Section~\ref{sec:quality}). Bold marks the best non-random method. Uniform random is the diversity ceiling but optimizes neither coherence nor availability. $^*$Batch variants were not separately rated and inherit the quality of the same model's independent run.}
\label{tab:generation-diversity}
\vspace{-6pt}
\end{table}

\subsection{Alien Sampling Produces More Diverse Research Portfolios}
\label{sec:diversity}

We now evaluate whether Alien broadens exploration into coherent, less-available regions. Table~\ref{tab:generation-diversity} shows that Claude's 300 independently sampled combinations amount to only 56.5 effective distinct ideas, with strong atom concentration (Appendix~\ref{app:diversity}). Batch generation improves diversity: Claude nearly matches Alien on effective atoms (191.7 vs.\ 198.0), but reuses more pairs (244 repeated occurrences vs.\ 106), and more ideas have a near-duplicate (81.7\% vs.\ 51.3\%). Alien achieves 225.2 effective distinct ideas, more than any LLM baseline. Uniform random remains the diversity ceiling but optimizes neither coherence nor availability.

Comparing sets of distinct unordered pairs and triples, Claude--Gemini Jaccard overlap is 11.57\% for pairs and 4.95\% for triples under independent sampling, and 8.85\% and 1.60\% under batch prompting. Alien's overlap with the four LLM conditions is only 0.93--2.33\% for pairs and 0--0.20\% for triples. Alien--Random overlap is similarly low (1.33\% for pairs; no shared triples). These descriptive results show shared concentration across LLM portfolios at both resolutions; low overlap alone does not establish quality. Full counts and protocol details appear in Appendix~\ref{app:combination_overlap}.

\begin{figure}[t]
    \centering
    \includegraphics[width=0.43\textwidth]{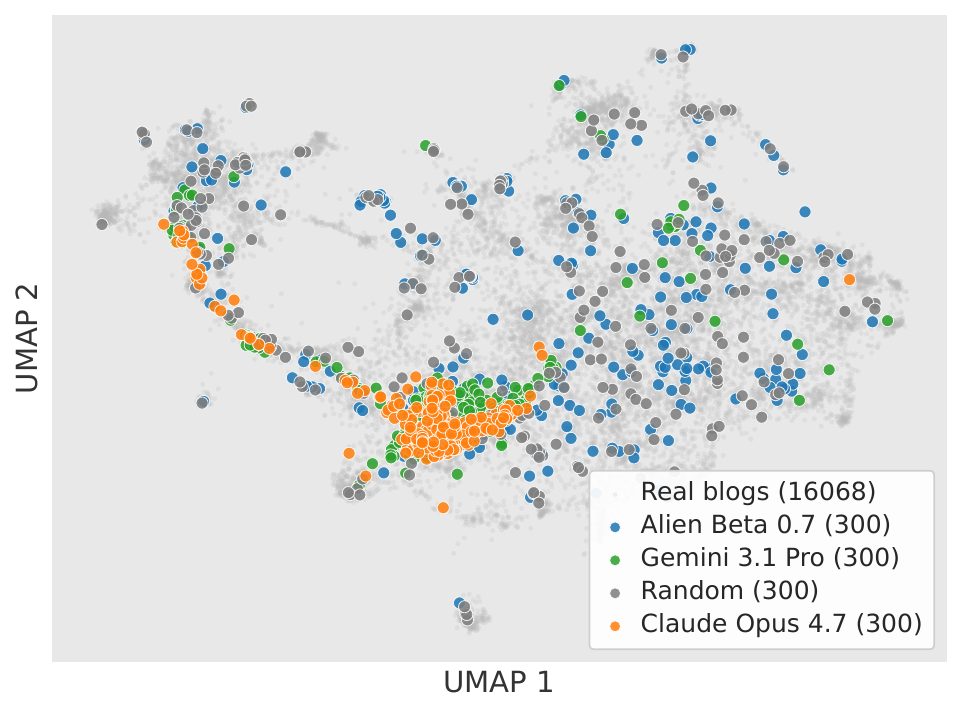}
    \hfill
    \includegraphics[width=0.49\textwidth]{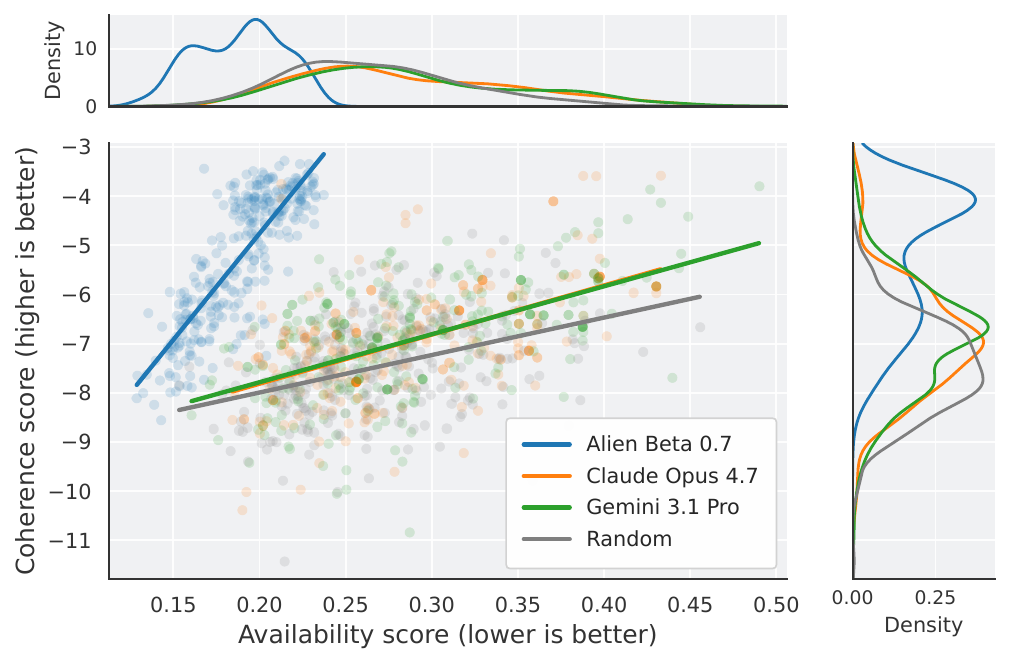}
    \vspace{-6pt}
    \caption{Generated ideas in embedding and score space. \textbf{Left:} UMAP embedding of method digests from the 16{,}068-paper corpus and decoded ideas from Claude, Gemini, Alien, and Random samples. Claude and Gemini concentrate around sparse-autoencoder and interpretability concepts, while Alien samples occupy a broader region. \textbf{Right:} Coherence--availability scores of generated three-atom combinations; higher coherence and lower availability are preferred. Alien concentrates in the intended high-coherence, low-availability region, LLM baselines stay in more available regions, and random sampling loses coherence.}
    \label{fig:coherence-availability-and-umap}
    \vspace{-8pt}
\end{figure}

Figure~\ref{fig:coherence-availability-and-umap} shows where this diversity lies. In embedding space, Claude and Gemini ideas cluster around sparse-autoencoder and interpretability concepts, while Alien ideas spread across the map of the literature. In score space, Alien samples concentrate in the high-coherence, low-availability region; LLM baselines remain in more available regions; and random combinations lose coherence while remaining surprisingly available. In a dense research domain, many random triples are still close to some author community, so reaching the low-availability tail requires explicit search (Appendix~\ref{app:k3_availability_space}). The methods select atoms with similar publication-age profiles, suggesting that Alien's broader exploration does not simply reflect a preference for newer concepts (Appendix~\ref{app:atom_age}).

\subsection{Generated Ideas Remain Competitive in Quality}
\label{sec:quality}

We finally ask whether the broader portfolios retain per-idea quality. We distinguish human-rated quality, LLM-judged coherent originality, and downstream report quality, all blind to method.

\paragraph{Human ratings.}
Ten paid raters with post-graduate degrees in Machine Learning or Computer Science each rated one idea per method, 40 blinded reviews in total over 10 ideas per method; these are the same ideas later run through autoresearch. Alien and Claude tie on overall quality (4.03), followed by Random (3.75) and Gemini (3.68); Alien has the highest feasibility (3.90) and novelty (3.90), Claude the highest coherence (4.30) and promise (4.40) (Appendix~\ref{app:human_survey}). This tie is a per-idea result: a single Alien idea and a single Claude idea read as equally good. The value of the sampler lies in the portfolio, since Alien delivers this quality while covering four times as many distinct directions as Claude queried independently (Table~\ref{tab:generation-diversity}).

\paragraph{Scaling evaluation with calibrated LLM judges.}
Evaluating LLM-research artifacts requires raters with advanced degrees and working knowledge of the LLM literature, and our evaluation runs to thousands of pairwise judgments. Expert raters at that scale are beyond reach, so the larger-sample evaluations below rely on frontier LLM judges, which raises a fair question: does an LLM judge of LLM-mediated outputs track expert opinion? We tested this where the concern is most acute, the autoresearch setting, in which both the research pipeline and the judge are LLMs. We recruited 50 participants online, screened for a graduate degree or doctorate and either verified computer-science expertise or a specialization in AI, ML, or NLP. Each compared one pair of autoresearch reports under the same criteria given to the judge, with methods hidden and presentation order randomized; each of 10 report pairs was judged by five participants (Appendix~\ref{app:judge_validation}). The judge agreed with individual experts on 68\% of judgments (95\% CI 54--80\%, bootstrapped over pairs), more often than two experts agreed with each other (58\%), and it matched the expert majority on 8 of the 10 pairs. The judge is thus not an outlier relative to expert opinion; it tracks the expert consensus more closely than a typical individual rater.

\begin{wrapfigure}{R}{0.30\textwidth}
    \vspace{-10pt}
    \centering
    \includegraphics[width=\linewidth]{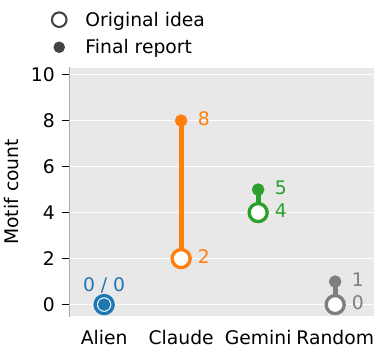}
    \vspace{-15pt} 
    \caption{Interpretability motifs in 10 initial ideas and final reports (after autoresearch loop) per method. (Appendix~\ref{app:motif_concentration}).}
    \label{fig:downstream-evaluation}
    \vspace{-10pt}
\end{wrapfigure}

\paragraph{Blind coherent-originality tournament.}
A GPT-5.5 judge compares decoded ideas in blind forced-choice matches and picks the one that better occupies the coherent-but-cognitively-unavailable region: plausible enough to investigate, but more likely to expose a new framing than to instantiate a familiar recipe. On a balanced 50-idea-per-method tournament, Alien wins 55.6\% of its matches, comparable to Gemini (54.4\%) and above Claude (48.9\%) and Random (41.1\%). The advantage grows in the tail, which is how an ideation tool is used in practice: after keeping the top half of each method's ideas and re-running the tournament, Alien's win rate rises to 72.2\% while Claude's falls to 27.8\% (Gemini 58.9\%, Random 41.1\%), and the pattern persists among the top ten (69.4\% vs.\ 30.6\%; Appendix~\ref{app:idea_pairwise_judge}).

\paragraph{Ideas remain preferable after preliminary experimentation.}
We ran the 40 ideas, 10 per method, through a fixed autoresearch pipeline that iteratively refines each idea, runs bounded experiments on A100 GPU nodes, and writes a standalone research report (Appendix~\ref{app:autoresearch_eval}). The calibrated GPT-5.5 judge compared the anonymized reports in a six-round Swiss-style schedule \citep{fide2026swiss} with both presentation orders, yielding 224 forced choices. Converting preferences to ranks, Alien achieves the best average rank percentile (0.633), followed by Claude (0.574), Gemini (0.495), and Random (0.295), and contributes 3 of the top 5 reports (Appendix~\ref{app:autoresearch_eval}).

We also examine the prevalence of interpretability motifs in the original ideas and final reports after the autoresearch loop (Figure~\ref{fig:downstream-evaluation}), and they reveal a collapse mode in the LLM baselines: Claude shifts from 2/10 ideas to 8/10 reports centered on interpretability, and Gemini from 4/10 to 5/10, while Alien stays at 0/10 (Random: 0/10 to 1/8 completed reports). The collapse happens \emph{during} the research loop: ideas not framed around interpretability are close enough to it in concept space that iterative refinement pulls them toward this familiar attractor, the homogenization documented for LLM outputs \citep{jiang2025artificial}, now appearing in execution. The evaluated Alien ideas do not show this collapse.

\paragraph{Diversity and quality jointly.}
To summarize both axes in one number, we standardize the Vendi score and each quality measure across the six methods and average them with equal weight. Batch variants were not separately rated, so they inherit the quality of the same model's independent run; this is conservative, since if batch generation costs quality, only the baselines' composites fall \citep{zhang2025noveltybench}. Alien ranks first under all three quality measures (Table~\ref{tab:generation-diversity}), and its lead widens under the larger-sample LLM measures. It stays first for any diversity weight up to 0.82, and it Pareto-dominates every LLM baseline: equal or better quality, with strictly higher diversity (Appendix~\ref{app:composite}).

\section{Discussion}

We argue that AI-assisted science should distinguish two quantities that are often conflated: whether an idea is scientifically plausible, and whether the current scientific community is likely to think of it. Modern LLMs are strong at the first kind of modeling. They can synthesize papers, produce coherent proposals, and extend familiar research programs. But because they are trained on the literature and prompted through language, they also inherit the distributional shape of that literature. They tend to search where the community has already placed conceptual mass. The result is a form of ideation that can be useful and fluent, but not necessarily complementary.

The alien space of science names this complementary target: directions that are coherent under existing knowledge but cognitively unavailable under the current community prior.  The claim is not that alien ideas are guaranteed breakthroughs, but distributional: if AI systems are to expand science rather than merely accelerate it, they should move some coherent search mass into regions the community is unlikely to explore on its own.

Alien Sampling is one operationalization of this principle. The method is simple once the distinction is made. Represent research directions compositionally; learn which combinations are coherent; learn which combinations are available to existing authors; and search for high coherence with low availability. The empirical results support this decomposition. Idea atoms preserve paper-level methodological content. Coherence and availability behave as distinct learnable axes. With the vocabulary and models built exclusively from pre-2025 data, the learned scores recover combinations that appear in 2025 papers far above chance, while weighting unavailability steers recovery toward strictly novel combinations. Most importantly, the sampler changes the distribution of generated ideas: it explores a much broader effective atom vocabulary than frontier LLM baselines while remaining competitive under human and downstream execution-based evaluation.

The broader implication is that AI-science systems should be evaluated not only by novelty, quality, or downstream task performance, but by \emph{availability-adjusted discovery value}: how much coherent scientific search mass they move outside the current community prior. An AI scientist that produces more high-quality versions of already-available ideas may improve throughput. An AI scientist that reliably surfaces coherent, low-availability directions can change what the field explores. This axis is especially important as automated research systems become more capable: without an explicit pressure away from the community prior, more powerful systems may simply industrialize the production of familiar ideas.

\paragraph{Why learned scorers rather than prompting.}
Our setting is unusually favorable to the LLM baselines. We restrict the domain to LLM research so that availability can be estimated from near-complete coverage, and the resulting 273-atom catalogue is small enough to fit in a single context window. Broader domains break this assumption: expanding to the AI field as a whole yields roughly 30k atoms \citep{weiss2026lacuna}. At that scale, a prompting approach must split the vocabulary across many calls, so no single call can reason over the whole portfolio. Its cost also scales poorly, because every new search or sampling run requires a fresh round of LLM calls. Alien sampling shifts this cost to a one-time preprocessing step. The LLM distills each paper into atoms once. Two small scorers, a 42.8M-parameter transformer and a dual encoder, are then trained from scratch in a few hours. From that point on, search is much cheaper. For instance, the scorers can rank all 3.35M candidate triples in less than 15 minutes without an LLM call on any individual candidate, and can be reused across runs and objectives. The LLM returns only at the end, to decode the few selected combinations into proposals. In this division of labor, the scorers do most of the exploratory work by selecting promising combinations of atoms. The LLM then works out how to turn those combinations into coherent research proposals.

\paragraph{Limitations and future work.}
The system can only recombine concepts already present in the literature. Producing new primitives would require online vocabulary expansion or mechanisms for creating atoms that are not latent in the corpus. Availability is inferred from publications, which omit private reading, tacit knowledge, unpublished projects, and failed attempts. Our estimates are therefore a lower bound, and they are relative to the community the corpus covers. Our human study is small, so most quality evidence rests on LLM judges that we calibrated against experts on report comparisons. Autoresearch pipelines also remain imperfect proxies for real scientific programs, and a larger expert evaluation is a natural next step. Feasibility is currently assessed only implicitly and late: an infeasible proposal fails during autoresearch, after compute has been spent on it. An explicit tractability critique between decoding and ranking, covering compute, data availability, and known hardness results \citep[e.g.,][]{foster2026ai}, would filter such proposals earlier and compose cleanly with the sampler. Finally, the strongest validation would be longitudinal: whether ideas estimated as coherent but unavailable later become productive research directions. Such evidence takes years to accumulate, so our one-year temporal validation should be read as an early proxy rather than a final test.

\paragraph{Outlook.}
A field's literature records not only what has been discovered, but also a map of what its researchers are likely to notice next. Models trained to imitate human outputs tend toward ideas we would eventually reach ourselves; AI systems can instead be designed to search the coherent blind spots of that map, the regions of idea space that communities struggle to reach because of their historical, social, and cognitive structure, complementing rather than replacing human creativity.

\section*{Ethical Concerns}
Every atom is distilled from peer-reviewed ML papers. The current instantiation therefore recombines public concepts at the granularity of a short research direction, and its outputs are proposals for researchers to evaluate. Other domains carry a specific risk. Availability cannot distinguish combinations that are unexplored because they are cognitively unavailable from those the community avoids for ethical or safety reasons, and in dual-use fields the latter are exactly what a low-availability search could surface. We therefore treat the corpus as the first guardrail and apply the method only to literatures where recombining published ideas is low-risk. Expanding to dual-use domains requires a dedicated safety analysis. We also recommend screening sampled combinations and decoded proposals before release, in addition to relying on the decoder's own refusal behavior.

\setlength{\parskip}{4pt plus 1pt minus 1pt}
\begin{ack}
This work was funded by Tiptree Advanced Systems Corporation. Model access and compute were supported by
Google for Startups Cloud Program credits, Microsoft for Startups Azure credits, and the Max Planck Institute for Human Development, which allowed the use of A100 GPUs for model training and inference.
Alejandro H. Artiles is supported by the Max Planck Artificial Intelligence Network. Martin Weiss is supported by an IVADO's Postdoctoral-Entrepreneur fellowship, 
Chris Pal and Hugo Larochelle are supported by a Canada CIFAR AI Chair.

\paragraph{Competing interests.}
Martin Weiss and Nasim Rahaman are co-founders of Tiptree Advanced Systems Corporation, which develops commercial
research-discovery tools for machine learning researchers. Alejandro H. Artiles performed this work during an internship at Tiptree Systems.
\end{ack}

\bibliography{references}
\bibliographystyle{plainnat}

\appendix
\newpage
\section{Implementation Details}
\label{app:implementation}

\begin{algorithm}[h]
\caption{Alien Sampling Pipeline}
\label{alg:pipeline}
\begin{algorithmic}[1]
\STATE \textbf{Input:} corpus of papers $\mathcal{P}$, researcher profiles $\mathcal{R}$
\STATE \textit{// Representation}
\STATE Extract conceptual units from each paper; cluster into atom vocabulary $\mathcal{V}$
\STATE Map each paper to its expressed atoms $\rightarrow$ atom sequences $\{S_p\}_{p \in \mathcal{P}}$
\STATE \textit{// Model training}
\STATE Train coherence model $C$ on atom sequences from papers
\STATE Train availability model $A$ on atom sets grouped by researcher profile
\STATE \textit{// Alien sampling}
\STATE Sample $N$ candidate sequences from $C$ at temperature $\tau$ (or by random sampling from the space of sets of atoms)
\STATE Score each candidate by coherence $C(S)$ and unavailability $U(S)=-A(S)$
\STATE Standardize both scores within the candidate pool
\STATE Select candidates with high combined coherence--unavailability score given $\beta$
\STATE \textbf{Output:} top-scoring candidates as alien research directions
\end{algorithmic}
\end{algorithm}

\subsection{Data Collection}
\label{app:data_collection}

We construct the corpus in two stages. First, we collect papers from selected ML and NLP venues using OpenReview and DBLP metadata, with Semantic Scholar metadata used to fill missing abstracts, keywords, and PDF links. Second, we apply the LLM field filter over title, abstract, and keywords, retaining papers that match at least one of these metadata fields: "llm", "language model", "language models", "large language model", "large language models", "gpt" and "prompting". This yields 16,068 papers with usable PDF URLs. Table~\ref{tab:corpus_summary} summarizes the selection procedure; Table~\ref{tab:corpus_venue_year} gives the exact venue-year distribution.

\paragraph{Author identities.} Availability is learned from author repertoires, so author identity matters. We identify authors by service-level identifiers rather than name strings: OpenReview profile IDs for 59.4\% of the 46{,}412 authors, DBLP's disambiguated author records (homonyms carry numeric suffixes) for 38.0\%, and an e-mail address for the remaining 2.6\%. The usual name-matching failures, two researchers merged under a shared name or one researcher split across name variants, are therefore largely avoided by construction.

\begin{table}[h]
    \centering
    \begin{tabular}{ll}
        \toprule
        Field & Value \\
        \midrule
        Metadata sources & OpenReview, DBLP, Semantic Scholar \\
        Selection filter & LLM match in title, abstract, or keywords \\
        Venues & NeurIPS, EMNLP, ACL, ICLR, ICML, AAAI, COLM, NAACL, IJCAI \\
        Years & 2017--2025 \\
        Total papers & 16,068 \\
        \bottomrule
    \end{tabular}
    \caption{Corpus construction summary.}
    \label{tab:corpus_summary}
\end{table}

\begin{table}[h]
    \centering
    \scriptsize
    \setlength{\tabcolsep}{3pt}
    \resizebox{\textwidth}{!}{%
    \begin{tabular}{lrrrrrrrrrr}
        \toprule
        Venue & 2017 & 2018 & 2019 & 2020 & 2021 & 2022 & 2023 & 2024 & 2025 & Total \\
        \midrule
        NeurIPS & 0 & 3 & 12 & 19 & 50 & 115 & 348 & 955 & 1,720 & 3,222 \\
        EMNLP & 12 & 32 & 0 & 0 & 0 & 272 & 500 & 954 & 1,253 & 3,023 \\
        ACL & 9 & 18 & 47 & 106 & 0 & 173 & 413 & 636 & 1,221 & 2,623 \\
        ICLR & 16 & 27 & 19 & 55 & 85 & 108 & 285 & 559 & 1,395 & 2,549 \\
        ICML & 2 & 4 & 6 & 8 & 15 & 32 & 110 & 536 & 956 & 1,669 \\
        AAAI & 2 & 13 & 12 & 40 & 52 & 66 & 95 & 307 & 758 & 1,345 \\
        COLM & 0 & 0 & 0 & 0 & 0 & 0 & 0 & 279 & 386 & 665 \\
        NAACL & 0 & 10 & 0 & 0 & 90 & 124 & 0 & 306 & 0 & 530 \\
        IJCAI & 4 & 3 & 9 & 6 & 11 & 26 & 34 & 118 & 231 & 442 \\
        \midrule
        Total & 45 & 110 & 105 & 234 & 303 & 916 & 1,785 & 4,650 & 7,920 & 16,068 \\
        \bottomrule
    \end{tabular}%
    }
    \caption{Distribution of LLM-filtered papers by venue and year. Zero entries indicate years with no papers from that venue in the filtered database.}
    \label{tab:corpus_venue_year}
\end{table}

\subsection{Model usage}
\label{app:model_usage}
Gemini 3 Flash was used for the atomization pipeline: reading papers, producing compressed summaries, extracting conceptual units, evaluating candidate clusters, and naming the resulting atoms. For the LLM baselines, Gemini 3.1 Pro and Claude Opus 4.7 were prompted to select novel yet feasible atom combinations from the atom vocabulary. Claude Opus 4.7 decodes atom combinations into natural-language research ideas for all methods; the decoder ablation (Appendix~\ref{app:decoder_ablation}) additionally uses Gemini 3.1 Pro, Gemini 3 Flash, Kimi K3, GLM 5.2, Qwen 3.6 27B, Llama 3.3 70B, and Ministral 3 8B as decoders. GPT-5.5 is the blind judge of the idea tournament (Appendix~\ref{app:idea_pairwise_judge}) and of the autoresearch reports (Appendix~\ref{app:autoresearch_eval}); it is the judge validated against human experts (Appendix~\ref{app:judge_validation}). Gemini 3.1 Pro is the judge in the $\beta$ sweep, the reconstructor comparisons, and the decoder ablation. The autoresearch pipeline runs on Claude Code with Claude Sonnet 4.6. No language model is involved in the temporal validation or the held-out-author test, which operate only on atom sets and the two small scoring models.

\subsection{Paper Compression}
\label{app:compression}

Each paper is distilled into a high-signal ``blog post'' summary, approximately 2 pages in length. This compression specifically targets methodology over results, removing formatting noise, verbose citations, and granular experimental details while preserving the core research contribution in accessible prose.

\subsection{Conceptual Unit Extraction}
\label{app:extraction}

Conceptual units are extracted from compressed blog posts using an LLM. The number of units per paper is flexible rather than fixed. Each unit must satisfy three quality criteria:
\begin{itemize}
    \item \textbf{Self-standing}: interpretable without the original paper context
    \item \textbf{Recombinable}: can meaningfully pair with units from other papers
    \item \textbf{No dangling references}: avoids paper-specific notation or undefined terms
\end{itemize}

\subsection{Atom Clustering}
\label{app:clustering}

We cluster conceptual units into a shared vocabulary of transferable atoms using BGE embeddings, UMAP dimensionality reduction, and HDBSCAN. HDBSCAN's hyperparameters are selected by the LLM-as-judge clustering evaluation described in Section~\ref{app:atoms_validation}.

\begin{table}[h]
    \centering
    \begin{tabular}{ll}
        \toprule
        Metric & Value \\
        \midrule
        Input conceptual units & 82,255 \\
        Initial HDBSCAN clusters & 273 \\
        Initial noise units & 62.8\% \\
        Selected coverage after reassignment & 80\% \\
        Mean atoms per paper before reassignment & 1.65 \\
        Mean atoms per paper at selected coverage & 3.38 \\
        Clustering algorithm & UMAP + HDBSCAN \\
        Embedding model & BAAI/bge-large-en-v1.5 \\
        \bottomrule
    \end{tabular}
\end{table}

Each cluster is summarized by an LLM into a canonical atom description representing the shared concept. Units initially labeled as noise are incrementally reassigned to their nearest cluster centroid, ordered by embedding similarity, until the selected 80\% coverage point is reached. This reassignment is used only to construct denser paper--atom sequences for training the coherence and availability models. It does not change the atom descriptions.

\begin{figure}[h]
    \centering
    \includegraphics[width=0.7\textwidth]{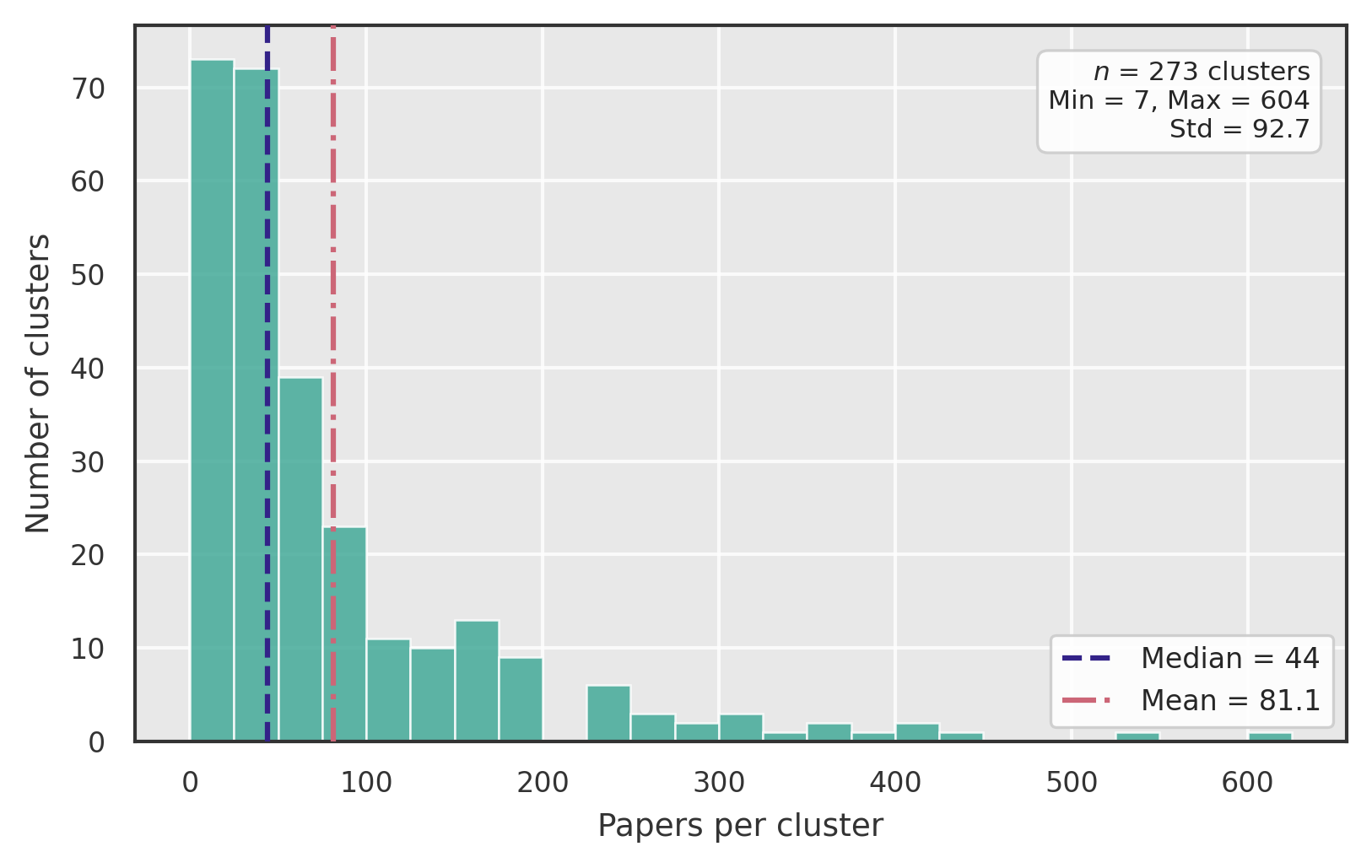}
    \caption{Distribution of papers per initial atom cluster, before noise reassignment. Most atom clusters are supported by tens of papers, while a long tail of broad, cross-cutting clusters pulls the mean (81.1) above the median (44).}
    \label{fig:papers-per-cluster}
\end{figure}

\paragraph{Paper and author coverage over atoms.}
The initial atom clusters, before noise reassignment, are broad but unevenly supported. Figure~\ref{fig:papers-per-cluster} shows the number of distinct papers contributing to each initial cluster. Across the 273 initial clusters, support ranges from 7 to 604 papers with standard deviation 92.7. Most clusters are supported by a few dozen papers: the median cluster appears in 44 papers, and the bulk of clusters draw from roughly 20--50 papers. The mean is higher, 81.1 papers per cluster, because a long tail of large cross-cutting clusters captures widely reused methodological motifs that appear across many subareas. After reassignment to the selected 80\% coverage used for scoring, mean support is 198.2 papers per atom.

\begin{figure}[!htbp]
    \centering
    \includegraphics[width=0.8\textwidth]{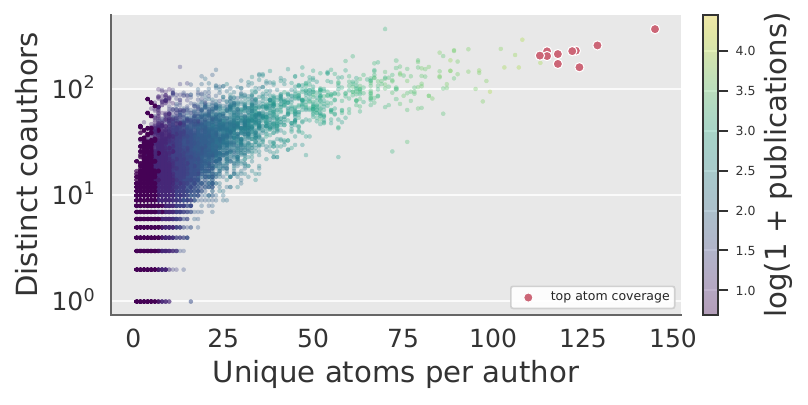}
    \caption{Author connectivity is coupled to atom coverage. Each point is an author node, positioned by unique atom coverage and distinct coauthor degree; color encodes $\log(1 + \mathrm{publications})$, and red points mark the top ten authors by atom coverage. The association shows that broad author profiles are also socially well connected, so individual prolific authors can bridge distant regions of the atom space.}
    \label{fig:author-coverage-connectivity}
\end{figure}

The author side has a similar heavy-tailed structure. The median author node covers only 4 atoms, but the broadest author node covers 145 of 273 atoms; the top 10 author nodes collectively cover 89.0\% of the atom universe. These broad profiles are not merely large atom sets: they are also socially well-connected. Figure~\ref{fig:author-coverage-connectivity} shows that atom coverage, distinct coauthor degree, and publication count are associated. This matters for availability modeling because a single broad author can connect otherwise distant regions of the atom space, even when no broader community independently supports the combination.

\subsection{Training Data}
\label{app:training_data}

Both learned models use the 80\% coverage atom representation described in Section~\ref{app:atoms_validation}. The training JSONL files contain atom IDs and entity IDs, but not venue or year fields, so the quantities below describe the actual model inputs.

\paragraph{Coherence records.}
The coherence dataset contains paper-level atom sequences for papers with at least two assigned atoms. Of the 16,068 corpus papers, 16,020 have mapped atoms; 15,203 satisfy this minimum-length filter and 817 mapped papers are skipped. Each included paper contributes ten serialized sequence variants, producing 152,030 coherence records split 90/10 into 136,827 training records and 15,203 validation records. Atom lengths below exclude the beginning- and end-of-sequence tokens. At the paper level, sequences contain 3.50 atoms on average: 18.3\% have length 2, 33.2\% length 3, 31.2\% length 4, 14.5\% length 5, and 2.8\% length 6--8.

\paragraph{Availability records.}
The availability dataset is built from 46,329 author atom repertoires. An author is included if their profile contains at least two atoms, retaining 44,809 profiles. Each record follows the subset-complement schema: the set tower receives a positive query subset $Q_i$ of length 2--4, while the author tower receives the remaining profile atoms $R_i \setminus Q_i$. This prevents direct overlap between the two towers for the matched pair. Dataset generation produces 47,887,640 distinct author--query records. The loader drops 23,589 records with empty complements, leaving 47,864,051 usable records: 43,077,647 for training and 4,786,404 for validation. The training split contains 39,791 distinct authors; their full repertoires form the inference author bank. The validation split is example-level rather than author-disjoint, so it measures held-out subset-complement pairs rather than retrieval for entirely unseen authors.

\begin{table}[h]
    \centering
    \small
    \setlength{\tabcolsep}{4pt}
    \begin{tabular}{lll}
        \toprule
        Quantity & Coherence data & Availability data \\
        \midrule
        Training objective & Next-atom prediction & Author--set contrastive retrieval \\
        Training records used & 136,827 & 43,077,647 \\
        Validation records used & 15,203 & 4,786,404 \\
        Source entities & 15,203 papers & 44,809 eligible authors \\
        Positive input length & Mean 3.50; median 3; p90 5; max 8 & Mean 3.80; median 4; p90 4; max 4 \\
        Author-tower input length & -- & Mean 27.66; median 24; p90 48; max 143 \\
        Vocabulary size & 273 atoms & 273 atoms \\
        Random seed & 42 & 42 \\
        \bottomrule
    \end{tabular}
    \caption{Training-record characteristics for the coherence and availability models. Lengths count atom IDs, excluding special sequence tokens for coherence. Availability statistics are computed after dropping empty-complement records, matching the records consumed by the dual-encoder loader.}
    \label{tab:training-data-summary}
\end{table}

\subsection{Model Training}
\label{app:model_training}

\subsubsection{Coherence Model.} We train a causal transformer where atoms serve as discrete tokens in a vocabulary of 273. These atoms were identified using the clustering selection procedure described in Section~\ref{app:atoms_validation}. Training uses autoregressive next-atom prediction on randomly ordered atom sequences derived from papers. The coherence model is trained on the 80\% coverage atom representation with the run settings and default architecture details in Table~\ref{tab:coherence-training-params}.

\begin{table}[h]
    \centering
    \small
    \begin{tabular}{ll}
        \toprule
        Parameter & Value \\
        \midrule
        Epochs & 30 \\
        Batch size & 64 \\
        Gradient accumulation steps & 2 \\
        Effective batch size & 128 \\
        Learning rate & $1\times10^{-4}$ \\
        Warmup steps & 1,000 \\
        Weight decay & 0.01 \\
        Maximum sequence length & 32 \\
        Model type & Decoder-only causal Transformer \\
        Vocabulary size & 274 tokens \\
        Hidden size & 768 \\
        Transformer layers & 6 \\
        Attention heads & 12 \\
        Feed-forward hidden size & 3,072 \\
        Dropout & 0.1 \\
        Parameters & 42.8M \\
        \bottomrule
    \end{tabular}
    \caption{Run settings and architecture details for the coherence transformer. Architecture values not passed explicitly in the command use the training script defaults. Vocabulary size is 273 atom tokens plus a shared BOS/EOS token.}
    \label{tab:coherence-training-params}

\end{table}

\subsubsection{Order sensitivity diagnostic.}
\label{app:coherence_order}
Because the coherence model is autoregressive, a single ordered serialization of an atom set can receive a different score from another ordering of the same atoms. We therefore measure order sensitivity by scoring every permutation of the same unordered $k$-atom set and computing the standard deviation of the resulting coherence scores. We compare this within-set order variation to the between-category separation among exact real-paper atom sets, uniformly random atom sets, and pairwise-disjoint incoherent atom sets. For real-paper coherent sets, the average permutation standard deviation is 0.37 for $k=3$ and 0.63 for $k=4$. This is much smaller than the coherent--random mean score gaps (3.52 for $k=3$, 4.18 for $k=4$) and the coherent--incoherent gaps (4.55 for $k=3$, 5.26 for $k=4$). Thus, although single-order coherence scores are not exactly order-invariant, the coherent-set signal is robust relative to random and deliberately incoherent baselines.

\begin{figure}[h]
    \centering
    \includegraphics[width=0.82\textwidth]{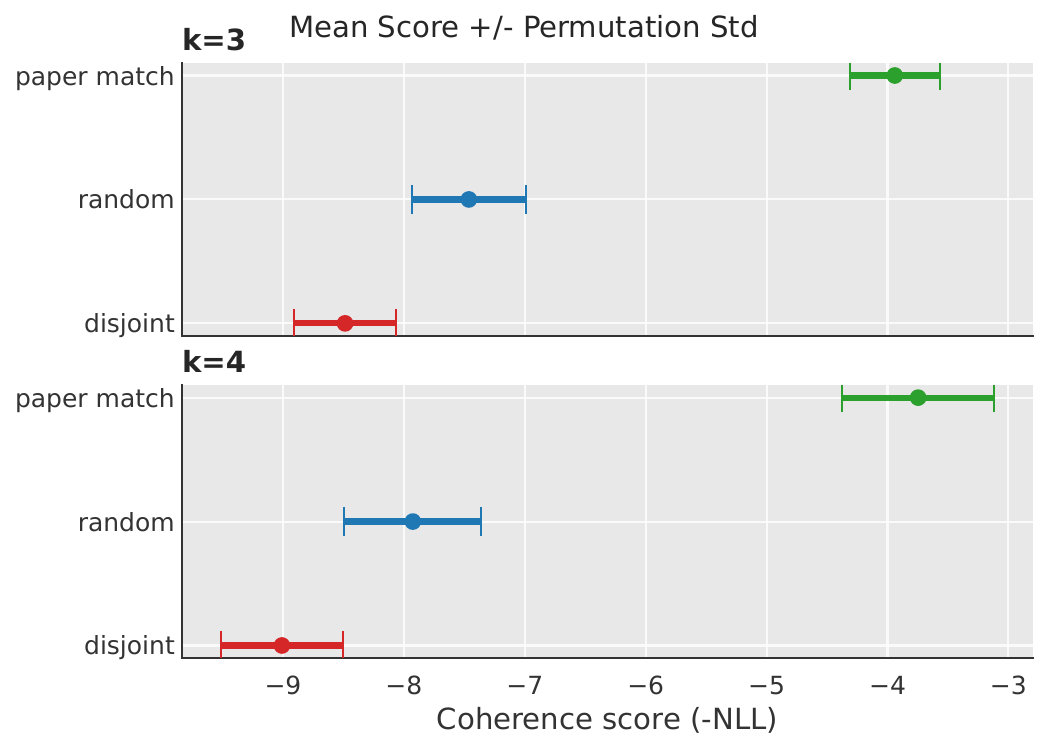}
    \caption{Order sensitivity of coherence scores. Points show the mean coherence score for random, real-paper coherent, and pairwise-disjoint incoherent atom sets; horizontal bars show $\pm 1$ standard deviation across permutations of the same atom set. Scores are $-\mathrm{NLL}$, so higher is more coherent. Although scores vary with atom order, the permutation-induced variation is small relative to the separation between real-paper coherent sets and the random/incoherent baselines.}
    \label{fig:coherence-order-sensitivity}
\end{figure}

\subsubsection{Availability Model.} We train a dual-encoder contrastive model that scores compatibility between an author and a candidate atom set. The author encoder represents the author's known atom repertoire, while the set encoder represents the candidate atoms with a bidirectional Transformer. At inference, low similarity to the nearest author community indicates low cognitive availability. The final availability model uses the settings in Table~\ref{tab:availability-training-params}.

\begin{table}[h]
\centering\footnotesize
\setlength{\tabcolsep}{4pt}
\begin{tabular}{ll}
\toprule
Parameter & Value \\
\midrule
Epochs & 100 \\
Optimizer & AdamW \\
Global batch / per GPU & 4,096 / 1,024 \\
Gradient accumulation steps & 1 \\
Peak learning rate & $5\times10^{-4}$ \\
Learning-rate schedule & 2,000-step linear warmup; cosine decay \\
Weight decay / gradient clipping & 0.03 / 1.0 \\
Query length / author input & 2--4 atoms / full complement (up to 143 atoms) \\
Hidden / feed-forward size & 512 / 3,072 \\
Transformer layers / attention heads & 6 / 8 \\
Dropout / author dropout & 0.1 / 0 \\
Maximum logit scale & 30 \\
Hardware / seed & 4 A100 GPUs / 42 \\
\bottomrule
\end{tabular}
\caption{Training settings for the dual-encoder availability model. The 100-epoch run took 13 h 42 min.}
\label{tab:availability-training-params}
\end{table}

\subsection{Availability Architecture Comparison}
\label{app:availability_architecture_comparison}

Availability is the part of the pipeline where the modeling choice matters most. A coherence model asks whether atoms form a plausible paper. An availability model asks a different question: whether an existing researcher community is positioned to produce the combination. We therefore compare two availability objectives using the same atom vocabulary and matched $k=3$ evaluation pools: an author-agnostic density estimator trained with random negatives, and the author-conditioned dual encoder used in the sampler.

\paragraph{Evaluation protocol.}
We use two complementary evaluations. The broad evaluation exhaustively partitions all 3,353,896 triplets into three pools: \emph{paper-supported} triplets that appear together in a paper; \emph{author-only} triplets covered by at least one author profile but not observed in one paper; and \emph{unavailable}, or zero-support, triplets with no author support. The targeted trap evaluation asks whether a model over-rates combinations with weak community evidence. Here \emph{community-supported} triplets have many supporting authors but no single paper, while \emph{prolific-only} triplets have no paper support and are covered only by one or two broad author profiles, but not by a broader author community. We report AUC-ROC because availability is used as a rank signal: AUC 0.5 means the model cannot order two pools, while higher values mean the positive pool receives higher availability scores. Spearman correlation with author-support count is 0.705 for the dual encoder and 0.224 for the density baseline.

\paragraph{Density estimator with random negatives.}
The density-estimator baseline uses the same unordered set interface as the set side of the dual encoder, but removes the author tower. Let $f_\theta(S) \in \mathbb{R}$ be the scalar score for atom set $S$, computed by embedding the atom IDs, applying a bidirectional Transformer with no positional embeddings or causal mask, mean-pooling over non-padding atoms, and passing the pooled vector through an MLP. For each observed positive set $S_i^+$, the training loader samples $K$ negative sets $S_{i1}^-,\ldots,S_{iK}^-$ with the same cardinality as $S_i^+$. Negative atoms are drawn independently from the empirical atom-frequency distribution $q(a)$, so $S_{ik}^- \sim q^{|S_i^+|}$. The model is trained by classifying the positive set against its random negatives:
\[
    \mathcal{L}_{i}
    =
    -\log
    \frac{\exp(f_\theta(S_i^+)/\tau)}
    {\exp(f_\theta(S_i^+)/\tau)
    +
    \sum_{k=1}^{K}\exp(f_\theta(S_{ik}^-)/\tau)}.
\]
The implementation forms these logits as $[f_\theta(S_i^+), f_\theta(S_{i1}^-), \ldots, f_\theta(S_{iK}^-)]/\tau$, clips them to $[-30,30]$, and applies cross-entropy with the positive set as class zero. Under the noise-contrastive interpretation of this objective, the learned score estimates a density ratio, increasing with $\log p_{\mathrm{data}}(S)-\log p_{\mathrm{noise}}(S)$, where $p_{\mathrm{noise}}$ is induced by the frequency-weighted random sampler. This makes the model strong at separating real-looking combinations from unavailable random controls. However, it has no explicit author input. As a result, it estimates global set familiarity rather than cognitive availability: it cannot answer ``available to whom'', and it completely fails to distinguish paper-supported triplets from author-only triplets (AUC 0.445).

\paragraph{Dual-encoder availability.}
The final model keeps the set encoder but adds a separate author encoder. Training pairs a query subset $Q_i$ with the complement of the same author's repertoire, $R_i \setminus Q_i$, so the task cannot be solved by literal atom overlap. The contrastive objective asks whether the query set retrieves the right author profile and whether the profile retrieves the right query set. At inference, we score a candidate against the 39,791 training-author profiles and aggregate the top 10 similarities by their median. This makes availability a community-compatibility score rather than a single-author bridge.

Table~\ref{tab:availability-architecture-auc} shows the main discrimination tests. The dual encoder provides stronger separation on all three support comparisons. The density baseline distinguishes supported sets from zero-support sets less reliably and reverses the paper-versus-author-only ordering. These measurements support the dual encoder as the availability scorer for this setting.

\begin{table}[h]
\centering\footnotesize
\setlength{\tabcolsep}{4pt}
\begin{tabular}{llccc}
\toprule
Model & Conditioning & Paper $>$ zero & Author-only $>$ zero & Paper $>$ author-only \\
\midrule
Density estimator & Random negatives & 0.683 & 0.708 & 0.445 \\
Dual encoder & Author--set compatibility & \textbf{0.996} & \textbf{0.792} & \textbf{0.950} \\
\bottomrule
\end{tabular}
\caption{AUC-ROC over all 50,036 paper-supported, 2,636,735 author-only and 667,125 zero-support triplets. Both models score exactly the same sets and support labels.}
\label{tab:availability-architecture-auc}
\end{table}

The targeted trap diagnostic in Table~\ref{tab:availability-architecture-trap} asks whether a model rates prolific-only triplets as available. We reuse five deterministic pools of 500 triplets each: paper-supported, community-supported without paper support, prolific-only, zero-support, and random. Prolific-only triplets have one or two supporting authors, all with at least 20 atoms in their repertoires, but no community supporting the combination. The density baseline places 131 prolific-only triplets in its top quartile (21.0\%) and ranks community-supported triplets below prolific-only controls (AUC 0.198). The dual encoder places none in its top quartile and preserves the intended support ordering.

\begin{table}[h]
\centering\footnotesize
\setlength{\tabcolsep}{4pt}
\begin{tabular}{lccc}
\toprule
Model & Community $>$ prolific & Paper $>$ prolific & Prolific in top quartile \\
\midrule
Density estimator & 0.1975 & 0.4293 & 21.0\% \\
Dual encoder & \textbf{0.9999} & \textbf{1.0000} & \textbf{0.0\%} \\
\bottomrule
\end{tabular}
\caption{Prolific-author diagnostic on five matched 500-triplet pools. The top quartile contains the 625 highest-scoring sets across all 2,500 candidates; ties use fixed candidate order. Prolific-only controls have no paper support.}
\label{tab:availability-architecture-trap}
\end{table}

Overall, the density estimator is a useful sanity check: the trained density baseline detects some corpus signal but provides substantially weaker support discrimination. But that objective does not match the construct of cognitive availability. Our sampler needs a score for whether a direction is reachable by an existing researcher community, and this requires conditioning on researcher profiles. We therefore use the dual encoder as the final availability model: it retains strong discrimination against unavailable and prolific-only controls, while supplying the retrieval evidence needed to interpret availability as support from a community of researchers.

\subsection{Atom-Author Hypergraph Comparison}
\label{app:availability_hypergraph_comparison}

The most important structured baseline is the Sourati--Evans human-aware science hypergraph \citep{evans2023accelerating}. We test whether this graph-based signal transfers to our availability problem in a dense collaboration graph like the AI research domain. We use the same atom universe as the dual encoder and the same deterministic $k=3$ candidate pools. Availability asks whether a direction is supported by a plausible researcher community, not merely whether a short social path can connect its atoms.

\paragraph{Hypergraph construction and default setting.}
We construct a publication hypergraph with atom and author nodes. The resulting graph contains 273 atom nodes, 46,329 author nodes, and 16,020 publication hyperedges. Each hyperedge connects the atoms and authors associated with one article. Hyperedges contain 3.38 atoms and 5.98 authors on average. We then learn atom embeddings from author--atom random walks and score a candidate atom set by mean pairwise cosine similarity in the learned availability embedding.

Our main atom-author hypergraph baseline follows the default walk balance described by \citet{evans2023accelerating}: $\alpha=1$, giving equal total sampling weight to atom nodes and author nodes after a hyperedge is selected. We adapt their 250,000-walk protocol to our atom vocabulary by using 916 walks per atom, for 250,068 total walks, using 20 publication-hyperedge transitions per walk. We keep the Word2Vec settings fixed across all alpha values: 128 dimensions, context window 8, 30 epochs, 5 negative samples, and learning rate $0.025 \rightarrow 0.0001$.

\begin{table}[h]
\centering\footnotesize
\setlength{\tabcolsep}{4pt}
\begin{tabular}{lcccc}
\toprule
Method & Comm. $>$ prolific & Comm. $>$ zero & Comm. $>$ random & Control share top Q \\
\midrule
Dual encoder & \textbf{0.9999} & \textbf{0.9999} & \textbf{0.9689} & \textbf{6.4\%} \\
Atom-author W2V, $\alpha=1.0$ (default) & 0.9525 & 0.9597 & 0.9100 & 24.6\% \\
Atom-author W2V, $\alpha=0.5$ & 0.8436 & 0.8397 & 0.7572 & 50.6\% \\
Atom-author W2V, $\alpha=0.1$ & 0.6095 & 0.5794 & 0.5026 & 76.6\% \\
\bottomrule
\end{tabular}
\caption{Availability diagnostics on four matched pools of 500 triplets: community-supported without paper support, prolific-only, zero-support, and random. The final column denotes the fraction of the
highest-scoring quartile that is prolific-only, zero-support, or random.}
\label{tab:sourati-evans-hypergraph-results}
\end{table}

The balanced atom-author hypergraph default is a strong baseline, but the availability-only metrics isolate a clear failure mode. The dual encoder ranks community-supported directions above prolific-only bridges, zero-support controls, and random controls with nearly perfect AUCs. The hypergraph also detects community support at $\alpha=1$, but its high-scoring region contains a substantially larger share of prolific-author trap or noisy candidates. Lower alpha values increasingly overweight author-mediated transitions and degrade all three community-discrimination tests.

\begin{figure}[h]
    \centering
    \includegraphics[width=0.92\textwidth]{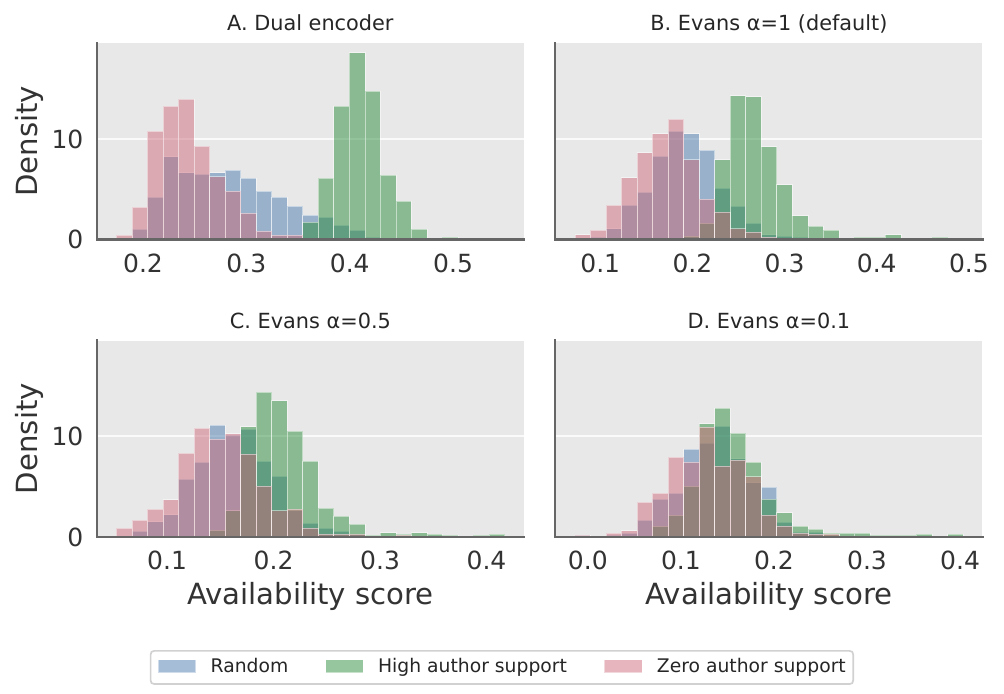}
    \caption{Availability distributions on 1,500 matched triplets: 500 with highest author support, 500 zero-support sets, and 500 random controls. These histogram pools allow paper support and differ from the no-paper community pools in Table~\ref{tab:sourati-evans-hypergraph-results}. The same sets are used for all four methods.}
    \label{fig:availability-dual-encoder-hypergraph-hist}
\end{figure}

\begin{figure}[h]
    \centering
    \includegraphics[width=\textwidth]{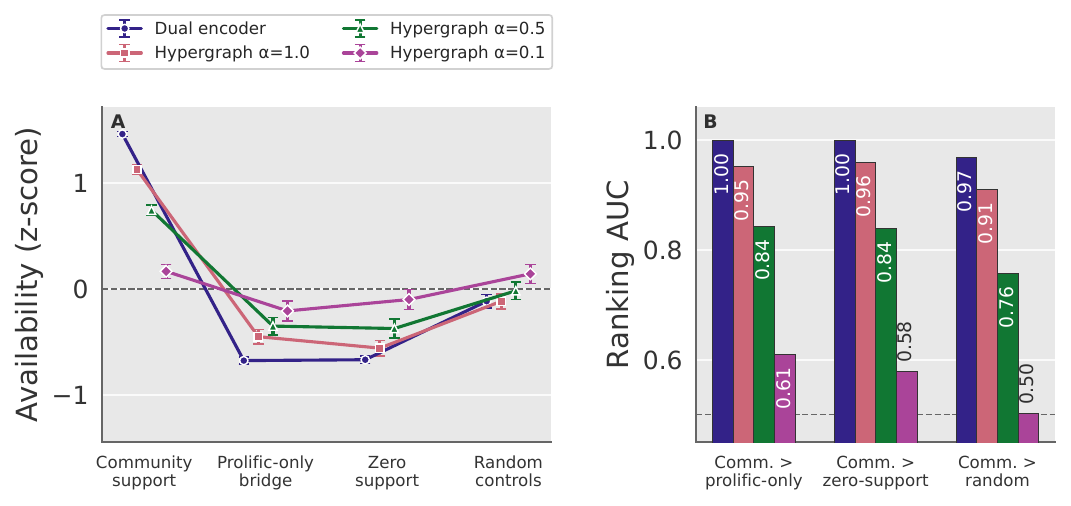}
    \caption{Availability on the four matched support pools of Table~\ref{tab:sourati-evans-hypergraph-results}. Panel A shows per-method standardized means over the 2,000 candidates, with $1.96$ standard-error bars. Panel B gives community-versus-control AUCs. Hypergraph $\alpha=1$ balances author and atom exploration; lower values increasingly weight author transitions.}
    \label{fig:availability-blackhole-robustness}
\end{figure}

\begin{figure}[h]
    \centering
    \includegraphics[width=0.72\textwidth]{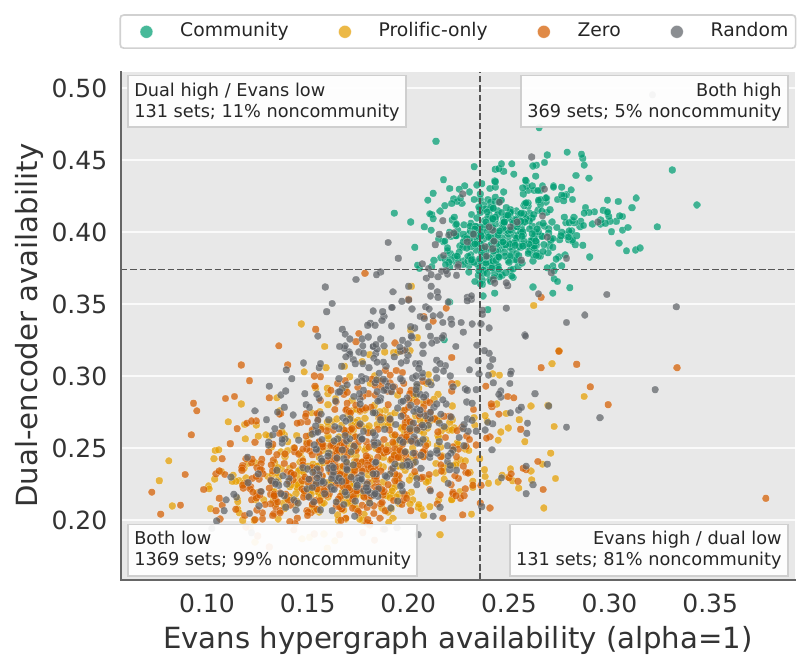}
    \caption{Model-disagreement quadrants on the same 2,000 support-controlled triplets. Dashed lines mark each method\textquotesingle s 75th percentile. The hypergraph-high/dual-low quadrant contains 131 sets, of which 80.9\% belong to control pools; the dual-high/hypergraph-low quadrant also contains 131 sets, with 11.5\% controls.}
    \label{fig:availability-head-to-head}
\end{figure}

\begin{figure}[h]
    \centering
    \includegraphics[width=\textwidth]{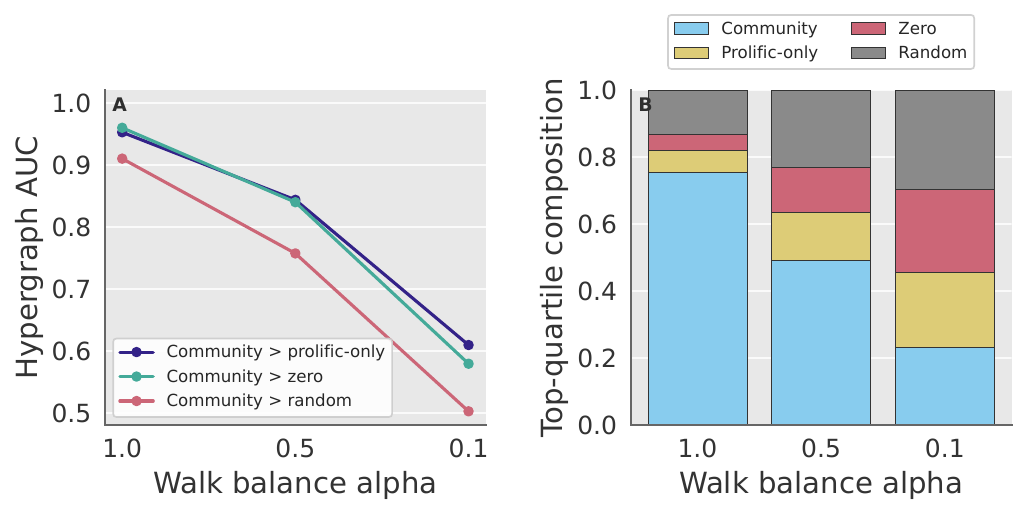}
    \caption{Sensitivity to author/atom walk balance. The balanced atom-author hypergraph default ($\alpha=1$) is the strongest setting. Panel A reports availability-only AUCs for community-supported triplets against three control classes. Panel B shows that lower alpha values increase the fraction of prolific-only, zero-support, and random candidates in the hypergraph high-scoring region.}
    \label{fig:sourati-evans-alpha-ablation}
\end{figure}

\paragraph{Alpha ablation.}
The Sourati--Evans walk parameter $\alpha$ controls the balance between atom (material in their use case) and author exploration after a publication hyperedge is sampled. In this implementation, atom candidates receive total weight $\alpha$ and author candidates receive total weight 1. Thus $\alpha=1$ is balanced, while smaller values overweight author-mediated transitions. Figure~\ref{fig:sourati-evans-alpha-ablation} reruns the hypergraph baseline with matched 250k-walk settings while varying only $\alpha$.

\begin{figure}[h]
    \centering
    \includegraphics[width=0.62\textwidth]{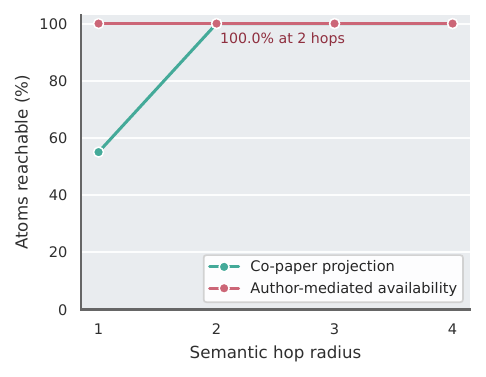}
    \caption{Dense author-mediated paths collapse the atom space. In the shared AI hypergraph, publication incidence already reaches all atoms by two steps, and adding author nodes reaches the full atom universe immediately under the semantic-hop convention. This explains why availability is difficult to infer from graph reachability alone.}
    \label{fig:hypergraph-reachability}
\end{figure}

\paragraph{Why density matters.}
The AI author--atom graph is extremely dense, with the broad author and coauthorship structure summarized in Section~\ref{app:clustering}. Starting from an atom, direct publication incidence reaches 55.1\% of the atom vocabulary at radius 1 and all atoms by radius 2. Once author nodes are included, the author-mediated graph reaches the full atom vocabulary immediately under the semantic-hop convention (Figure~\ref{fig:hypergraph-reachability}). Thus path existence is too weak a signal: many directions are reachable through short social paths even when they are not supported by a coherent author community.

Overall, the atom-author hypergraph is a strong and appropriate structured baseline when run at its balanced default setting. The difference is one of inductive bias. The hypergraph is effective at exploiting publication and author-mediated proximity, but dense author reachability can make unsupported directions look available. The dual encoder instead aggregates similarity across the retrieved author community, so a direction is judged available only when the nearest author neighborhood agrees, not merely when one broad author node bridges the atoms.
\clearpage

\subsection{Availability Space for Three-Atom Combinations}
\label{app:k3_availability_space}

\begin{figure}[h]
    \centering
    \includegraphics[width=0.86\textwidth]{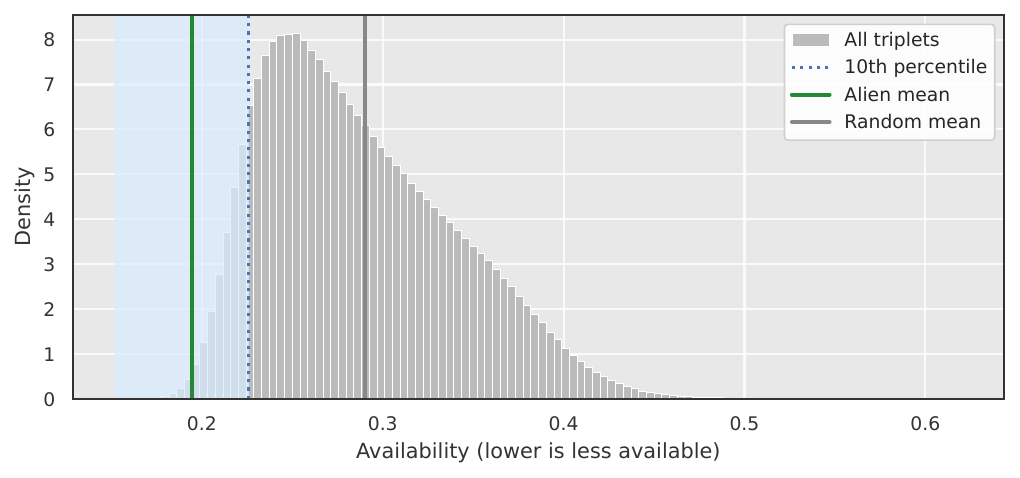}
    \caption{Availability over all 3,353,896 three-atom combinations. Lower scores are less available. The shaded region is the bottom 10\%; vertical lines mark its boundary and the means of the fixed random sample and the numerical Alien top 300 at $\beta=0.7$.}
    \label{fig:k3-availability-space}
\end{figure}

Figure~\ref{fig:k3-availability-space} shows the availability distribution over the full $k=3$ candidate space. This diagnostic explains why random recombination is not enough to obtain cognitively unavailable directions in our domain. The LLM literature is dense enough that many uniformly random triples still fall near some author community under the dual encoder. Consequently, the random mean is more available than the Alien $\beta=0.7$ mean. Low availability is concentrated in a left-tail region of the exhaustive space, so reaching it requires explicit optimization against availability rather than random sampling.

\subsection{Sampling Parameters}
\label{app:sampling}

For $k=3$ we generate from all possible combinations of our 273 atoms and use the coherence and availability models to search in this space. For $k=4$, we sample $N = 10{,}000$ candidate atom sequences from the coherence model at temperature $T = 1$. Coherence is measured by length-normalized log-likelihood under the coherence model, while unavailability is measured as the negative top-community similarity under the availability model. We standardize both scores within the candidate pool and select the top-300 candidates by the combined coherence--unavailability score used in the main text. Each selected atom sequence is reconstructed into a natural language research idea using the reconstruction pipeline described in Section~\ref{sec:sampling}.

\subsubsection{\texorpdfstring{Selecting $\beta$}{Selecting beta}}
\label{app:beta_selection}
The fusion weight $\beta$ controls a real trade-off rather than a nuisance hyperparameter: larger values move the sampler toward lower availability, but eventually sacrifice the coherence constraint that distinguishes alien science from random recombination. We therefore selected $\beta$ before the final baseline comparison using a separate sweep over $\beta \in \{0.0,0.1,\ldots,1.0\}$. For each atom-set size $k$, the sweep held the candidate pool fixed and re-ranked the same candidates under each $\beta$, selecting the top 300 for reconstruction. For $k=3$ the sweep exhaustively scored all 3,353,896 triples; for $k=4$ it scored 52,952 unique sets sampled from 100,000 coherence-model generations. This makes neighboring $\beta$ values directly comparable: only the fusion weight changes.

The evaluation has two separate components. First, we reconstruct the beta-selected atom sets with each candidate reconstructor. This lets us ask whether decoder capacity changes how much signal can be extracted from the same recombinations. Second, within each reconstructor and each $k$, we compare beta values by pairwise originality judgments. For every pair of beta methods, the judge sees two reconstructed ideas from the corresponding beta-selected sets and chooses which is the more original recombination; presentation order is randomized, ties count as half wins, and the resulting tournament is summarized with a position-adjusted Bradley--Terry score. We also compare each beta method against a random baseline. Coherence is not evaluated pairwise: each reconstructed idea is rated independently with a five-level coherence rubric, and we aggregate the per-sample scores for each beta method.

\begin{figure}[h]
    \centering
    \includegraphics[width=0.86\textwidth]{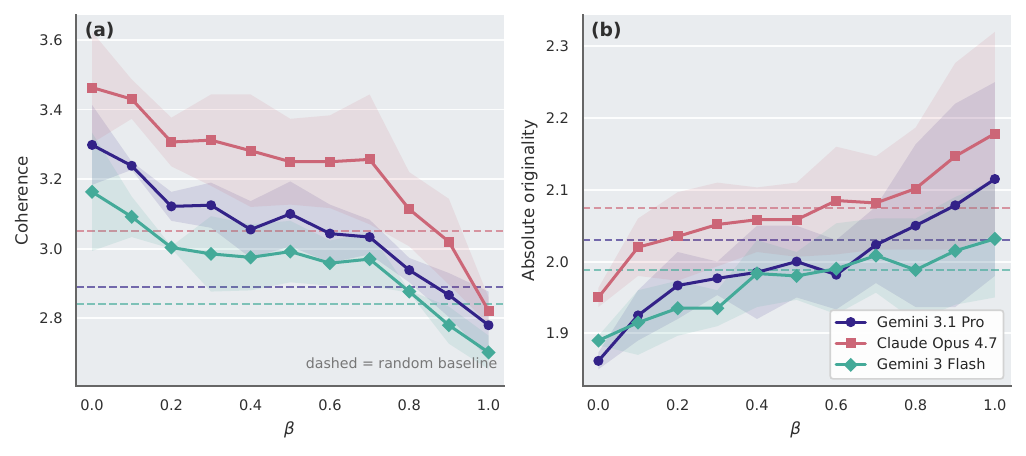}
    \caption{Reconstruction-model capability across beta values. The same beta-selected atom combinations are reconstructed with different decoders. Coherence and absolute originality are rated independently per sample; dashed lines show the random baseline. Stronger reconstructors extract more signal from the same recombinations, but the coherence--originality trend as $\beta$ changes is similar across decoder capacity.}
    \label{fig:reconstruction-model-capability}
\end{figure}

\begin{figure}[h]
    \centering
    \includegraphics[width=0.52\textwidth]{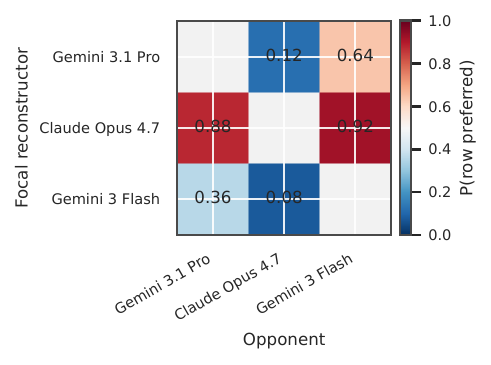}
    \caption{Pairwise originality comparison between reconstructors while holding atom sets fixed. Each cell is the focal reconstructor's win rate against the opponent. Claude Opus 4.7 is the strongest of the three decoders compared here, so the main beta choice is made in the Claude reconstruction setting.}
    \label{fig:reconstructor-winrate-heatmap}
\end{figure}

This separation matters because decoder capacity changes the absolute quality of reconstructions but not the basic beta trend. Figure~\ref{fig:reconstruction-model-capability} shows that stronger reconstructors, especially Claude Opus 4.7, extract more useful signal from a fixed atom-selection procedure: they produce higher coherence and higher absolute originality across most beta values. At the same time, all reconstructors follow the same qualitative pattern as $\beta$ increases: originality rises while coherence falls. Figure~\ref{fig:reconstructor-winrate-heatmap} confirms the reconstructor choice more directly by holding atom sets fixed and running pairwise originality comparisons between decoders. Claude Opus 4.7 wins most head-to-head comparisons against both Gemini 3.1 Pro and Gemini 3 Flash, so we choose $\beta$ in the same decoder setting used for the main reconstructed outputs: the $k=3$ Claude Opus 4.7 beta sweep. A broader comparison including open-weight decoders is reported in Appendix~\ref{app:decoder_ablation}.

\begin{figure}[h]
    \centering
    \includegraphics[width=0.72\textwidth]{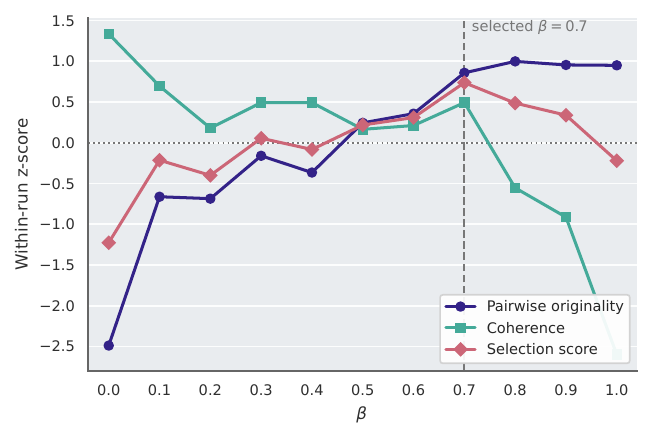}
    \caption{Beta selection for the final $k=3$ Claude Opus 4.7 setting. Pairwise originality is obtained from beta-versus-beta originality comparisons; coherence is rated independently per reconstructed sample. The selection score combines the two standardized signals, and is maximized at $\beta=0.7$.}
    \label{fig:claude-k3-beta-selection}
\end{figure}

Figure~\ref{fig:claude-k3-beta-selection} and Table~\ref{tab:beta-selection-main} show the decisive $k=3$ Claude sweep. We use within-run $z$-scores so pairwise originality and individually judged coherence are on a common scale. In the best-reconstructor setting, $\beta=0.7$ is the best beta under this selection rule. Moving from $\beta=0.7$ to $\beta=0.8$ gives a small additional gain in pairwise originality, but it flips coherence from above-average to below-average and drops the raw coherence score below the random baseline for this run (3.22 vs. 3.32). We therefore choose $\beta=0.7$ for the rest of the paper.

\begin{table}[h]
    \centering
    \small
    \begin{tabular}{rrrrr}
        \toprule
        $\beta$ & Pairwise originality $z$ & Coherence $z$ & Coherence mean & Trade-off \\
        \midrule
        0.6 & 0.36 & 0.21 & 3.38 & 0.31 \\
        \textbf{0.7} & \textbf{0.86} & \textbf{0.49} & \textbf{3.44} & \textbf{0.74} \\
        0.8 & 1.00 & $-0.55$ & 3.22 & 0.49 \\
        0.9 & 0.95 & $-0.91$ & 3.14 & 0.34 \\
        1.0 & 0.95 & $-2.60$ & 2.78 & $-0.22$ \\
        \bottomrule
    \end{tabular}
    \caption{Beta-selection neighborhood for the final $k=3$ Claude Opus 4.7 reconstruction setting. Pairwise originality comes from beta-versus-beta originality comparisons; coherence is rated independently per sample. Both are standardized within the beta sweep. The trade-off column is $0.67 \cdot z_{\mathrm{pairwise}} + 0.33 \cdot z_{\mathrm{coherence}}$; $\beta=0.7$ is best under this criterion. The random baseline's raw coherence mean in this run is 3.32.}
    \label{tab:beta-selection-main}
\end{table}

\subsection{Decoder stability.} As a diagnostic, we measured how consistently the reconstruction decoder maps a fixed atom combination to text. For each generation method, we sampled five atom combinations and generated five independent reconstructions per combination using Claude Opus 4.7 with the same reconstruction prompt. We embedded the resulting texts and computed pairwise cosine similarity only among reconstructions of the same atom combination. The mean within-idea cosine similarity was 0.902, showing that the decoder is generally stable: repeated decodes usually preserve the same semantic research direction.

However, the distribution also shows that the mapping from atom combinations to reconstructed ideas is not strictly one-to-one. Some reconstruction pairs for the same atom combination have substantially lower similarity, with values near 0.8 and a minimum around 0.76. This suggests that a fixed atom set can support multiple nearby, but meaningfully distinct, research interpretations. Thus, while most exploration in our current pipeline comes from the sampler, the decoder itself can introduce limited semantic variation. Future variants could make this source of diversity explicit by prompting for multiple distinct interpretations of the same atom set or conditioning later reconstructions on previous attempts. In this work, we keep the reconstruction protocol fixed for simplicity and use the first decoded reconstruction for each sampled atom combination.

\begin{figure}[h]
    \centering
    \includegraphics[width=1\textwidth]{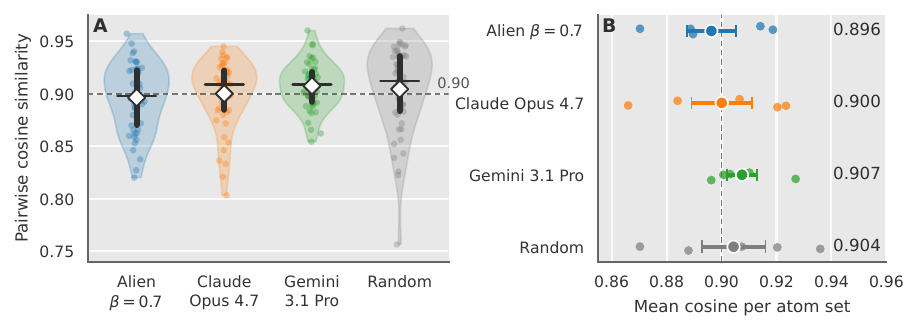}
    \caption{Decoder stability under repeated reconstruction. For each method, five atom combinations were reconstructed five times with Claude Opus 4.7. Panel A shows within-idea pairwise cosine similarities between reconstructions, and Panel B shows atom-combination means with SEM across ideas. The decoder is generally stable, but occasional lower-similarity pairs indicate that a fixed atom combination can admit multiple distinct reconstructions.}
    \label{fig:stability_vs_rating}
\end{figure}

\subsection{Decoder Ablation: Is the Method Decoder-Agnostic?}
\label{app:decoder_ablation}

The sampler selects atom combinations; the decoder only verbalizes them. To check that the pipeline does not depend on one particular, proprietary decoder, we held 50 Alien combinations ($\beta=0.7$) fixed and decoded each with eight models under the identical reconstruction prompt, spanning open and proprietary models from 8B parameters to frontier scale: Claude Opus 4.7, Gemini 3.1 Pro, and Gemini 3 Flash (proprietary), and Kimi K3, GLM 5.2, Qwen 3.6 27B, Llama 3.3 70B, and Ministral 3 8B (open weights, served at provider defaults). All 28 decoder pairs were compared on all 50 combinations, giving 1{,}400 pairwise judgments with presentation order balanced. Gemini 3.1 Pro judged which reconstruction better integrates the three atoms into a single non-obvious mechanism, with instructions not to reward polish or length, and we fit a Bradley--Terry model \citep{bradley1952rank} with a presentation-position term.

\begin{table}[h]
    \centering
    \small
    \setlength{\tabcolsep}{5pt}
    \begin{tabular}{rlccccr}
        \toprule
        Rank & Decoder & Open & Parameters & BT score & Win share & Mean words \\
        \midrule
        1 & Kimi K3 & yes & 2.8T (104B active) & 4.50 & 0.976 & 1{,}340 \\
        2 & Claude Opus 4.7 & no & --- & 2.38 & 0.809 & 1{,}086 \\
        3 & Qwen 3.6 27B & yes & 27B & 0.93 & 0.623 & 1{,}079 \\
        4 & GLM 5.2 & yes & 744B (40B active) & 0.46 & 0.549 & 862 \\
        5 & Gemini 3.1 Pro & no & --- & $-0.42$ & 0.404 & 786 \\
        6 & Ministral 3 8B & yes & 8.4B & $-0.95$ & 0.320 & 1{,}429 \\
        7 & Gemini 3 Flash & no & --- & $-0.97$ & 0.317 & 790 \\
        8 & Llama 3.3 70B & yes & 70B & $-5.94$ & 0.003 & 534 \\
        \bottomrule
    \end{tabular}
    \caption{Decoder ablation on 50 fixed Alien combinations. Parameter counts are from the official model cards (not disclosed for proprietary models). Bradley--Terry (BT) scores are position-adjusted; win share counts ties as half wins; mean words is the average length of each decoder's reconstructions.}
    \label{tab:decoder-ablation}
\end{table}

An open-weight decoder ranks first: Kimi K3 wins 341 of its 350 comparisons and beats Claude Opus 4.7 in 46 of their 50 head-to-head comparisons (3 losses, 1 tie), and two further open models, Qwen 3.6 27B and GLM 5.2, rank above Gemini 3.1 Pro (Table~\ref{tab:decoder-ablation}). Decoding can therefore run entirely on open weights, and since the two scoring models are small networks trained from scratch, sampling has no proprietary dependency; only atomization currently uses a proprietary model (Appendix~\ref{app:model_usage}). Quality is not monotonic in model size: the 8B model ranks near the bottom, the dense 27B Qwen model ranks above the much larger GLM 5.2 and two proprietary models, and Llama 3.3 70B fails under this prompt and provider defaults. The judge ranked its own outputs fifth of eight, so judge self-preference does not seem to drive the ranking.

\clearpage

\section{Conceptual Units and Clustering Validation}
\label{app:atoms_validation}

We validate the representation layer in two parts. First, we verify that conceptual units preserve the methodology of the original distilled blog post. Second, we choose a clustering method, clustering hyperparameters, and reassignment coverage by testing whether candidate clusters form separable semantic groups.

\paragraph{Conceptual-unit reconstruction.} For each paper, we reconstructed the distilled blog post from its paper-specific conceptual units and compared the reconstruction against the original distilled blog. The LLM judge used a five-point scale: full match (5), mostly match (4), partial match (3), minimal match (2), and no match (1), with instructions to evaluate mechanism equivalence rather than exact wording.

Conceptual units received full-match judgments for 97\% of evaluated papers. We interpret this as a validation of the extraction step: conceptual units retain enough paper-specific detail to reconstruct the methodological content of the source blog. We do not use this result to claim that clustered atoms can reconstruct papers; atoms are intentionally more abstract and are evaluated separately as a shared vocabulary.

\paragraph{LLM-judge clustering metrics.}
Clustering quality is semantic: two units should share an atom when they express the same reusable research idea, even if their wording differs. Internal clustering metrics over embeddings do not directly test this property, and the corpus does not provide ground-truth atom labels. We therefore evaluate candidate clusterings using LLM judges over the original unit text, with two complementary metrics:

\begin{itemize}
    \item \textbf{Intruder accuracy (within-cluster purity):} The judge receives four units from a source cluster and one unit from a nearby cluster, and must identify the unit that does not belong. A coherent cluster should make the foreign unit easy to detect.

    \item \textbf{Overlap ARI (boundary separability):} The judge receives four units from each of two nearby clusters and partitions the eight units into two coherent groups. We measure agreement with the original clustering labels using the adjusted Rand index (ARI). If the clusters represent distinct concepts, the judge should recover the same partition.
\end{itemize}

Both tests use nearby clusters rather than random pairs to focus on difficult semantic boundaries. For model selection plots, we combine the two metrics into a single score, $Q=(\mathrm{overlap\ ARI}+\mathrm{intruder\ accuracy})/2$.

\paragraph{Clustering method and hyperparameter selection.} We use the intruder and overlap tests to compare clustering methods and select HDBSCAN hyperparameters. Candidate methods include HDBSCAN on the original embedding space, PCA+HDBSCAN, UMAP+HDBSCAN, and graph clustering variants. For UMAP+HDBSCAN, we sweep UMAP dimensionality and HDBSCAN parameters, including \texttt{min\_cluster\_size}, \texttt{min\_samples}, cluster selection method, and cluster selection epsilon. Each candidate clustering is evaluated with 200 intruder trials and 200 overlap trials.

\begin{figure}[p]
    \centering
    \includegraphics[width=\textwidth]{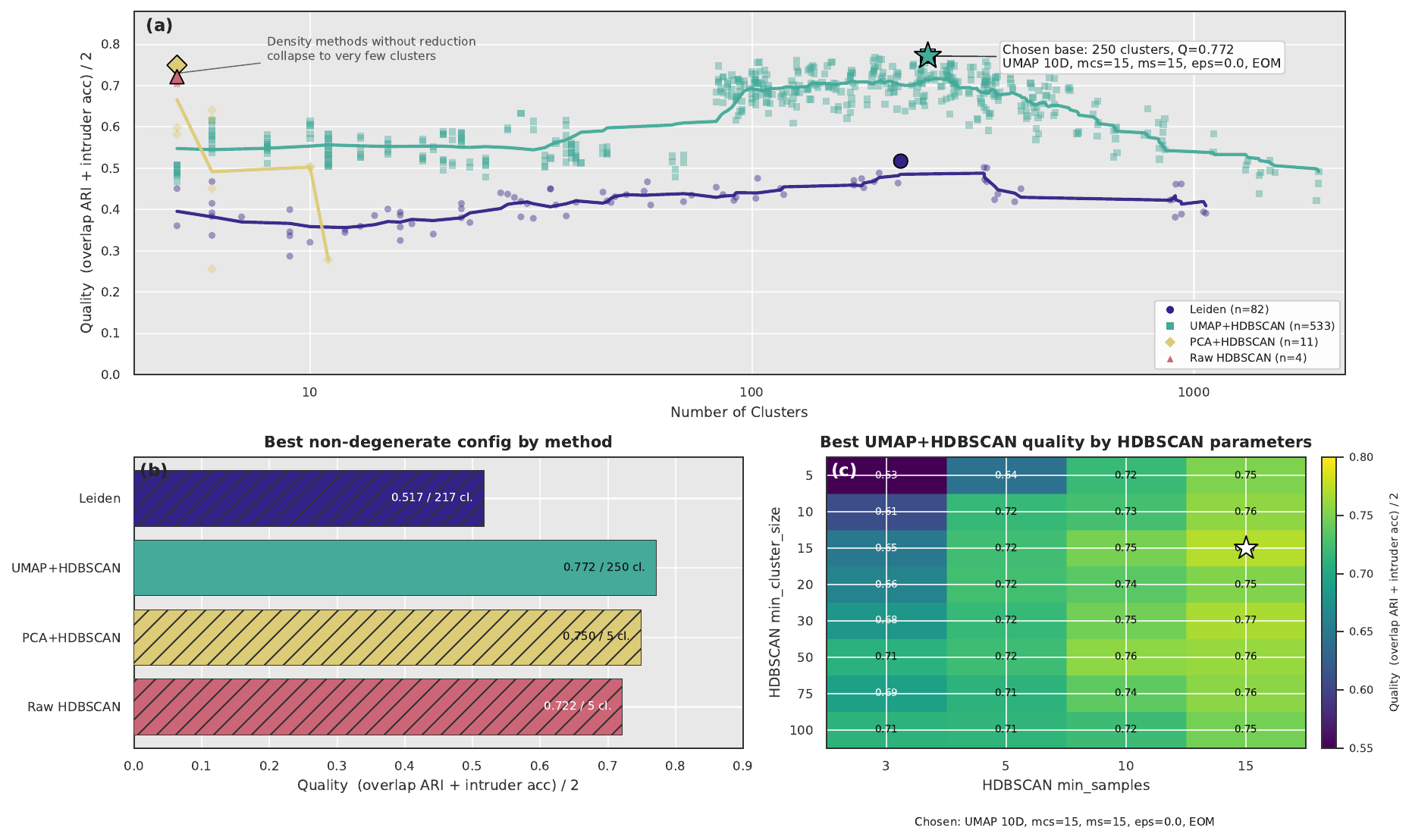}
    \caption{Clustering method and hyperparameter selection. Top: quality as a function of the number of clusters across candidate methods, with the selected UMAP+HDBSCAN base configuration marked. Bottom left: best non-degenerate configuration by method. Bottom right: UMAP+HDBSCAN quality across HDBSCAN \texttt{min\_cluster\_size} and \texttt{min\_samples}. Quality is $Q=(\mathrm{overlap\ ARI}+\mathrm{intruder\ accuracy})/2$.}
    \label{fig:clustering-method-selection}
\end{figure}

\begin{figure}[p]
    \centering
    \includegraphics[width=\textwidth]{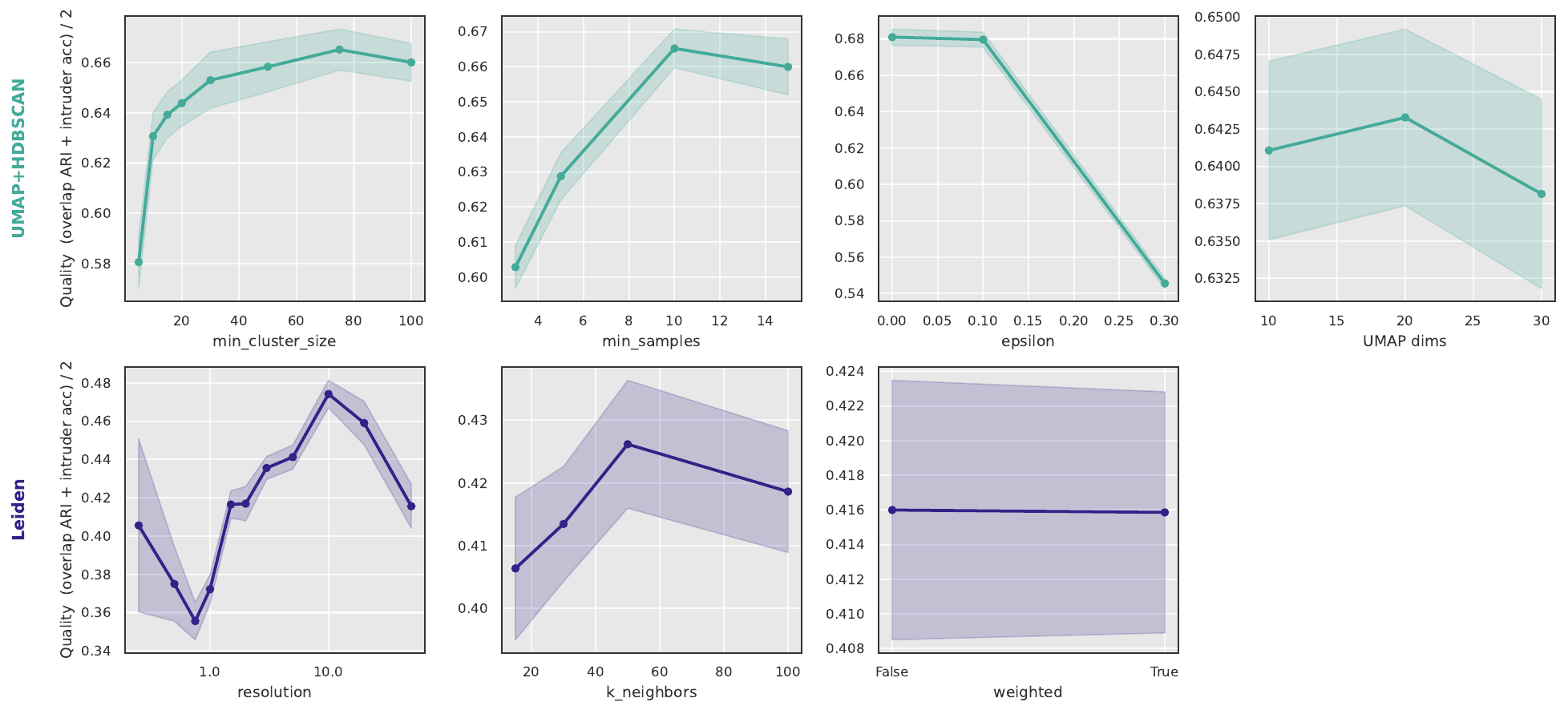}
    \caption{Parameter sensitivity for UMAP+HDBSCAN and Leiden candidates. Lines show mean LLM-judged quality as one parameter varies, with bands indicating variability across the remaining parameters. The selected UMAP+HDBSCAN configuration lies in a broad high-quality region rather than depending on a single narrow parameter setting.}
    \label{fig:clustering-parameter-sensitivity}
\end{figure}

Figure~\ref{fig:clustering-method-selection} summarizes the sweep. Density clustering without dimensionality reduction tends to collapse into very few clusters, which can score well on local purity but is not a useful atom vocabulary. UMAP+HDBSCAN is the strongest non-degenerate family at a useful vocabulary size, and the selected base configuration has $Q=0.772$ with 250 clusters in the sweep run. Re-running this setting for the final corpus export yields the 273 initial clusters reported above. Figure~\ref{fig:clustering-parameter-sensitivity} shows that performance is mostly stable across a broad range of HDBSCAN parameters, with quality improving for moderate-to-large \texttt{min\_cluster\_size} and \texttt{min\_samples}.

The selected configuration uses BGE embeddings, UMAP with 10 dimensions, 30 neighbors, and \texttt{min\_dist=0.0}, followed by HDBSCAN with \texttt{min\_cluster\_size=15}, \texttt{min\_samples=15}, \texttt{cluster\_selection\_method=eom}, and \texttt{cluster\_selection\_epsilon=0.0}. This configuration gives high LLM-judged separability before reassignment: intruder accuracy 0.815 and overlap ARI 0.729 in the hyperparameter sweep.

\begin{figure}[h]
      \centering
      \includegraphics[width=0.7\textwidth]{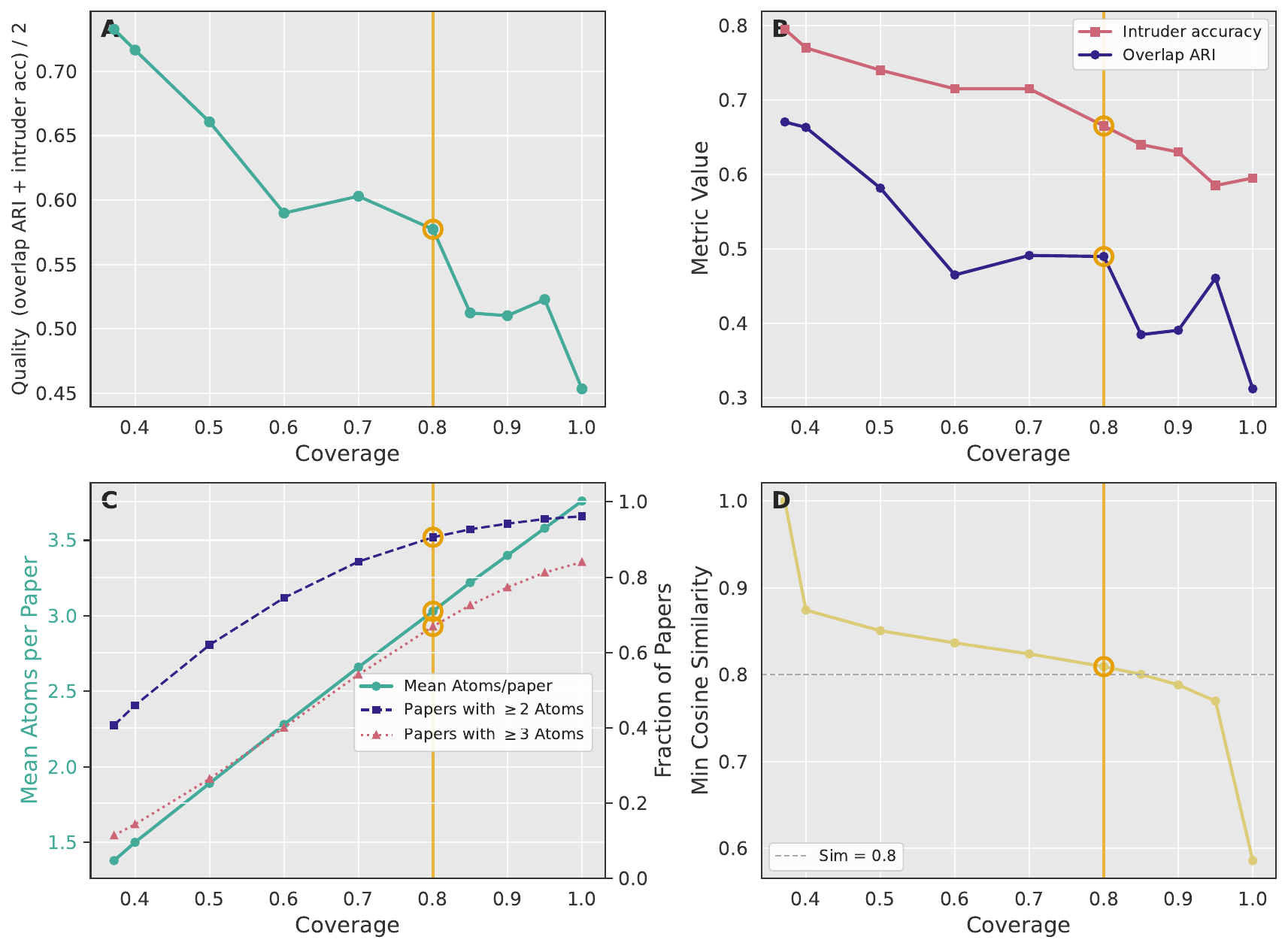}
      \vspace{-6pt}
      \caption{
      Clustering quality as noise units are reassigned to increase atom coverage. Higher coverage increases the
  average number of atoms per paper, which is the key source of training signal for the downstream co-occurrence
  models, but it also makes clusters less semantically pure. We choose 80\% coverage: it increases the average from
  1.65 to 3.38 atoms per paper while keeping the minimum reassignment similarity above 0.8 and preserving reasonable
  LLM-judged cluster separability.}
      \label{fig:clustering-reassignment}
      \vspace{-8pt}
  \end{figure}

\paragraph{Coverage selection by reassignment.} The initial HDBSCAN clustering is conservative: it produces high-purity clusters but leaves 62.8\% of units as noise. To choose a usable training representation, we reassign noise units to their nearest cluster centroid in order of embedding similarity and evaluate coverage checkpoints with the same intruder and overlap tests. The purpose is to increase the amount of paper--atom signal available to the models in training. The main quantity we want to increase is the average number of atoms per paper: when papers have only one or two atoms, they provide little signal about which atoms co-occur. We select the 80\% coverage point because it raises the average from 1.65 to 3.38 atoms per paper, covers 99.5\% of papers with at least one atom and 90.6\% with at least two, while keeping the minimum reassignment similarity above 0.8. Figure~\ref{fig:clustering-reassignment} shows this trade-off.

\clearpage

\section{Score Validity and Temporal Validation}
\label{app:score_validity}

We validate the learned scores in three complementary tests. First, controlled-pool diagnostics test whether the main coherence and availability models recover the intended orderings among atom sets with known paper-co-occurrence or author-support properties. Second, temporal validation tests whether the combined scores recover research directions published in 2025. Third, the held-out-author test evaluates whether availability generalizes to researchers absent from training. The latter two experiments use a pipeline rebuilt entirely from pre-2025 data (Appendix~\ref{app:frozen_temporal}): a vocabulary clustered from pre-2025 conceptual units only, and coherence and availability models trained only on papers through 2024. No language model is involved in scoring or evaluating atom sets in any of the three tests.

\subsection{Controlled-Pool Score Diagnostics}
\label{app:score-diagnostics}

With a usable atom representation in hand, we next validate the two scores that define the search objective. Coherence should identify atom sets that plausibly fit into a single paper. Availability should identify atom sets that current author communities are positioned to generate. Before combining them, we test whether each score recovers the intended ordering on controlled pools.

For coherence, we compare exact $k$-atom paper matches, uniformly random $k$-sets, and pairwise-disjoint sets whose atom pairs never co-occur in any training paper. Because the coherence model is autoregressive and order-sensitive, each set score marginalizes over all $k!$ permutations and reports $-\mathrm{NLL}$, so higher is more coherent. For availability, we compare high-support sets that occur in many author repertoires, uniformly random sets, and zero-support sets absent from every author repertoire. The availability score is the median of the top-10 author--set cosine similarities, so higher means more cognitively available.

\begin{figure}[H]
    \centering
    \includegraphics[width=1\textwidth]{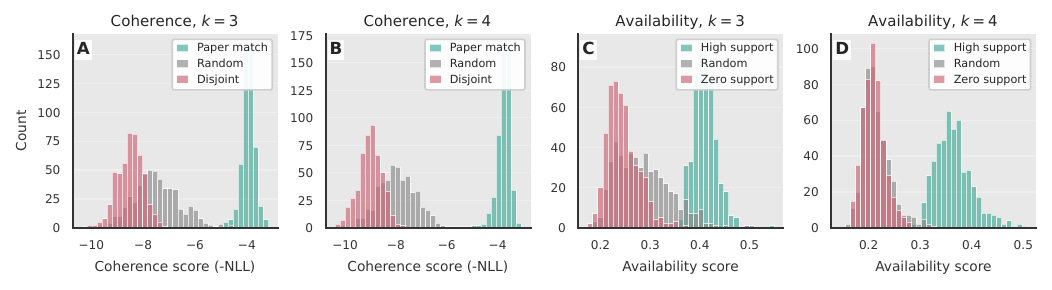}
    \vspace{-6pt}
    \caption{
    Direct diagnostics of the learned coherence model for $k=3$ (a) and $k=4$ (b), and the availability model for $k=3$ (c) and $k=4$ (d). Panels (a)–(b) evaluate the coherence model on paper matches, random combinations, and pairwise-disjoint combinations, while panels (c)–(d) evaluate the availability model on combinations with high author support, random support, and zero author support. Higher scores indicate greater coherence in panels (a)–(b) and greater cognitive availability in panels (c)–(d).
}
    \label{fig:score-model-diagnostics}
    \vspace{-8pt}
\end{figure}

Figure~\ref{fig:score-model-diagnostics} shows that both scores behave as intended at $k=3$ and $k=4$. The coherence model separates exact paper matches from random combinations and scores disjoint combinations lowest. The availability model assigns high-author-support combinations much higher scores than random or zero-support controls. These diagnostics establish that the two objectives carry distinct usable signal before we combine them for alien sampling.

\subsection{Frozen-Vocabulary Temporal Validation}
\label{app:frozen_temporal}

\paragraph{Frozen vocabulary.} We re-run the atomization pipeline of Section~\ref{sec:units} on the 41{,}321 conceptual units of the 8{,}148 papers published through 2024, with the same embedding model, UMAP, and HDBSCAN settings as the main vocabulary. This yields 191 atoms. As for the main vocabulary, pre-2025 noise units are reassigned to their nearest atom centroid in order of similarity until 80\% of units are covered; the lowest accepted cosine similarity, $\tau=0.808$, is frozen as the projection threshold. Centroids use only the original pre-2025 cluster members.

\paragraph{Projecting 2025 papers.} Each of the 40{,}934 conceptual units from 2025 papers is assigned to its nearest frozen centroid if its cosine similarity is at least $\tau$, and is otherwise dropped. No 2025 unit changes a centroid or forms a new atom. This maps 73.6\% of 2025 units and gives 98.9\% of 2025 papers at least one atom. The 2{,}628 papers of 2025 with exactly three mapped atoms define 2{,}424 distinct target atom sets. Of these, 299 are \emph{recurring}: a pre-2025 paper has exactly the same atom set. The remaining 2{,}125 are \emph{exact-new}: no pre-2025 paper has exactly that three-atom representation. Of these, 714 occurred within larger pre-2025 atom sets, leaving 1{,}411 \emph{strictly novel} triples whose atoms never co-occurred in any pre-2025 paper (Section~\ref{sec:future_validation}).

\paragraph{Scoring.} Coherence and dual-encoder availability models are trained exclusively on pre-2025 papers for this separate temporal experiment. We score all $\binom{191}{3}=1{,}143{,}135$ triples (coherence averaged over the $3!$ orderings), standardize within this pool, rank by $F_\beta$, and count exact matches with the 2025 targets among the top $K$. The random expectation is a recall of $K/1{,}143{,}135$. To analyze where recovered targets lie, we split targets into thirds by their score under the frozen availability model.

\begin{table}[h]
    \centering
    \small
    \setlength{\tabcolsep}{5pt}
    \begin{tabular}{rcccccc}
        \toprule
        & \multicolumn{2}{c}{Top 1{,}000} & \multicolumn{3}{c}{Top 10{,}000} \\
        \cmidrule(lr){2-3}\cmidrule(l){4-6}
        $\beta$ & All targets & Exact-new & All targets & Exact-new & Exact-new, least available third \\
        \midrule
        0.0 & 160 (75.4$\times$) & 18 (9.7$\times$) & 762 (35.9$\times$) & 463 (24.9$\times$) & 5 (0.8$\times$) \\
        0.1 & 148 (69.8$\times$) & 10 (5.4$\times$) & 763 (36.0$\times$) & 464 (25.0$\times$) & 10 (1.6$\times$) \\
        0.2 & 129 (60.8$\times$) & 7 (3.8$\times$) & 733 (34.6$\times$) & 434 (23.3$\times$) & 13 (2.1$\times$) \\
        0.3 & 109 (51.4$\times$) & 3 (1.6$\times$) & 663 (31.3$\times$) & 368 (19.8$\times$) & 32 (5.2$\times$) \\
        0.4 & 77 (36.3$\times$) & 3 (1.6$\times$) & 453 (21.4$\times$) & 194 (10.4$\times$) & 43 (6.9$\times$) \\
        0.5 & 27 (12.7$\times$) & 1 (0.5$\times$) & 124 (5.8$\times$) & 40 (2.2$\times$) & 32 (5.2$\times$) \\
        0.6 & 6 (2.8$\times$) & 1 (0.5$\times$) & 31 (1.5$\times$) & 17 (0.9$\times$) & 17 (2.7$\times$) \\
        0.7 & 1 (0.5$\times$) & 0 (0.0$\times$) & 10 (0.5$\times$) & 7 (0.4$\times$) & 7 (1.1$\times$) \\
        \midrule
        Targets & 2{,}424 & 2{,}125 & 2{,}424 & 2{,}125 & 708 \\
        \bottomrule
    \end{tabular}
    \caption{Exact recovery of 2025 three-atom sets with every component frozen at 2024: hits among the top $K$ of all 1{,}143{,}135 candidates, with enrichment over the random expectation. Exact-new targets exclude exact pre-2025 paper representations but may occur within larger earlier atom sets; they differ from the strictly novel targets in Figure~\ref{fig:future-validation}. For $\beta\geq0.8$, at most four targets are recovered.}
    \label{tab:frozen-temporal}
\end{table}

\paragraph{Results.} Table~\ref{tab:frozen-temporal} reports the $\beta$ sweep. With coherence alone, recurring combinations are recovered almost completely (142 of 299 in the top 1{,}000 and all 299 in the top 10{,}000), as expected for directions that keep being combined across years; the exact-new combinations are recovered at $9.7\times$ chance in the top 1{,}000 (exact hypergeometric $p=1.6\times10^{-12}$) and $24.9\times$ in the top 10{,}000. Increasing $\beta$ reduces total recovery but shifts it toward targets far from pre-2025 author communities: in the top 10{,}000, the least available third accounts for 4\% of hits at $\beta=0$, 15\% at $\beta=0.4$, and 43\% at $\beta=0.5$, and recovery of exact-new combinations in that third rises from 5 (below chance) to 43 ($6.9\times$) at $\beta=0.4$. A stricter projection threshold shrinks the target population (348 targets remain at $\tau=0.86$) but leaves recall among the remaining targets essentially unchanged (5.6--6.6\% in the top 1{,}000 and 29--31\% in the top 10{,}000 for $\tau\in[0.808,0.86]$).

\paragraph{Recovered papers.} At $\beta=0$, recovered papers lie in dense, central areas such as tool use, multi-agent reinforcement learning, and efficient training. At moderate and high $\beta$, the recovered set includes work recognized for its novelty and work bridging distant communities. \emph{Judge Decoding: Faster Speculative Sampling Requires Going Beyond Model Alignment} \citep{bachmann2025judge}, an ICLR 2025 oral, rises from rank 1{,}489 at $\beta=0$ to rank 20 at $\beta=0.2$, rank 1 at $\beta=0.5$--$0.6$, and rank 21 at $\beta=0.7$. \emph{Knowledge Entropy Decay during Language Model Pretraining Hinders New Knowledge Acquisition} \citep{kim2025knowledge}, also an ICLR 2025 oral, rises from rank 297 to 18 at $\beta=0.4$. Further examples include vocabulary-guided gait recognition (rank 1{,}100 to 18 at $\beta=0.4$), a clinical-diagnosis language model (rank 1{,}726 to 55 at $\beta=0.5$), and a cross-cultural emotion benchmark whose atom combination had never been used before 2025 (rank 340 at $\beta=0.5$).

\subsection{Held-Out-Author Test}
\label{app:heldout_authors}

\paragraph{Cohort.} Using the frozen pre-2025 availability model, we evaluate the 21{,}794 authors whose earliest paper in our database appears in 2025. Of these, 19{,}342 have at least one paper mapped into the frozen vocabulary and 3{,}186 have at least two; none of them appears in the model's training data. ``First appearing in 2025'' refers to our corpus, not to a researcher's first publication anywhere.

\paragraph{Protocols.} Each author contributes one query per protocol. The candidate pool for an author consists of the targets of the other held-out authors, after removing exact duplicate targets and the coauthors of the target's source papers. We report query-averaged pairwise AUC in both retrieval directions, with 5{,}000 author-level bootstrap samples for confidence intervals, and mean reciprocal rank (MRR) relative to its exact random expectation.
\begin{enumerate}[leftmargin=*]
    \item \textbf{Never-published combination.} Three atoms that the author used across at least two of their papers but never within a single paper; the three atoms are removed from the author's visible repertoire, and at least two other atoms must remain.
    \item \textbf{Withheld real paper.} One of the author's papers is held out and the others form the context, restricted to authors whose held-out paper shares no atom with the context.
    \item \textbf{Strict disjoint papers.} Context and target are built from disjoint sets of the author's papers, and the target uses only atoms exclusive to the held-out side.
\end{enumerate}

\begin{table}[h]
    \centering
    \small
    \setlength{\tabcolsep}{5pt}
    \begin{tabular}{lrccc}
        \toprule
        Protocol & $n$ & AUC, author $\rightarrow$ target & AUC, target $\rightarrow$ author & MRR / random \\
        \midrule
        Never-published combination & 2{,}779 & 0.656 [0.646, 0.666] & 0.673 [0.663, 0.683] & 4.2$\times$ / 4.7$\times$ \\
        Withheld real paper & 1{,}983 & 0.551 [0.539, 0.564] & 0.563 [0.550, 0.576] & 1.7$\times$ / 1.6$\times$ \\
        Strict disjoint papers & 762 & 0.592 [0.572, 0.612] & 0.606 [0.586, 0.625] & 1.9$\times$ / 1.8$\times$ \\
        \midrule
        Atom-overlap baselines & & 0.445--0.451 & 0.445--0.451 & 0.2--0.3$\times$ \\
        Context size only & & 0.500 & 0.500 & --- \\
        \bottomrule
    \end{tabular}
    \caption{Held-out-author test with 95\% bootstrap confidence intervals. Chance AUC is 0.5. All model AUCs exceed chance with Wilcoxon $p<10^{-14}$. Overlap baselines (atom Jaccard, binary atom cosine, overlap count) fall below chance because the target shares no atoms with the visible repertoire by construction.}
    \label{tab:heldout-authors}
\end{table}

\paragraph{Results.} The availability model matches unseen authors to combinations within their reach far above chance in all three protocols (Table~\ref{tab:heldout-authors}). The signal is strongest for never-published combinations, which no record of the author's work contains, and remains above chance for real withheld papers and under the strict paper split. Literal-overlap baselines and context size cannot solve the task, so the result is not explained by seeing the target atoms, by prolific authors, by coauthor retrieval, or by recognizing an author seen in training. The test supports the predictive validity of the availability score as a corpus-relative lower bound on what is within a researcher's reach; it does not measure researchers' mental states directly.

\clearpage

\section{Alien Ideas Evaluation}
\label{app:sampler_eval}

We evaluate the Alien sampler against baselines along two axes, portfolio diversity and per-idea quality (LLM judging, human ratings, and downstream execution), and combine them in a composite score.

\subsection{Experimental Setting}
\label{app:sampler_setting}

All methods produce 300 combinations of three atoms, close to the mean of 3.38 atoms per paper.
\begin{itemize}
    \item \textbf{Alien:} all $\binom{273}{3}=3{,}353{,}896$ three-atom sets are scored, and the top 300 under $F_\beta$ with $\beta=0.7$ are kept.
    \item \textbf{Claude Opus 4.7 and Gemini 3.1 Pro, independent:} each model is queried 300 times. In each query, the full catalogue of 273 atom descriptions is provided in context, in an order shuffled per call to mitigate positional bias, and the model is prompted to select one combination that is both novel and feasible.
    \item \textbf{Claude Opus 4.7 and Gemini 3.1 Pro, batch:} a single call provides the full catalogue (shuffled with a fixed seed) and asks for exactly 300 combinations of three distinct atoms under the same coherence, novelty, and integration criteria, returned as JSON (temperature 1.0). Because the whole collection is generated in one context, the model sees every combination it has already proposed. The model is prompted to select 300 distinct combinations that are novel and feasible. Both models returned 300 valid entries in one call.
    \item \textbf{Uniform random:} 300 three-atom sets drawn uniformly.
\end{itemize}
For downstream evaluation, the selected combinations of every method are decoded into natural-language research ideas by the same decoder and prompt (Section~\ref{sec:sampling}).

\subsection{Diversity Analysis}
\label{app:diversity}

\paragraph{Metrics.} For a method's 300 combinations, let $p(a)$ be the empirical frequency of atom $a$ over all atom slots. \emph{Effective atoms} is $\exp(H(p))$, the number of equally frequent atoms with the same entropy. The \emph{Vendi score} \citep{friedman2023vendi} is $\exp\bigl(-\sum_i \lambda_i\log\lambda_i\bigr)$, where $\lambda_i$ are the eigenvalues of $K/300$ and $K_{jl}$ is the Jaccard similarity of combinations $j$ and $l$; it equals 300 if all combinations are disjoint, 1 if all are identical, and $m$ if the set collapses to $m$ disjoint templates. \emph{Repeated pairs} counts atom-pair occurrences beyond each pair's first appearance (out of $3\times300=900$ pairs), and \emph{near-duplicates} is the share of combinations for which some other combination shares at least two of the three atoms. We also report the share of combinations that contain one of the method's ten most-used atoms (\emph{top-10 concentration}).

\begin{figure}[h]
    \centering
    \includegraphics[width=\textwidth]{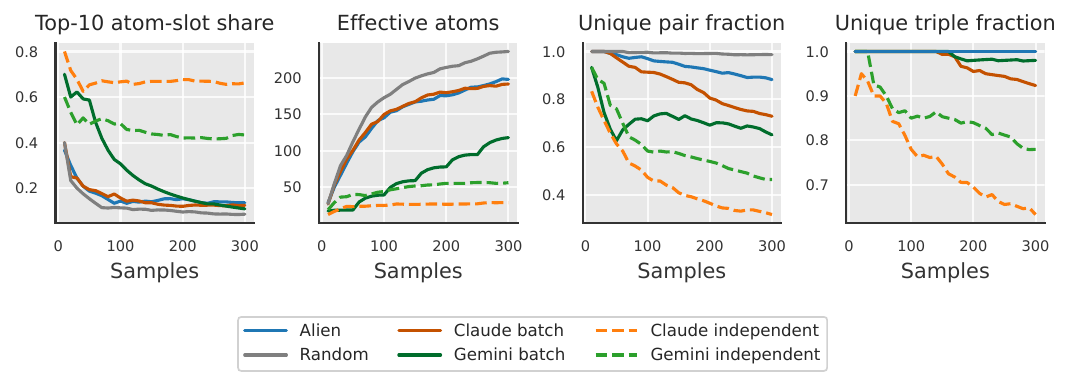}
    \caption{Diversity accumulation for the same 300-combination portfolios as Table~\ref{tab:diversity}: top-10 atom-slot share, effective atoms, and unique-pair and unique-triple fractions.}
    \label{fig:generation-diversity-dynamics}
\end{figure}

\paragraph{Results.} Table~\ref{tab:diversity} complements Table~\ref{tab:generation-diversity} and Figure~\ref{fig:generation-diversity-dynamics} shows how the measures evolve as samples accumulate.

\begin{table}[h]
    \centering
    \small
    \begin{tabular}{lcccc}
        \toprule
        Method & Atoms used & Unique triples & Top-10 concentration $\downarrow$ & Vendi $\uparrow$ \\
        \midrule
        Alien ($\beta=0.7$) & 251 & 300 & 34.3\% & 225.2 \\
        Claude Opus 4.7, batch & 235 & 277 & 32.3\% & 197.9 \\
        Gemini 3.1 Pro, batch & 120 & 294 & 28.0\% & 182.7 \\
        Gemini 3.1 Pro, independent & 118 & 234 & 76.7\% & 97.4 \\
        Claude Opus 4.7, independent & 85 & 190 & 95.7\% & 56.5 \\
        Uniform random & 266 & 300 & 25.0\% & 250.4 \\
        \bottomrule
    \end{tabular}
    \caption{Additional diversity statistics of each method's 300 combinations.}
    \label{tab:diversity}
\end{table}

We quantify a critical limitation of LLMs: when prompted to select novel atom combinations, they repeatedly favor the same atoms, limiting diversity. The Alien sampler has higher Vendi than the LLM baselines, while uniform random remains the diversity ceiling.

Analysis of independently sampled LLM selections reveals systematic biases:
\begin{itemize}
    \item Both LLM baselines select Sparse Autoencoders as their top atom: Claude includes it in 42.67\% of samples and Gemini includes it in 25.67\%.
    \item Claude concentrates heavily on mechanistic-interpretability and reasoning-verification atoms, including concept vectors, activation patching, Process Reward Models, and symbolic verification.
    \item Gemini is less concentrated than Claude but still repeatedly selects Sparse Autoencoders, Flow Matching, activation steering, machine unlearning, and biophysical-constraint modeling.
    \item The Alien sampler has lower peak atom concentration than the independent LLM baselines: its two most frequent atoms appear in 6.67\% and 6.00\% of samples. Its top atoms concern reversal reasoning, textual optimization, and data curation.
\end{itemize}

\paragraph{Top Atoms by Method.} Below we show the three most frequently selected atoms for Alien and the two independently sampled LLM baselines. Percentages in parentheses indicate the fraction of the 300 samples that included the atom. Atom labels are typeset in sans serif to distinguish canonical atom identities from ordinary prose.

{\small
\paragraph{Claude Opus 4.7.}
\begin{enumerate}[leftmargin=*, itemsep=3pt, parsep=0pt]
  \item \textsf{\textbf{Atom 111}} \textsf{(42.67\%)}: \emph{Sparse Autoencoders (SAEs) resolve the problem of neural superposition—where individual neurons simultaneously encode multiple, overlapping concepts—by functioning as dictionary learners that project dense, polysemantic model activations into a higher-dimensional latent space constrained by sparsity. This mechanism decomposes complex internal representations into discrete, 'monosemantic' features, where each active dimension corresponds to a single, human-interpretable concept. By isolating these individual computational primitives, SAEs enable researchers to transparently audit a neural network's internal logic, map causal feature circuits, and perform precise, surgical interventions on specific behaviors without the need for opaque fine-tuning.}
  \item \textsf{\textbf{Atom 178}} \textsf{(34.00\%)}: \emph{High-level semantic concepts and behavioral traits in neural networks are encoded as distinct linear directions within high-dimensional activation spaces, a phenomenon formalized as the Linear Representation Hypothesis. These specific 'concept vectors' can be mathematically isolated through contrastive activation analysis—calculating the vector difference between the internal states of paired inputs representing opposing traits (e.g., true versus false, or safe versus harmful) to cancel out shared syntactic and contextual noise. Once isolated, these vectors enable direct operational control, allowing researchers to mathematically probe a network's latent cognitive states or surgically steer its behavior during generation by adding or subtracting the concept vector, entirely bypassing the need for model retraining.}
  \item \textsf{\textbf{Atom 183}} \textsf{(24.33\%)}: \emph{Activation patching—along with variants like path patching and interchange intervention—is a causal interpretability technique that isolates the functional components of a neural network by surgically transplanting internal activations from a source execution run into a counterfactual or corrupted target run; if this targeted substitution alters or restores the model's final prediction to match the source, it definitively identifies the patched layers, tokens, or neural circuits as the causally responsible pathways for that specific reasoning step or behavior.}
\end{enumerate}

\paragraph{Gemini 3.1 Pro.}
\begin{enumerate}[leftmargin=*, itemsep=3pt, parsep=0pt]
  \item \textsf{\textbf{Atom 111}} \textsf{(25.67\%)}: \emph{Sparse Autoencoders (SAEs) resolve the problem of neural superposition—where individual neurons simultaneously encode multiple, overlapping concepts—by functioning as dictionary learners that project dense, polysemantic model activations into a higher-dimensional latent space constrained by sparsity. This mechanism decomposes complex internal representations into discrete, 'monosemantic' features, where each active dimension corresponds to a single, human-interpretable concept. By isolating these individual computational primitives, SAEs enable researchers to transparently audit a neural network's internal logic, map causal feature circuits, and perform precise, surgical interventions on specific behaviors without the need for opaque fine-tuning.}
  \item \textsf{\textbf{Atom 9}} \textsf{(22.33\%)}: \emph{Flow Matching unifies diverse generative and alignment tasks by learning a continuous vector field that defines deterministic, optimal transport trajectories between a source distribution, such as random noise or sub-optimal model states, and a target data distribution. By modeling transformations as Ordinary Differential Equations that move samples along mathematically efficient, straight-line or geodesic paths, this approach bypasses the complex, stochastic iterative denoising of traditional diffusion models and the sequential predictions of autoregressive models, resulting in faster inference, enhanced training stability, and precise control for generating complex multi-dimensional data or steering model behaviors.}
  \item \textsf{\textbf{Atom 177}} \textsf{(14.67\%)}: \emph{Inference-time activation steering enables causal control over neural network behavior without weight fine-tuning by isolating high-level concepts—such as truthfulness, safety, or reasoning style—as specific directional vectors within the model's latent space. By dynamically intervening on a model's internal hidden states along these conceptual axes during the forward pass, researchers can robustly align outputs to desired traits, bypassing the vulnerabilities of external prompt manipulation while utilizing targeted calibrations to preserve the model's foundational capabilities and linguistic fluency.}
\end{enumerate}

\paragraph{Alien Sampler.}
\begin{enumerate}[leftmargin=*, itemsep=3pt, parsep=0pt]
  \item \textsf{\textbf{Atom 24}} \textsf{(6.67\%)}: \emph{The 'Reversal Curse' in autoregressive language models is a structural limitation arising from next-token prediction training, which optimizes strictly for forward conditional probabilities and encodes information as directional statistical dependencies rather than symmetric logical identities. Because gradient-based left-to-right optimization inherently stores knowledge as a one-way sequential association, language models fail to automatically infer inverse relationships or apply logic in reverse without explicit bidirectional training methods like permutation language modeling, exposing a fundamental gap between sequence-level pattern matching and true conceptual reasoning.}
  \item \textsf{\textbf{Atom 97}} \textsf{(6.00\%)}: \emph{The optimization of non-differentiable AI components---such as system prompts, reasoning chains, and multi-agent instructions---can be automated through 'Textual Gradients' or 'Symbolic Optimization.' By treating a Large Language Model as a meta-optimizer that analyzes execution errors and generates natural language critiques as directional feedback, AI systems can iteratively rewrite and evolve their own instructions to improve task performance and scale complexity, entirely bypassing the need for manual prompt engineering, human-annotated datasets, or access to internal numerical model weights.}
  \item \textsf{\textbf{Atom 63}} \textsf{(5.33\%)}: \emph{Advanced dataset curation for machine learning relies on multi-stage, hierarchical filtering pipelines that shift the focus from raw data volume to information density and semantic diversity. By utilizing high-capacity foundational models to distill qualitative metrics---such as pedagogical value, logical density, and factual accuracy---into lightweight proxy classifiers, researchers can autonomously evaluate and filter multi-billion-token corpora at scale. When combined with automated semantic deduplication and targeted expert-in-the-loop validation, this methodology isolates high-signal, diverse content, maximizing the computational efficiency of model training and preventing performance degradation caused by noisy, repetitive, or low-value data.}
\end{enumerate}
}

\subsection{Overlap of Atom Pairs and Complete Combinations}
\label{app:combination_overlap}

We compare the six 300-sample conditions in Appendix~\ref{app:sampler_setting}: Claude Opus 4.7 and Gemini 3.1 Pro under independent and batch prompting, Alien at $\beta=0.7$, and Uniform random. Each sample contains three distinct atoms from the same 273-atom vocabulary. Both batch calls use the same shuffled catalog order (seed 42).

\paragraph{Metrics.} For condition $A$, let $S_k(A)$ be the set of distinct unordered $k$-atom subsets occurring in its samples. Unlike the Jaccard similarity between individual ideas used to compute Vendi, this metric compares the sets of pairs or triples explored by different methods. We report the shared count $|S_k(A)\cap S_k(B)|$ and Jaccard overlap
\begin{equation}
    J_k(A,B)=\frac{|S_k(A)\cap S_k(B)|}{|S_k(A)\cup S_k(B)|},
    \qquad k\in\{2,3\}.
\end{equation}
Each three-atom sample contributes three pairs and one triple before deduplication; repeated pairs or triples count only once within each condition. Pair overlap includes pairs belonging to shared triples, so the two resolutions are related diagnostics. The number of distinct pairs equals 900 minus the repeated-pair count in Table~\ref{tab:generation-diversity}. Table~\ref{tab:combination-support} gives the resulting support sizes, and Table~\ref{tab:combination-overlap} reports all 15 between-condition comparisons.

\begin{table}[htbp]
    \centering
    \small
    \begin{tabular}{@{}lcc@{}}
        \toprule
        Method & Distinct pairs / 900 occurrences & Distinct triples / 300 samples \\
        \midrule
        Claude independent & 286 & 190 \\
        Gemini independent & 418 & 234 \\
        Claude batch & 656 & 277 \\
        Gemini batch & 586 & 294 \\
        Alien ($\beta=0.7$) & 794 & 300 \\
        Random & 888 & 300 \\
        \bottomrule
    \end{tabular}
    \caption{Distinct unordered atom pairs and triples in each 300-sample generation condition. Duplicates are retained in the sample budget but counted once in the overlap analysis.}
    \label{tab:combination-support}
\end{table}

\begin{table}[htbp]
    \centering
    \small
    \begin{tabular}{@{}lcc@{}}
        \toprule
        Comparison & Two-atom pairs & Three-atom combinations \\
        \midrule
        Claude independent $\leftrightarrow$ Gemini independent & 73 (11.57\%) & 20 (4.95\%) \\
        Claude batch $\leftrightarrow$ Gemini batch & 101 (8.85\%) & 9 (1.60\%) \\
        Claude independent $\leftrightarrow$ Claude batch & 44 (4.90\%) & 2 (0.43\%) \\
        Gemini independent $\leftrightarrow$ Gemini batch & 40 (4.15\%) & 1 (0.19\%) \\
        Claude independent $\leftrightarrow$ Gemini batch & 29 (3.44\%) & 2 (0.41\%) \\
        Gemini independent $\leftrightarrow$ Claude batch & 59 (5.81\%) & 4 (0.79\%) \\
        \midrule
        Claude independent $\leftrightarrow$ Alien & 10 (0.93\%) & 1 (0.20\%) \\
        Gemini independent $\leftrightarrow$ Alien & 20 (1.68\%) & 0 (0.00\%) \\
        Claude batch $\leftrightarrow$ Alien & 33 (2.33\%) & 0 (0.00\%) \\
        Gemini batch $\leftrightarrow$ Alien & 23 (1.69\%) & 0 (0.00\%) \\
        \midrule
        Claude independent $\leftrightarrow$ Random & 4 (0.34\%) & 0 (0.00\%) \\
        Gemini independent $\leftrightarrow$ Random & 11 (0.85\%) & 0 (0.00\%) \\
        Claude batch $\leftrightarrow$ Random & 21 (1.38\%) & 0 (0.00\%) \\
        Gemini batch $\leftrightarrow$ Random & 14 (0.96\%) & 0 (0.00\%) \\
        Alien $\leftrightarrow$ Random & 22 (1.33\%) & 0 (0.00\%) \\
        \bottomrule
    \end{tabular}
    \caption{Shared count (Jaccard percentage) for distinct unordered pairs and triples. Independent denotes one combination per call; Alien uses $\beta=0.7$. All conditions contain 300 samples.}
    \label{tab:combination-overlap}
\end{table}

\paragraph{Results and interpretation.} Claude--Gemini overlap is higher than LLM--Alien overlap at both resolutions under both independent and batch prompting. In the batch conditions, pair and triple Jaccards are approximately 23\% and 68\% lower than in the independent conditions. However, the number of shared pairs increases from 73 to 101 as each model explores more distinct pairs, so lower Jaccard should not be interpreted as fewer shared pairs. Alien's low overlap with LLM outputs extends to constituent pairs as well as complete triples.

Alien--Random overlap is also low: 22 shared pairs (1.33\% Jaccard) and no shared triples. This comparison makes clear that separation alone does not establish coherence, scientific originality, or cognitive unavailability. The overlap results describe shared concentration among LLM outputs; the scoring and downstream evaluations assess the usefulness of Alien's alternative exploration.

\subsection{Diversity--Quality Composite}
\label{app:composite}

The composite of Table~\ref{tab:generation-diversity} is $\tfrac12\,[z(\mathrm{Vendi})+z(Q)]$, where both terms are standardized across the six methods and $Q$ is, in turn, the human overall rating (the mean of coherence, feasibility, novelty, and promise), the win rate in the full round of the idea tournament, and the average rank percentile of the autoresearch reports. Batch quality is imputed from the same model's independent run. First, using human \emph{core} quality (coherence, feasibility, and promise, excluding novelty so that diversity is not counted twice), Alien scores 0.841, ahead of Claude batch (0.729), uniform random (0.337), Claude independent ($-0.296$), Gemini batch ($-0.497$), and Gemini independent ($-1.114$). Second, varying the diversity weight $w$ in $w\,z(\mathrm{Vendi})+(1-w)\,z(Q)$, Alien ranks first (tied with Claude at $w=0$ on human ratings) for every $w\le0.82$ with human ratings and every $w\le0.88$ with the two LLM-judged measures; only at larger weights does uniform random, the diversity ceiling, overtake it.

\subsection{Publication Age and Popularity of Selected Atoms}
\label{app:atom_age}

We ask whether Alien selects less widely used concepts simply because they are older. Here, an atom's \emph{popularity}, or crowding, is the number of corpus papers that express it. We measure age using the publication years of those papers in the 80\%-coverage representation used to train both models. These measures describe an atom's publication history within our corpus, rather than when the underlying concept was invented.

\paragraph{Measuring publication age.} An atom's \emph{first year} is the year of its earliest paper in the corpus. Its \emph{emergence year} is the 10th percentile of its papers' publication years, marking roughly when the first 10\% of those papers had appeared.

Because 49.3\% of the corpus was published in 2025, we also compare each atom's share of older papers with the corpus as a whole. Specifically, 21.8\% of all corpus papers were published before 2024. We define an \emph{age index} as
\[
    \text{age index}(a) =
    \frac{\text{fraction of atom } a\text{'s papers published before 2024}}{\text{fraction of all corpus papers published before 2024}}.
\]
An age index of 1 means that the proportion of pre-2024 papers associated with an atom matches that of the corpus overall. Values above 1 indicate a larger proportion of older papers; values below 1 indicate a smaller proportion. For example, an index of 0.96 means that the pre-2024 share is 4\% lower than the corpus share, not that the atom is 4\% younger in calendar years. Across individual atoms, the 10th and 90th percentiles of the index are 0.74 and 1.20.

\paragraph{Separating age from popularity.} At first, atoms with earlier first appearances seem to be more widely used: the Spearman correlation between first year and paper count is $-0.587$. However, an atom represented by more papers has more chances to include an early paper. To account for this, we randomly draw the same number of papers from the corpus for each atom and repeat this 200 times. Averaged across atoms, the observed and baseline first years differ by only 0.004 years, and the differences for individual atoms are uncorrelated with paper count ($\rho=0.012$). The age index also has only a weak correlation with paper count ($\rho=0.125$). Thus, within this corpus, being more widely used does not necessarily mean having an older publication profile.

\begin{table}[h]
    \centering
    \small
    \begin{tabular}{lccccc}
        \toprule
        Method & Age index & First year & Emergence & Papers/atom & Papers/top-10 atom \\
        \midrule
        Alien ($\beta=0.7$) & 0.96 & 2018.68 & 2022.55 & 131.2 & \textbf{108.6} \\
        Claude, independent & 1.01 & 2018.51 & 2022.49 & 141.2 & 114.6 \\
        Gemini, independent & 1.00 & 2018.48 & 2022.40 & 154.2 & 141.5 \\
        Claude, batch & 0.99 & 2018.04 & 2022.48 & 210.4 & 244.8 \\
        Gemini, batch & 0.95 & 2018.22 & 2022.56 & 210.2 & 128.0 \\
        Uniform random & 0.98 & 2018.26 & 2022.50 & 197.6 & 174.4 \\
        \bottomrule
    \end{tabular}
    \caption{Publication age and popularity of selected atoms. Values are averaged over the 900 atom occurrences in each method's 300 combinations, so repeatedly selected atoms count repeatedly. The final column averages the number of papers containing each of the method's ten most-selected atoms. Across the full vocabulary, the average is 198.2 papers per atom.}
    \label{tab:atom-age}
\end{table}

\paragraph{Comparison across methods.} Table~\ref{tab:atom-age} averages the measures over all 300 three-atom combinations from each method. The methods' mean age indices range from 0.95 to 1.01, all within approximately 5\% of the corpus reference value of 1. This comparison concerns each method's average; individual selected atoms can have older or more recent publication profiles. Mean first-appearance years range from 2018.04 to 2018.68, and mean emergence years range from 2022.40 to 2022.56.

The small differences between methods can still be statistically detectable. Alien's mean age index is lower than those of the independent baselines ($-0.049$ relative to Claude and $-0.042$ relative to Gemini; two-sided permutation tests over combinations with 100,000 relabellings, $p<0.001$), indicating a slightly smaller share of older papers. It does not differ detectably from Gemini batch ($p=0.735$).

The clearer distinction is in popularity: Alien's ten most-selected atoms appear in 108.6 papers on average, fewer than for any other method and below the vocabulary average of 198.2. Alien therefore concentrates on less widely used concepts without showing a general preference for older ones.

\subsection{Pairwise LLM Judge Evaluation of Generated Ideas}
\label{app:idea_pairwise_judge}

The diversity and novelty metrics above measure whether a method explores a broad, less familiar region of idea space. We also ran a forced-choice LLM evaluation of coherent originality over decoded ideas. Each comparison showed two methodology sketches without method labels, and the judge selected the sketch that better satisfied the coherent novelty criterion: plausible enough to investigate, but less likely to be proposed by the research community. The prompt explicitly asked whether the idea opens a new framing, research object, empirical target, or line of follow-on work rather than merely specifying a convenient first experiment.

We used GPT-5.5 with medium reasoning effort, temperature 1.0, and JSON-mode forced-choice outputs. The tournament design was deliberately simple. In the full round, we used 50 ideas per method. For each unordered method pair and each of three seeds, we sampled 20 one-to-one cross-method comparisons, avoiding duplicate unordered idea pairs where possible and balancing A/B presentation sides. With four methods this gives $6 \times 3 \times 20 = 360$ pairwise judgments, all usable. We then ran two elimination diagnostics: select the top half of each method by item-level wins in the previous round and repeat the same balanced design with 25 ideas per method and 10 comparisons per method pair per seed (180 judgments), then select the top 10 per method and repeat with 4 comparisons per method pair per seed (72 judgments).

\begin{table}[h]
    \centering
    \small
    \setlength{\tabcolsep}{7pt}
    \begin{tabular}{lccc}
        \toprule
        Method & Full 50 & Top 25 & Top 10 \\
        \midrule
        Random & 41.1\% & 41.1\% & 41.7\% \\
        Alien sampler ($\beta=0.7$) & 55.6\% & 72.2\% & 69.4\% \\
        Gemini 3.1 Pro & 54.4\% & 58.9\% & 58.3\% \\
        Claude Opus 4.7 & 48.9\% & 27.8\% & 30.6\% \\
        \bottomrule
    \end{tabular}
    \caption{Raw method win rates in the coherent-unavailability pairwise judge. Each entry is the fraction of all matches involving that method that the method won in the corresponding round.}
    \label{tab:idea-judge-winrates}
\end{table}

\begin{table}[h]
    \centering
    \small
    \setlength{\tabcolsep}{7pt}
    \begin{tabular}{lccc}
        \toprule
        Method & Full 50 & Top 25 & Top 10 \\
        \midrule
        Random & 0.763 & 0.749 & 0.766 \\
        Alien sampler ($\beta=0.7$) & 1.184 & 2.109 & 1.888 \\
        Gemini 3.1 Pro & 1.144 & 1.336 & 1.305 \\
        Claude Opus 4.7 & 0.967 & 0.474 & 0.530 \\
        \bottomrule
    \end{tabular}
    \caption{Bradley--Terry method abilities fit separately within each judge round and normalized so that the method mean is 1. Scores above 1 indicate above-average pairwise preference under the coherent-unavailability judge for that round.}
    \label{tab:idea-judge-bt}
\end{table}

\begin{figure}[h]
    \centering
    \includegraphics[width=0.78\textwidth]{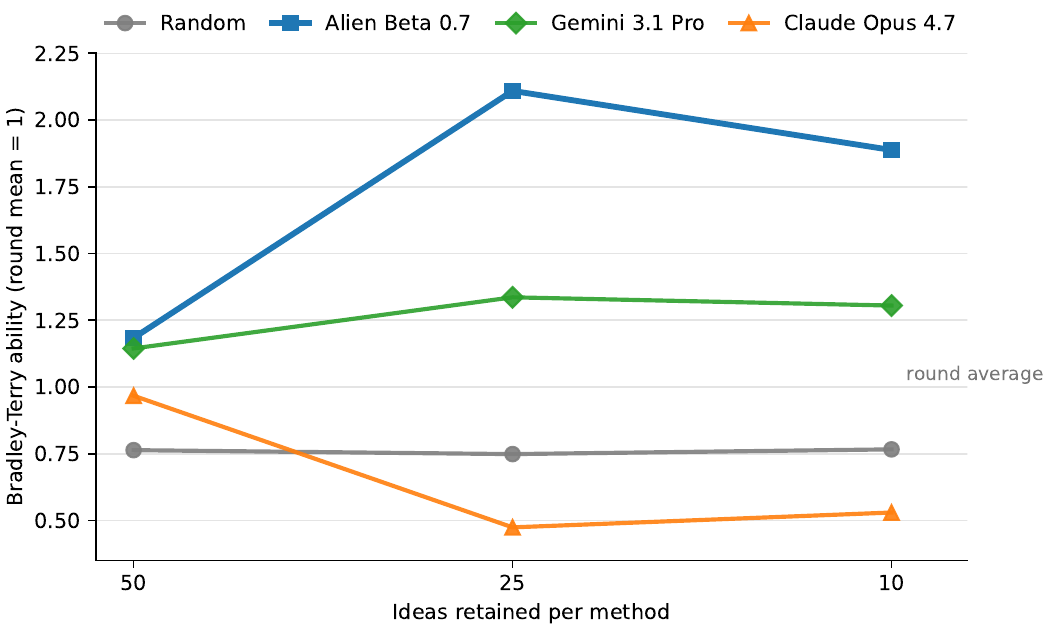}
    \caption{Bradley--Terry method abilities across elimination rounds. Alien strengthens sharply after selecting the top half of each method's outputs, while the random baseline remains flat and Claude drops under the coherent-unavailability criterion.}
    \label{fig:llm-judge-elimination-bt}
\end{figure}

\begin{figure}[h]
    \centering
    \includegraphics[width=\textwidth]{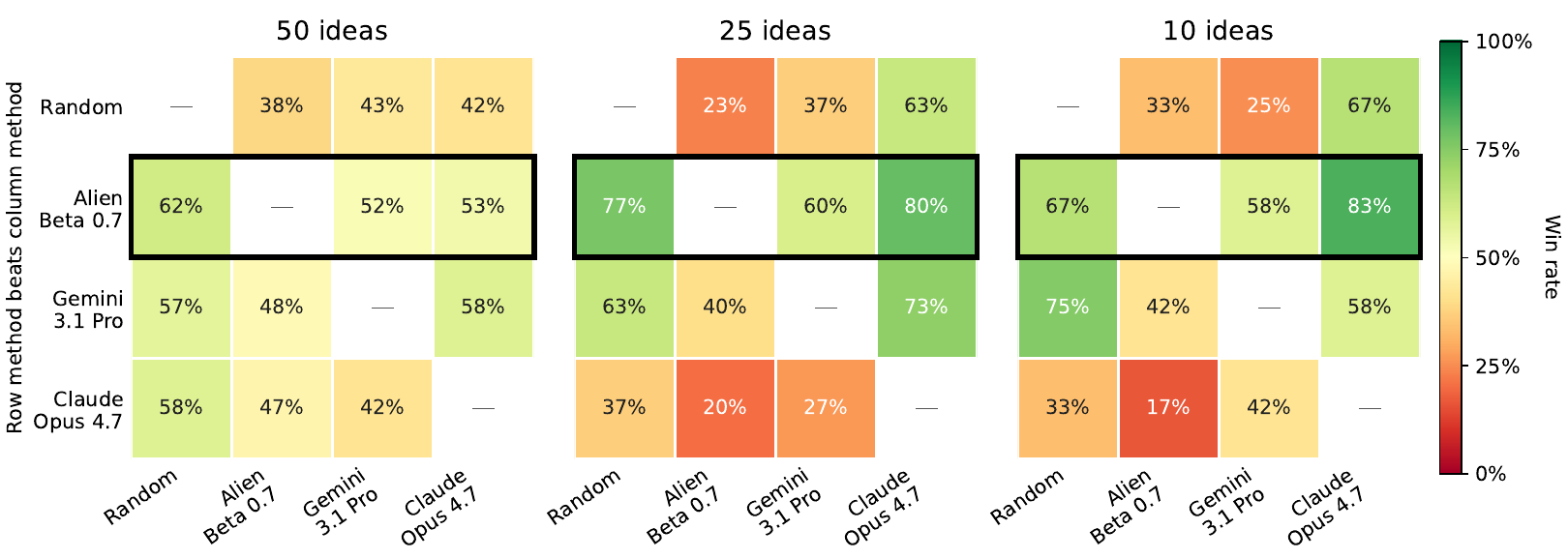}
    \caption{Pairwise win-rate matrices for the same judge rounds. Each cell is the row method's win rate against the column method; the diagonal is omitted. The black outline highlights the Alien sampler row.}
    \label{fig:llm-judge-winrate-heatmaps}
\end{figure}

\clearpage

\section{Downstream Evaluation Details}
\label{app:downstream_eval}

We ran two downstream evaluations beyond atom-level diversity, novelty, and coherence, on the same 10 decoded ideas per method. First, human raters evaluated the idea descriptions directly. Second, we ran a fixed autoresearch pipeline that attempted to turn each idea into a bounded empirical research note, then evaluated the resulting notes with blind pairwise LLM judgments, a judge that we validated against 50 human experts (Appendix~\ref{app:judge_validation}). These evaluations test a different question from the atom metrics: whether generated ideas remain useful once interpreted as concrete research directions.

\subsection{Human Survey}
\label{app:human_survey}

Ten paid raters with post-graduate degrees in machine learning or computer science each reviewed four ideas, one per source method, giving 40 blinded reviews over the 10 ideas per method that were also run through autoresearch. Reviewers rated each idea on five-point scales for coherence, feasibility, novelty, promise (``how worthwhile would this be for a researcher or lab to pursue?''), and obviousness. We report \emph{overall} quality as the mean of coherence, feasibility, novelty, and promise, and \emph{core} quality as the mean of coherence, feasibility, and promise. Table~\ref{tab:human-survey-results} reports means and standard errors. The results are close: Alien and Claude tie on overall quality (4.025 each), Alien has the highest feasibility and novelty means, and Claude has the highest coherence, promise, and core means (4.100 vs.\ 4.067 for Alien).

\begin{table}[h]
    \centering
    \small
    \setlength{\tabcolsep}{4pt}
    \begin{tabular}{lcccccc}
        \toprule
        Method & $n$ & Coherence & Feasibility & Novelty & Promise & Overall \\
        \midrule
        Alien 
        & 10 
        & $4.20 \pm 0.20$ 
        & $\mathbf{3.90 \pm 0.23}$ 
        & $\mathbf{3.90 \pm 0.28}$ 
        & $4.10 \pm 0.18$ 
        & $4.03 \pm 0.11$ \\
        
        Claude 
        & 10 
        & $\mathbf{4.30 \pm 0.30}$ 
        & $3.60 \pm 0.22$ 
        & $3.80 \pm 0.29$ 
        & $\mathbf{4.40 \pm 0.27}$ 
        & $4.03 \pm 0.23$ \\
        
        Gemini 
        & 10 
        & $4.00 \pm 0.33$ 
        & $3.20 \pm 0.29$ 
        & $3.70 \pm 0.37$ 
        & $3.80 \pm 0.39$ 
        & $3.68 \pm 0.30$ \\
        
        Random 
        & 10 
        & $3.70 \pm 0.21$ 
        & $3.70 \pm 0.21$ 
        & $3.60 \pm 0.22$ 
        & $4.00 \pm 0.15$ 
        & $3.75 \pm 0.04$ \\
        
        \bottomrule
    \end{tabular}
    \caption{Human ratings of generated idea descriptions. Entries are means $\pm$ standard errors over clean reviews.}
    \label{tab:human-survey-results}
\end{table}

\begin{figure}[h]
    \centering
    \includegraphics[width=0.6\linewidth]{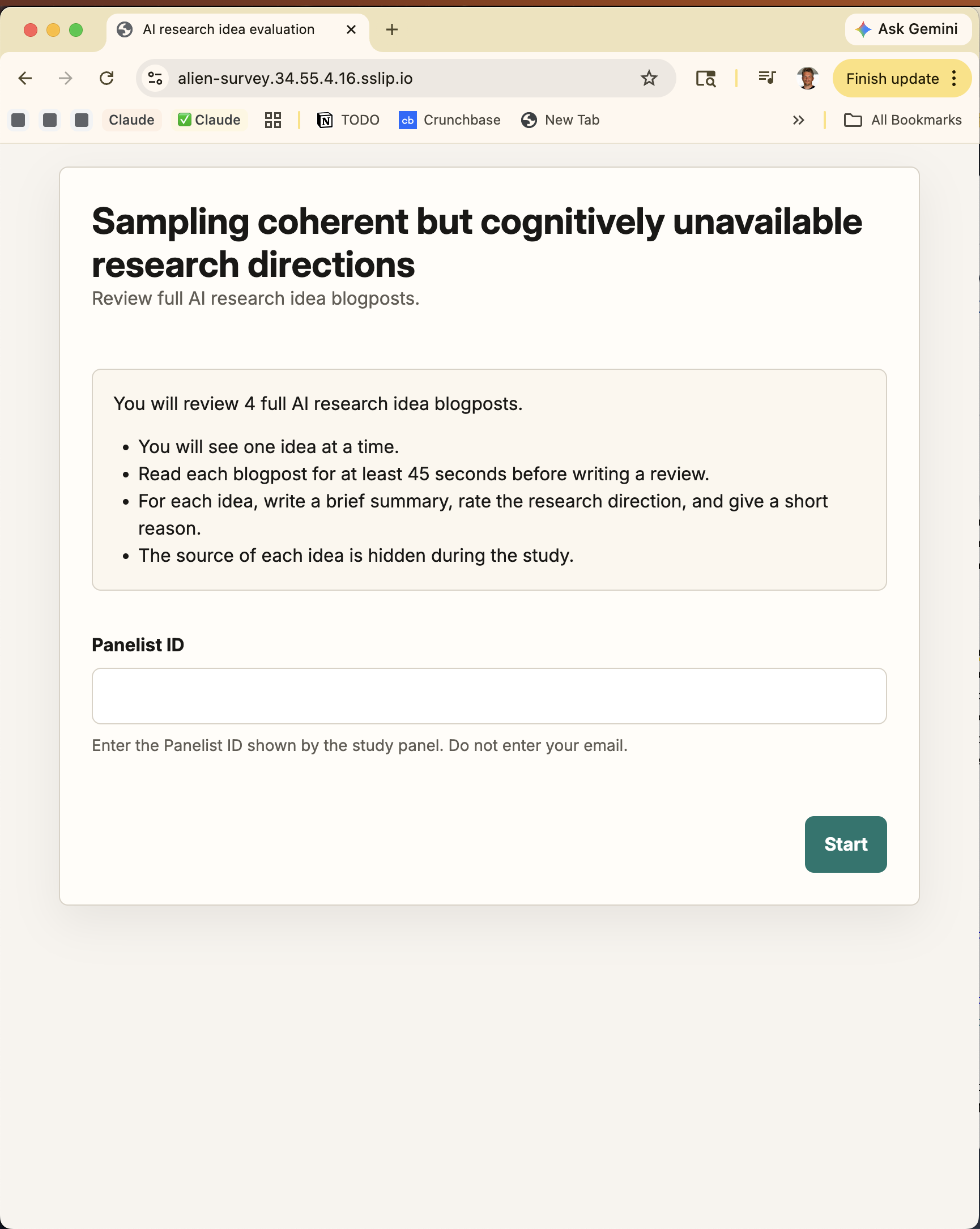}
    \caption{Human survey landing page. Reviewers entered a study-issued panelist ID and read brief instructions before starting. Each session covered four idea blogposts, presented one at a time, with the source method of each idea hidden.}
    \label{fig:human_eval_screenshot_0}
\end{figure}

\begin{figure}[h]
    \centering
    \includegraphics[width=0.6\linewidth]{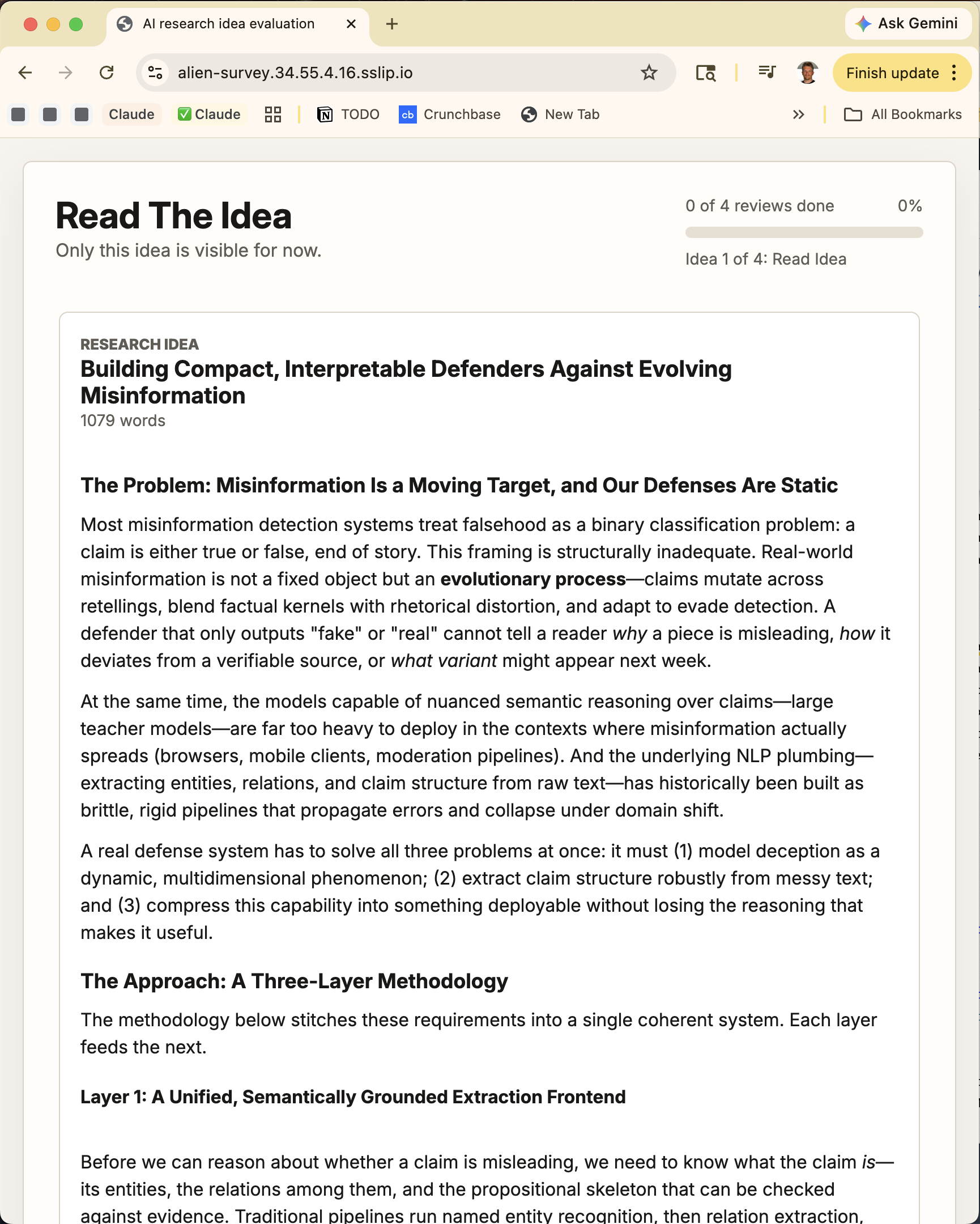}
    \caption{Idea reading view. Each blogpost was rendered in full (with the title and word count shown above the body) and only the current idea was visible at any time. A progress indicator on the top right tracked completion across the four ideas.}
    \label{fig:human_eval_screenshot_1}
\end{figure}

\begin{figure}[h]
    \centering
    \includegraphics[width=0.85\linewidth]{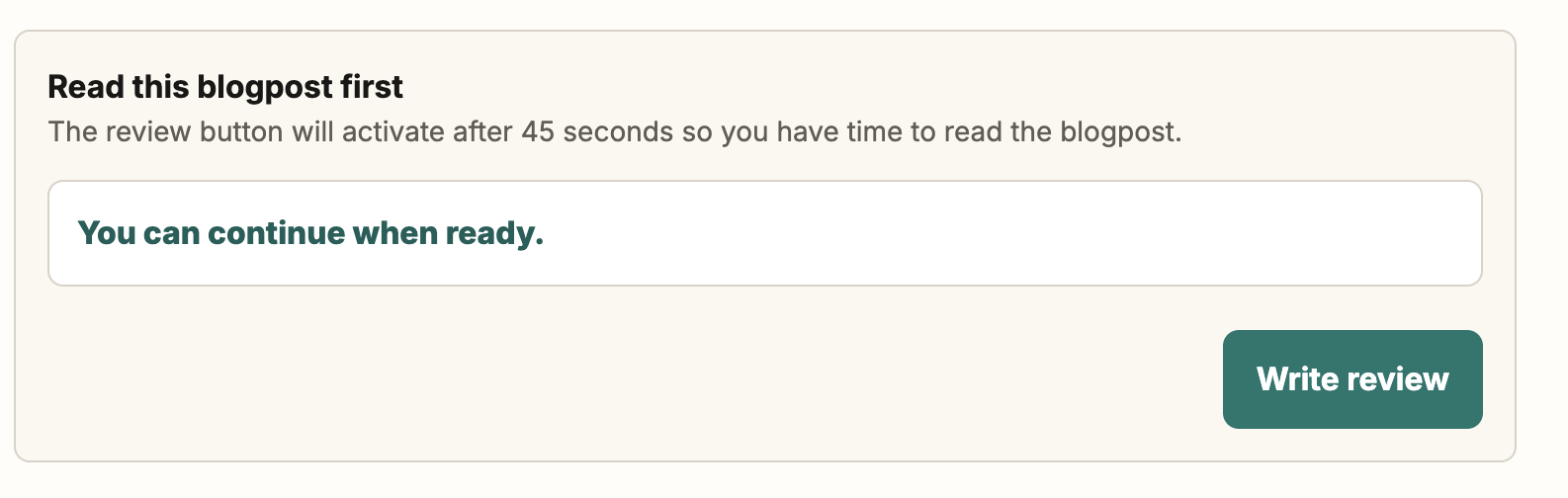}
    \caption{Read gate. The ``Write review'' button was disabled for the first 45 seconds on each idea, enforcing a minimum reading window before any rating could be entered.}
    \label{fig:human_eval_screenshot_2}
\end{figure}

\begin{figure}[h]
    \centering
    \includegraphics[width=0.6\linewidth]{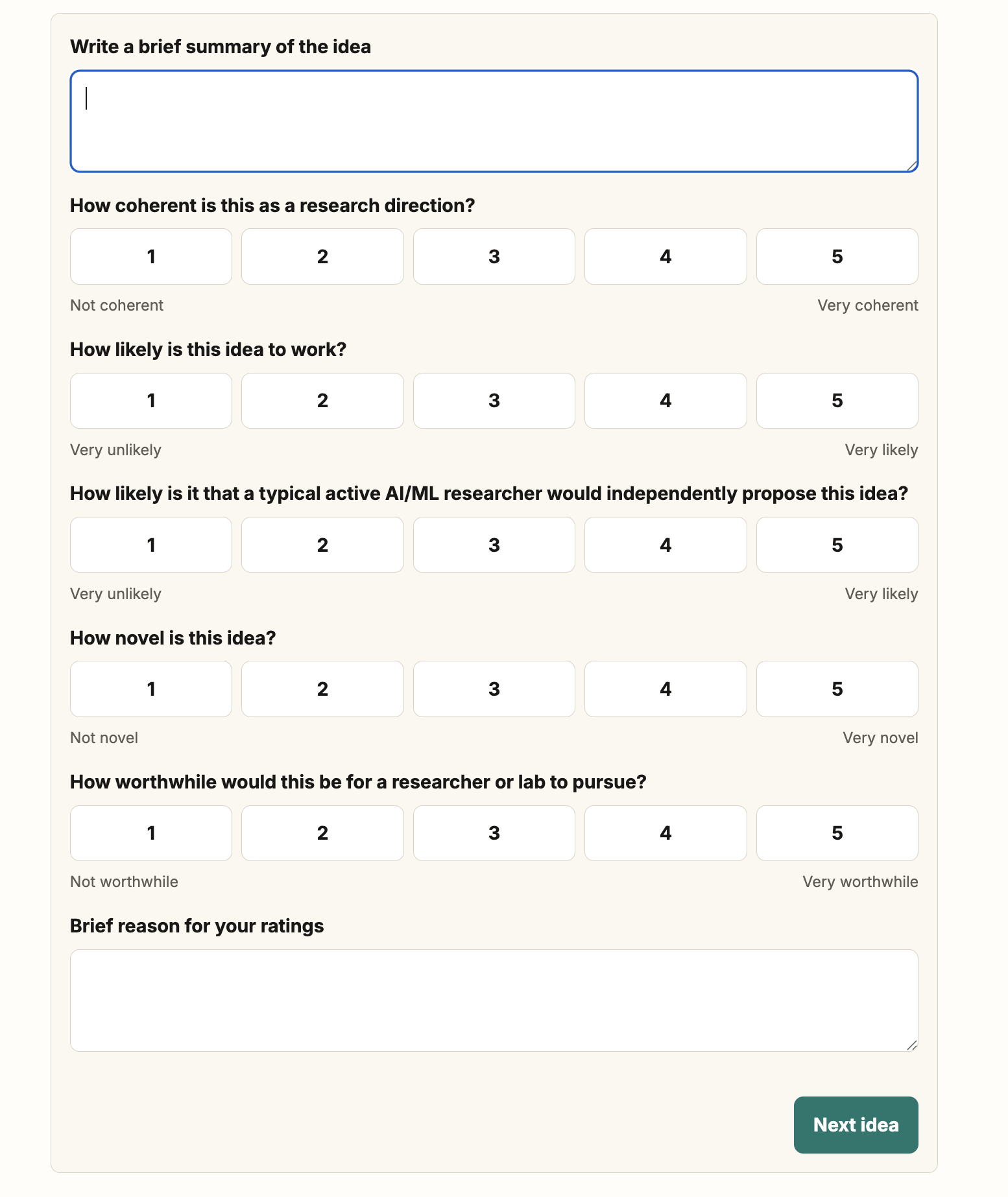}
    \caption{Rating form. For each idea, reviewers wrote a brief free-text summary, rated five ordinal dimensions on five-point scales---coherence (``How coherent is this as a research direction?''), feasibility (``How likely is this idea to work?''), obviousness (``How likely is it that a typical active AI/ML researcher would independently propose this idea?''), novelty (``How novel is this idea?''), and worthwhileness (``How worthwhile would this be for a researcher or lab to pursue?'')---and provided a short free-text justification.}
    \label{fig:human_eval_screenshot_3}
\end{figure}

\paragraph{Survey instrument.}
The survey was self-administered through a private web app via per-rater unique URLs; no identifying information was collected beyond a study-issued panelist ID. After a short instructions screen (Figure~\ref{fig:human_eval_screenshot_0}), each rater saw four idea blogposts one at a time (Figure~\ref{fig:human_eval_screenshot_1}); the source method of each idea was hidden. A read gate (Figure~\ref{fig:human_eval_screenshot_2}) disabled the rating button for the first 45 seconds of each idea to enforce a minimum reading window. The rating form (Figure~\ref{fig:human_eval_screenshot_3}) then collected a brief free-text summary, five ordinal ratings on five-point scales (coherence, feasibility, obviousness, novelty, and worthwhileness), and a short free-text justification.

\subsection{Autoresearch Pipeline and Pairwise Report Judging}
\label{app:autoresearch_eval}

For each source method, we selected 10 generated ideas and ran the same autoresearch pipeline. The pipeline used Claude Code CLI v2.1.101 with the \texttt{claude-sonnet-4-6} model setting on Mila compute nodes with optional A100 40GB GPU access. Each run consisted of a deep-research stage, an experimenter stage, a writer stage, a critic stage, and one experimenter--writer repair loop. The final artifact was a standalone \texttt{research\_note.md}.

Of the 40 attempted ideas, 38 produced reports; two Random runs failed because of unfeasible experimentation. The 38 completed reports were anonymized and evaluated with blind pairwise preference judgments by GPT-5.5. We used a six-round Swiss-style comparison schedule: reports were paired against nearby reports in the current ordering, duplicate unordered pairings were avoided where possible, and each scheduled comparison was shown in both A/B and B/A order. Pairwise preferences were converted into ranks using Bradley--Terry scores. Completed reports received $(40-r)/39$ for rank $r$; failed runs received zero. Method means include all 10 attempts, and failures are assigned rank 40 for the median.

\begin{table}[h]
    \centering
    \small
    \setlength{\tabcolsep}{5pt}
    \begin{tabular}{lccccc}
        \toprule
        Method & Avg. rank percentile & Reports & Top 5 & Top 10 & Median rank \\
        \midrule
        Alien & 0.633 & 10 & 3 & 4 & 14.5 \\
        Claude & 0.574 & 10 & 1 & 4 & 14.0 \\
        Gemini & 0.495 & 10 & 1 & 2 & 24.0 \\
        Random & 0.295 & 8 & 0 & 0 & 28.0 \\
        \bottomrule
    \end{tabular}
    \caption{LLM-judge rankings of completed reports. Each method attempted 10 ideas; mean scores and median ranks include the two Random failures as described above.}
    \label{tab:autoresearch-rank-results}
\end{table}

\subsection{Validating the LLM Judge Against Human Experts}
\label{app:judge_validation}

Most quality evidence in this paper comes from LLM judges, so we tested whether the judge tracks expert opinion in the setting where LLM mediation is most pronounced: comparing autoresearch reports, where both the research pipeline and the judge are LLMs.

\paragraph{Participants.} We recruited 50 paid participants through an online research platform, screened for a graduate degree or doctorate and either verified computer-science expertise or a master's or doctoral specialization in AI, ML, or NLP.

\paragraph{Procedure.} We drew 10 pairs of reports from the 38 completed autoresearch reports of Appendix~\ref{app:autoresearch_eval}, the same reports compared by the GPT-5.5 judge. Each participant compared one pair under the same criteria given to the judge, with the source methods hidden and the presentation order randomized; each pair was judged by five participants, giving 50 human judgments. The survey was self-administered through a private web app via per-rater
unique URLs; no identifying information was collected beyond a study-issued panelist ID. The general evaluation criteria were shared with the first study (Appendix~\ref{app:human_survey}), adapted to report-pair comparison.

\paragraph{Analysis and results.} Judge--expert agreement is the share of individual human judgments that match the judge's verdict on the same pair: 68\% (95\% CI 54--80\%, bootstrapped over pairs). Expert--expert agreement, averaged over all pairs of participants who judged the same report pair, is 58\%. The judge therefore agrees with an individual expert more often than two experts agree with each other, and it matches the majority verdict of the five experts on 8 of the 10 pairs. Report comparisons are hard even for experts, and the judge is not an outlier relative to them; if anything, it tracks the expert consensus more closely than a typical individual rater.

\subsection{Motif Concentration in Autoresearch Reports}
\label{app:motif_concentration}

We coded each original idea and each final report for membership in a sparse-autoencoder / feature-level mechanistic-interpretability cluster. A report counts once if it uses SAE or sparse-feature language, or feature-level interpretability machinery such as activation steering, causal tracing, activation patching, concept vectors, monosemantic features, or polysemantic representations. Generic uses of words such as ``mechanistic'' or ``interpretable'' were not counted.

\begin{table}[h]
    \centering
    \small
    \begin{tabular}{lcc}
        \toprule
        Method & Original ideas in cluster & Final reports in cluster \\
        \midrule
        Alien & 0/10 & 0/10 \\
        Claude & 2/10 & 8/10 \\
        Gemini & 4/10 & 5/10 \\
        Random & 0/10 & 1/8 \\
        \bottomrule
    \end{tabular}
    \caption{Concentration around the SAE / feature-level mechanistic-interpretability cluster before and after autoresearch. Claude and Gemini reports are much more likely to route through this toolkit than Alien reports.}
    \label{tab:motif-concentration}
\end{table}

\subsection{Idea Identity Preservation}
\label{app:idea_identity_preservation}

As a sanity check, we asked whether the autoresearch pipeline preserves the identity of the starting idea. For each completed final report, we embedded the final report and all 40 original ideas with \texttt{text-embedding-3-small}. We then ranked all original ideas by cosine distance to the final report and recorded the rank of the report's own starting idea. If the pipeline preserves idea identity, the final report should be nearest, or at least very close, to its own original title.

\begin{table}[h]
    \centering
    \small
    \begin{tabular}{lcccc}
        \toprule
        Method & Final reports & Own nearest & Own in top 3 & Mean own rank \\
        \midrule
        Alien & 10 & 7 & 8 & 2.40 \\
        Claude & 10 & 7 & 9 & 2.30 \\
        Gemini & 10 & 8 & 9 & 2.60 \\
        Random & 8 & 3 & 6 & 2.88 \\
        \bottomrule
    \end{tabular}
    \caption{Title-level identity preservation through the autoresearch pipeline. Most final reports remain closest, or close, to their own starting idea title.}
    \label{tab:idea-identity-preservation}
\end{table}

The identity-preservation result suggests that the autoresearch pipeline is not simply rewriting all ideas into unrelated generic reports. However, final reports are still more stylistically and semantically similar to each other than the initial ideas, as expected from a shared empirical-report format and bounded experimentation pipeline. We therefore treat idea-to-report nearest-neighbor preservation as a sanity check, and treat motif concentration as the more interpretable downstream diversity diagnostic.

\section{Example Outputs}
\label{app:examples}

\subsection{Example Atoms}
\label{app:example_atoms}

Below we show example atoms along with representative constituent conceptual units. The \textbf{Atom} field contains the canonical description passed to the LLM during reconstruction, the general idea captured by the atom. The \textbf{LLM justification} explains what the clustered units have in common, the analysis the LLM does on the conceptual units before naming the atom.

\begin{tcolorbox}[
  colback=gray!5,
  colframe=gray!50,
  boxrule=0.5pt,
  arc=2pt,
  left=8pt,
  right=8pt,
  top=6pt,
  bottom=6pt,
  title={\textbf{Atom Example: Sparse Autoencoders}},
  fonttitle=\small,
  breakable
]

\textbf{Atom 111:} Sparse Autoencoders (SAEs) resolve the problem of neural superposition—where individual neurons simultaneously encode multiple, overlapping concepts—by functioning as dictionary learners that project dense, polysemantic model activations into a higher-dimensional latent space constrained by sparsity. This mechanism decomposes complex internal representations into discrete, 'monosemantic' features, where each active dimension corresponds to a single, human-interpretable concept. By isolating these individual computational primitives, SAEs enable researchers to transparently audit a neural network's internal logic, map causal feature circuits, and perform precise, surgical interventions on specific behaviors without the need for opaque fine-tuning.

\vspace{6pt}
\textbf{LLM justification:} \textit{The cluster uniformly focuses on the application of Sparse Autoencoders (SAEs) and dictionary learning in mechanistic interpretability. The common thread is the mechanism of projecting dense, overlapping neural activations into a higher-dimensional, sparse space to resolve 'polysemanticity' (superposition). This effectively decomposes messy signals into distinct, single-concept ('monosemantic') features, enabling researchers to transparently audit internal logic, map causal circuits, and surgically steer model behavior.}

\vspace{8pt}
\textbf{Constituent Conceptual Units:}
\begin{enumerate}[leftmargin=*, itemsep=2pt, parsep=0pt]
  \item Sparse Autoencoders (SAEs) can disentangle conceptual 'superposition' in neural networks—where single neurons represent multiple unrelated concepts—by projecting hidden activations into a higher-dimensional space where individual features map to specific, interpretable semantic concepts.
  \item Sparse Autoencoders (SAEs) can act as 'interpretable dictionaries' for opaque models by mapping dense, entangled internal representations into a high-dimensional, sparse latent space where each active dimension corresponds to a single, monosemantic concept.
  \item Training objectives for Sparse Autoencoders balance a reconstruction error (ensuring the sparse code preserves the original data's information) with an L1 regularization penalty that forces the model to represent complex inputs using the smallest possible set of active concepts.
  \item Sparse Autoencoders can interpret opaque neural network activations by projecting them into a high-dimensional, sparse hidden layer; this decomposes complex internal signals into a set of nearly independent 'feature directions' that correspond to human-understandable concepts like 'baseball' or 'ships'.
  \item Feature discovery in neural networks can be categorized into three primary approaches: unsupervised methods (like Sparse Autoencoders) that find hidden structures without labels, supervised subspace methods (like Distributed Alignment Search) that learn rotations of neural space to find causal directions for specific labels, and masking methods (like Differential Binary Masking) that identify discrete subsets of existing neurons responsible for a concept.
  \item Sparse Autoencoders (SAEs) facilitate model interpretability and control by decomposing dense, polysemantic neural activations into a high-dimensional space of sparse, individual features, allowing researchers to 'clamp' specific features to influence model output.
  \item Sparse Autoencoders (SAEs) address the problem of neural superposition—where a single activation represents multiple, entangled concepts—by decomposing dense hidden states into thousands of sparse, monosemantic features that each correspond to a single, human-understandable concept.
  \item Mechanistic interpretability tools like Sparse Autoencoders offer an alternative to prompt engineering by identifying specific neural features that correlate with task performance, enabling researchers to directly amplify these internal representations to steer the model toward desired outputs more precisely.
  \item Sparse Autoencoders (SAEs) can be used to decompose the polysemantic, high-dimensional activation space of Large Language Models into a set of sparse, interpretable linear directions called 'latents,' which isolate specific conceptual features that would otherwise be overlapping and messy.
  \item Neural superposition occurs in Large Language Models when single neurons represent multiple unrelated concepts simultaneously to increase efficiency; Sparse Autoencoders (SAEs) can mitigate this by mapping messy activations into a high-dimensional space where individual 'features' represent specific, human-understandable concepts.
  \item \ldots{} (88 additional conceptual units omitted)
\end{enumerate}
\end{tcolorbox}

\begin{tcolorbox}[
  colback=gray!5,
  colframe=gray!50,
  boxrule=0.5pt,
  arc=2pt,
  left=8pt,
  right=8pt,
  top=6pt,
  bottom=6pt,
  title={\textbf{Atom Example: Code Structure Graphs}},
  fonttitle=\small,
  breakable
]

\textbf{Atom 253:} Transforming source code from linear text into explicit structural and semantic representations—such as Abstract Syntax Trees, control-flow graphs, and project-wide dependency networks—decouples underlying algorithmic logic from surface-level syntax. This architectural abstraction enables language models and automated systems to transcend token-based context limitations by navigating software as an interconnected topology, allowing them to systematically resolve non-local dependencies, extract minimal viable execution contexts, enforce deterministic semantic rules, and perform complex reasoning tasks across large-scale, multi-file repositories.

\vspace{6pt}
\textbf{LLM justification:} \textit{The cluster revolves around the limitation of treating source code as flat text and the necessity of using formal graph-based abstractions (ASTs, Control Flow Graphs, Dependency Trees, etc.) to capture programming logic. By mapping codebase architecture, these structural tools constrain and guide language models, allowing them to manage context windows, resolve distant dependencies, and reason accurately about project-wide semantics.}

\vspace{8pt}
\textbf{Constituent Conceptual Units:}
\begin{enumerate}[leftmargin=*, itemsep=2pt, parsep=0pt]
  \item Structural filtering using Abstract Syntax Tree (AST) analysis—which maps code into a hierarchical representation of its logical operations—allows training systems to identify functional equivalence between snippets, ensuring that code with identical execution logic is treated as similar even if its surface-level text or naming differs.
  \item Professional software development is characterized by high interdependency, where research indicates that roughly 73\% of functions in large-scale repositories rely on project-specific modules, internal utilities, or local architectural patterns rather than being self-contained.
  \item A distinction exists between 'static' code features (properties visible without execution, like token count) and 'dynamic' code features (properties that emerge during runtime, like the number of execution steps), which are processed by different cognitive subsystems during mental simulation.
  \item Complex software repositories can be converted into portable, executable units by tracing the dependency tree of a specific target file—analyzing imports and internal file paths—to extract only the essential code and data needed for that specific functionality, creating a 'compact workspace.'
  \item Software library evolution can be modeled as a Knowledge Graph where API elements are nodes and version transitions (such as renames or parameter updates) are directed edges; this enables the calculation of a cumulative 'net change' between non-adjacent versions by traversing and merging the transformation rules across multiple version jumps.
  \item Variable definitions located in remote directories can be recovered for code completion context by traversing a repository-level dataflow graph from the point of use, following logical dependencies back to their source assignments and type declarations regardless of file structure.
  \item Source code foundation models can be utilized as parametric knowledge bases where, instead of retrieving existing code from an external database, the model is 'probed' to synthesize new code fragments that explain the logic of low-level machine instructions.
  \item A 'Removal-then-Completion' strategy for code generation addresses potential errors in a user's draft by stripping away partial implementations and generating code based solely on high-level specifications (like docstrings) to ensure a clean, functional foundation.
  \item Performance bottlenecks in complex software can be identified by applying PageRank-style ranking to program call graphs, where execution costs are propagated from callees to callers to rank functions based on their cumulative influence on overall system performance.
  \item Software 'domain' can be categorized into a hierarchy of granularity—including organization, project/repository, and module/folder—meaning that a distributional shift in code models can occur at various levels of stylistic and architectural context.
  \item \ldots{} (90 additional conceptual units omitted)
\end{enumerate}
\end{tcolorbox}

\subsection{Example Research Ideas}
\label{app:example_ideas}

Below we show example research ideas generated by the Alien Sampler and by Claude Opus 4.7. Each example shows the input atoms and the reconstructed natural language research idea produced by the reconstruction pipeline.

\begin{tcolorbox}[
  colback=gray!5,
  colframe=gray!50,
  boxrule=0.5pt,
  arc=2pt,
  left=8pt,
  right=8pt,
  top=6pt,
  bottom=6pt,
  title={\textbf{Alien Sampler -- Example Idea 1}},
  fonttitle=\small,
  breakable
]

\textbf{Atoms:}
\begin{itemize}[leftmargin=*, itemsep=2pt, parsep=0pt]
  \item Formal grammar frameworks---such as Context-Free Grammars and their probabilistic extensions---serve as deterministic bridges between abstract mathematical logic and complex natural language. By enforcing explicit, hierarchical rules for syntax generation and parsing, these frameworks allow researchers to exert rigorous control over structural complexity. This structural determinism is utilized to generate unbiased, procedurally complex synthetic datasets, strictly constrain neural agent outputs to valid logical action spaces, and establish absolute mathematical ground truths. Consequently, researchers can isolate and evaluate a machine learning model's underlying syntactic reasoning, spatial logic, and predictive uncertainty entirely independently from real-world semantic confounders or benchmark memorization.
  \item To overcome the 'plausibility trap'---where artificial intelligence explanations sound convincing but fail to represent the system's actual logic---the structural faithfulness of a model's interpretability is quantified through causal perturbation and predictive reconstruction frameworks. By employing mechanisms such as necessity and sufficiency testing (ablating or isolating key features to observe proportional shifts in model confidence) alongside simulatability tasks (where independent proxy models attempt to reproduce the original decision using solely the generated explanation), researchers can objectively verify that an interpretable narrative maps directly to the true parametric mechanisms and causal pathways driving the system's output.
  \item Iterative self-improvement loops enable machine learning models to autonomously bootstrap their capabilities by functioning as both data generators and evaluators. By producing synthetic outputs, filtering them through automated verification mechanisms---such as self-consistency checks, environmental execution, or adversarial self-critique---and fine-tuning on the resulting high-quality reasoning trajectories or isolated error cases, models establish a recursive training flywheel. This continuous feedback cycle progressively shifts the learning frontier, allowing models to transcend the limitations of their initial training distributions, master increasingly complex tasks, and significantly reduce reliance on human-annotated labels.
\end{itemize}

\vspace{6pt}
\textbf{Research Idea:}

{\bfseries Grammar-Grounded Self-Improvement with Faithful Interpretability}\vspace{4pt}

\vspace{6pt}\noindent\textbf{The Problem: Reasoning Models We Can Neither Trust Nor Verify}\vspace{2pt}

Modern machine learning models, especially those trained on web-scale corpora, suffer from three interlocking failure modes that prevent them from being reliable reasoners.

First, \textbf{evaluation is contaminated}. When we test reasoning on natural-language benchmarks, we cannot disentangle genuine inference from memorization, semantic shortcut-taking, or distributional priors baked in during pretraining.

Second, \textbf{explanations lie}. A model can produce a fluent chain-of-thought rationale that bears no causal relationship to the parameters that actually produced its answer. This "plausibility trap" makes interpretability tools dangerous when used as trust signals: they sound right, but verify nothing.

Third, \textbf{improvement is bottlenecked by human labels}. Scaling reasoning capability by hand-annotating ever-harder problems is expensive, slow, and ultimately capped by what annotators themselves can solve.

This methodology proposes a unified research approach that addresses all three problems simultaneously by combining \textbf{formal grammars as a structural backbone}, \textbf{causal interpretability tests for faithfulness}, and \textbf{self-improvement loops for capability expansion}.

\vspace{6pt}\noindent\textbf{The Approach: A Three-Layer Methodology}\vspace{2pt}

The core idea is to build the entire training and evaluation pipeline on top of a \textbf{formal grammar substrate}, then use that substrate to (a) guarantee ground truth, (b) verify the faithfulness of model explanations, and (c) drive iterative self-improvement on synthetic but rigorously specified tasks.

\vspace{6pt}\noindent\textbf{Layer 1: Formal Grammars as the Ground-Truth Substrate}\vspace{2pt}

We start by adopting context-free grammars (and their probabilistic extensions) as the generative engine for the task universe. A grammar is a deterministic bridge between abstract mathematical logic and natural-language-like structure: it specifies, hierarchically and explicitly, what counts as a well-formed input and what counts as a valid derivation.

This choice is consequential:

\begin{itemize}[leftmargin=*, itemsep=2pt, parsep=0pt]
\item \textbf{Procedurally generated, unbiased datasets.} Because we control the grammar, we can sample arbitrarily complex examples without accidentally embedding the regularities of any particular human corpus. The complexity dial---nesting depth, production-rule branching factor, ambiguity---is ours to turn.
\item \textbf{Hard ground truth.} For every generated sample, the grammar yields the correct parse, the correct logical structure, and the correct answer. There is no annotator disagreement.
\item \textbf{Constrained action spaces.} When a neural agent emits outputs (e.g., a derivation, a parse, or a decision sequence), we can require those outputs to lie inside the grammar. Invalid productions are simply unrepresentable. This eliminates a large class of trivial errors and isolates genuine reasoning failures.
\item \textbf{Confounder-free evaluation.} Because the inputs are synthetic and procedurally diverse, benchmark memorization is impossible. What remains, when a model succeeds, is an estimate of its actual syntactic and structural reasoning ability.
\end{itemize}

In short: the grammar gives us a sandbox where every claim about the model's reasoning can, in principle, be checked against an absolute reference.

\vspace{6pt}\noindent\textbf{Layer 2: Faithfulness via Causal Perturbation and Simulatability}\vspace{2pt}

Having ground truth is necessary but not sufficient. A model that produces correct answers may still be producing wrong \textit{explanations}---reasoning traces that look principled but don't reflect the mechanism that actually generated the prediction.

To break the plausibility trap, we evaluate explanations using two complementary causal tests:

\begin{enumerate}[leftmargin=*, itemsep=2pt, parsep=0pt]
\item \textbf{Necessity and sufficiency testing.} For each feature, intermediate step, or rule application that an explanation cites as load-bearing, we perform targeted ablations and isolations. If the explanation says "step \textit{k} of the derivation determined the answer," then ablating step \textit{k} should produce a proportional drop in the model's confidence, and providing only step \textit{k} should approximately recover the prediction. Disproportionate or absent shifts are evidence that the explanation is not causally faithful.
\item \textbf{Simulatability.} We hand the generated explanation---and only the explanation---to an independent proxy model and ask it to reproduce the original system's decision. An explanation that genuinely captures the decision pathway should make the original output predictable from itself; one that is decorative will not.
\end{enumerate}

Crucially, both tests are \textit{quantitative}. Faithfulness becomes a measurable property rather than a vibe. And because the underlying tasks come from the grammar substrate, we know the true causal structure that explanations \textit{should} be tracking, giving these faithfulness scores a meaningful upper bound.

\vspace{6pt}\noindent\textbf{Layer 3: Self-Improvement Driven by Verifiable Synthetic Signal}\vspace{2pt}

The third layer turns the system into a recursive training flywheel. The model plays both roles in its own curriculum:

\begin{itemize}[leftmargin=*, itemsep=2pt, parsep=0pt]
\item \textbf{As a generator}, it produces candidate solutions, derivations, or reasoning trajectories on grammar-sampled problems.
\item \textbf{As an evaluator}, it filters those candidates using automated verification: self-consistency across multiple samples, execution against the grammar (does this parse? does this derivation reach the target?), or adversarial self-critique where one pass tries to falsify another.
\end{itemize}

Verified-correct trajectories become high-quality fine-tuning data. Persistent failure modes become hard-example sets for targeted training. Because verification is grounded in the grammar---not in another fallible model's opinion---the filtering signal is trustworthy.

Iterating this loop has a striking effect: the model's competence frontier shifts outward. Each round of fine-tuning enables it to solve slightly harder grammar-sampled problems, which in turn become the seed material for the next round. Reliance on human-labeled supervision drops sharply, and the model can move into regions of task complexity that no annotator was ever asked to cover.

\vspace{6pt}\noindent\textbf{Why the Three Layers Need Each Other}\vspace{2pt}

Each layer fails on its own; together, they form a closed loop.

\begin{itemize}[leftmargin=*, itemsep=2pt, parsep=0pt]
\item \textbf{Grammars without self-improvement} give us a clean evaluation but no path to scaling capability.
\item \textbf{Self-improvement without grammars} is dangerous: without absolute verification, the model's filter is itself biased, and the flywheel can amplify its own errors.
\item \textbf{Self-improvement without faithfulness checks} produces models that get better at the task but increasingly opaque, and we cannot distinguish genuine reasoning gains from sophisticated pattern matching.
\item \textbf{Faithfulness checks without grammars} lack a reliable referent: we can show that an explanation is internally causal, but not that it tracks the \textit{correct} causal structure.
\end{itemize}

When stitched together, the methodology works as follows. The grammar defines the problem space and supplies ground truth. The model bootstraps its capabilities through verified self-generated data, filtered against grammar-grounded checks. At each stage, its explanations are stress-tested via necessity/sufficiency ablation and simulatability, ensuring that capability gains are accompanied by genuinely faithful internal reasoning---not merely better-sounding rationalizations.

\vspace{6pt}\noindent\textbf{Takeaway for Practitioners}\vspace{2pt}

If you are building reasoning systems and want to escape the trio of memorization, plausible-but-empty explanations, and label scarcity, the recipe is: anchor your task distribution in a formal grammar; constrain model outputs to that grammar; use causal perturbation and simulatability to audit explanations; and close the loop with self-generated, grammar-verified training data. The grammar is what makes everything else honest.
\end{tcolorbox}

\begin{tcolorbox}[
  colback=gray!5,
  colframe=gray!50,
  boxrule=0.5pt,
  arc=2pt,
  left=8pt,
  right=8pt,
  top=6pt,
  bottom=6pt,
  title={\textbf{Alien Sampler -- Example Idea 2}},
  fonttitle=\small,
  breakable
]

\textbf{Atoms:}
\begin{itemize}[leftmargin=*, itemsep=2pt, parsep=0pt]
  \item Continuous relaxation transforms discrete sequence generation into a differentiable optimization problem by representing discrete variables---such as text tokens---as continuous probability distributions, soft logits, or dense embeddings. By applying gradient descent to these continuous proxies and systematically projecting the updated vectors back into discrete space using techniques like straight-through estimators, Voronoi partitioning, or gradient-guided coordinate substitution, the intractability of combinatorial discrete search is bypassed. This paradigm allows complex constraints, sequential dependencies, and targeted objective functions to be efficiently optimized across entire high-dimensional language sequences simultaneously.
  \item Continuous relaxations and gradient estimation techniques---such as the Gumbel-Softmax trick, Straight-Through Estimators, and Hard Concrete distributions---enable end-to-end gradient-based optimization of neural networks containing discrete, non-differentiable operations. By mathematically transforming hard categorical selections, binary pruning masks, and architectural routing decisions into smooth probability distributions or surrogate gradients, continuous approximations allow standard backpropagation algorithms to simultaneously optimize continuous model parameters and discrete structural or sequential choices.
  \item Translating artificial personas from superficial stylistic mimicry into stable psychological simulations requires decomposing human identity into multi-dimensional constructs that combine standardized psychometric traits, episodic memory anchors, and distinct cognitive styles. By enforcing these structured profiles through explicit behavioral constraints, in-character self-reflection loops, and specialized architectural routing, AI agents resist collapsing into a generic baseline, enabling both the quantifiable validation of virtual identities via clinical psychological inventories and the deployment of behaviorally consistent synthetic populations for complex social modeling.
\end{itemize}

\vspace{6pt}
\textbf{Research Idea:}

{\bfseries Optimizing Discrete Personas: A Gradient-Based Approach to Stable Psychological Simulation in Language Models}\vspace{4pt}

\vspace{6pt}\noindent\textbf{The Problem: Personas That Collapse}\vspace{2pt}

When we ask large language models to roleplay as specific individuals --- a clinically anxious teenager, a stoic engineer, a citizen with particular political leanings --- they tend to drift. After a few turns, surface mimicry erodes and the model regresses toward a generic, helpful baseline. This is not merely an aesthetic flaw. It is a fundamental obstacle for any application that needs \textit{behaviorally consistent synthetic populations}: social simulation, clinical training, agent-based modeling, or stress-testing policies on virtual cohorts.

The root cause is that "being a persona" is not a single objective. It is a \textbf{discrete, structured commitment}: a choice of psychometric trait levels, a particular set of episodic memories, a cognitive style that shapes how the agent reasons. Each of these decisions is categorical or combinatorial, and current LLMs handle them only through prompting --- a brittle, non-optimized channel.

This blog sketches a methodology that treats persona instantiation as a \textbf{differentiable optimization problem over discrete identity structure}, fusing three threads: continuous relaxation of discrete generation, gradient-based optimization of discrete architectural choices, and a multi-dimensional psychological decomposition of identity itself.

\vspace{6pt}\noindent\textbf{The Core Idea}\vspace{2pt}

Rather than hand-engineer prompts or fine-tune on imitation data, we represent every discrete component of a persona --- token choices in self-narrations, on/off memory activations, and routing decisions to specialized cognitive modules --- as a continuous proxy that can be optimized end-to-end. We then project these continuous proxies back into discrete persona behavior, and we validate the resulting agents against established clinical psychometric inventories.

The methodology has three layers, each addressing one idea atom.

\vspace{6pt}\noindent\textbf{Layer 1: Continuous Relaxation of Persona Utterances}\vspace{2pt}

A persona is ultimately observable through what it says. To optimize what it says against a target psychological profile, we need gradients through token selection. We use the standard toolkit:

\begin{itemize}[leftmargin=*, itemsep=2pt, parsep=0pt]
\item \textbf{Soft logits and dense embedding proxies} stand in for hard token IDs during the forward pass.
\item \textbf{Gradient descent on these continuous representations} allows us to push outputs toward objectives like trait-consistency, memory-faithfulness, or style adherence.
\item \textbf{Projection back to discrete tokens} uses straight-through estimators, Voronoi partitioning over the embedding space, or gradient-guided coordinate substitution to anchor the soft optimum to a real token sequence.
\end{itemize}

This converts the intractable combinatorial search over "what would \textit{this} person say next?" into smooth optimization across an entire response simultaneously. Sequential dependencies and global constraints (e.g., "the response must reflect high neuroticism \textit{and} reference the character's childhood event") become joint objectives rather than greedy decoding heuristics.

\vspace{6pt}\noindent\textbf{Layer 2: Gradient-Based Optimization of Discrete Architecture}\vspace{2pt}

A stable persona is more than its utterances; it has internal \textit{structure}. Some memories should be active in some contexts, others suppressed. Different cognitive styles should route through different sub-networks: an analytical persona might engage a chain-of-thought module, an impulsive one might bypass it.

These are inherently discrete decisions --- binary memory masks, categorical routing --- and they too need to be learned, not hardcoded. Here we apply the second idea atom:

\begin{itemize}[leftmargin=*, itemsep=2pt, parsep=0pt]
\item \textbf{Gumbel-Softmax} for categorical routing among cognitive-style modules.
\item \textbf{Hard Concrete distributions} for binary masks gating which episodic memories are accessible.
\item \textbf{Straight-Through Estimators} for any remaining hard selections.
\end{itemize}

Crucially, this lets standard backpropagation jointly optimize \textit{continuous} model parameters (the language model weights or adapter parameters) and \textit{discrete} structural choices (which memories, which modules, which traits dominate). The persona is not just a prompt --- it is an architectural configuration that is itself learned.

\vspace{6pt}\noindent\textbf{Layer 3: A Structured Psychological Profile as the Optimization Target}\vspace{2pt}

Layers 1 and 2 are useless without a well-posed target. The third idea atom supplies it: a persona must be decomposed into multi-dimensional constructs:

\begin{enumerate}[leftmargin=*, itemsep=2pt, parsep=0pt]
\item \textbf{Standardized psychometric traits} (e.g., Big Five-style continuous trait vectors) provide the quantitative skeleton.
\item \textbf{Episodic memory anchors} --- concrete biographical events --- provide the narrative substrate that prevents traits from being abstract numbers.
\item \textbf{Distinct cognitive styles} define \textit{how} the persona reasons, not just \textit{what} it believes.
\end{enumerate}

These three dimensions are then enforced through:

\begin{itemize}[leftmargin=*, itemsep=2pt, parsep=0pt]
\item \textbf{Explicit behavioral constraints} that enter the loss function (e.g., trait-prediction heads scoring generated text).
\item \textbf{In-character self-reflection loops}, in which the model critiques its own output through the persona's lens, providing additional gradient signal.
\item \textbf{Specialized architectural routing} (the discrete choices from Layer 2) that physically separates cognitive modes.
\end{itemize}

Because the target is structured and quantitative, we can validate --- not just vibe-check --- the resulting agents by administering \textbf{clinical psychological inventories} and comparing scores to the target profile.

\vspace{6pt}\noindent\textbf{Putting It Together}\vspace{2pt}

The training/inference loop looks like this:

\begin{enumerate}[leftmargin=*, itemsep=2pt, parsep=0pt]
\item \textbf{Specify} a target persona as (trait vector, memory set, cognitive-style assignment).
\item \textbf{Generate} candidate behavior using soft logits / embedding proxies (Layer 1).
\item \textbf{Score} the soft generation against the structured psychological target --- trait classifiers, memory-grounding checks, style adherence --- and against in-character self-reflection critiques (Layer 3).
\item \textbf{Backpropagate} through the soft generation and through Gumbel-Softmax / Hard Concrete gates that control memory access and module routing (Layer 2).
\item \textbf{Project} soft outputs back to discrete tokens, and harden routing/memory masks at deployment.
\item \textbf{Validate} the deployed agent by administering psychometric inventories; iterate if scores deviate from the target.
\end{enumerate}

\vspace{6pt}\noindent\textbf{Why It Works}\vspace{2pt}

Each layer plugs a hole the others cannot:

\begin{itemize}[leftmargin=*, itemsep=2pt, parsep=0pt]
\item \textbf{Without Layer 1}, we cannot get gradient signal into the actual utterances; we are stuck doing prompt engineering.
\item \textbf{Without Layer 2}, the persona has no internal structure that resists collapse; memories and reasoning modes degenerate into a soft average.
\item \textbf{Without Layer 3}, optimization has no principled target --- we mimic surface style rather than simulating identity.
\end{itemize}

Together, they convert persona construction from a prompting craft into a well-defined optimization problem with measurable outcomes. The same continuous-relaxation machinery that has been used to optimize adversarial text or prune networks is repurposed to sculpt stable psychological agents --- and the same psychometric instruments used on humans become the validation harness for synthetic ones.

The payoff is populations of LLM agents that don't drift, whose differences are quantifiable, and whose internal structure is the product of learning rather than improvisation.
\end{tcolorbox}

\begin{tcolorbox}[
  colback=gray!5,
  colframe=gray!50,
  boxrule=0.5pt,
  arc=2pt,
  left=8pt,
  right=8pt,
  top=6pt,
  bottom=6pt,
  title={\textbf{Claude Opus 4.7 -- Example Idea 1}},
  fonttitle=\small,
  breakable
]

\textbf{Atoms:}
\begin{itemize}[leftmargin=*, itemsep=2pt, parsep=0pt]
  \item The 'cross-modal alignment gap' in Vision-Language Models exposes a critical safety vulnerability where models heavily aligned to refuse harmful discrete text fail to generalize those guardrails to continuous visual embeddings. Attackers exploit this geometric mismatch by distributing malicious intent across modalities, embedding harmful text visually, or pairing individually benign images and prompts to create emergent harm; therefore, securing these systems requires enforcing safety boundaries directly on fused internal activations, ensuring that the integrated multi-modal context reliably maps to the model's latent refusal mechanisms.
  \item High-level semantic concepts and behavioral traits in neural networks are encoded as distinct linear directions within high-dimensional activation spaces, a phenomenon formalized as the Linear Representation Hypothesis. These specific 'concept vectors' can be mathematically isolated through contrastive activation analysis---calculating the vector difference between the internal states of paired inputs representing opposing traits (e.g., true versus false, or safe versus harmful) to cancel out shared syntactic and contextual noise. Once isolated, these vectors enable direct operational control, allowing researchers to mathematically probe a network's latent cognitive states or surgically steer its behavior during generation by adding or subtracting the concept vector, entirely bypassing the need for model retraining.
  \item Sparse Autoencoders (SAEs) resolve the problem of neural superposition---where individual neurons simultaneously encode multiple, overlapping concepts---by functioning as dictionary learners that project dense, polysemantic model activations into a higher-dimensional latent space constrained by sparsity. This mechanism decomposes complex internal representations into discrete, 'monosemantic' features, where each active dimension corresponds to a single, human-interpretable concept. By isolating these individual computational primitives, SAEs enable researchers to transparently audit a neural network's internal logic, map causal feature circuits, and perform precise, surgical interventions on specific behaviors without the need for opaque fine-tuning.
\end{itemize}

\vspace{6pt}
\textbf{Research Idea:}

{\bfseries Closing the Cross-Modal Safety Gap with Sparse, Interpretable Concept Steering}\vspace{4pt}

\vspace{6pt}\noindent\textbf{The Problem: Safety Doesn't Survive the Jump to Vision}\vspace{2pt}

Vision-Language Models (VLMs) inherit the safety alignment of their text backbones, but that alignment turns out to be surprisingly brittle once a vision encoder is bolted on. The root cause is what we might call the \textbf{cross-modal alignment gap}: refusal behaviors are heavily trained against \textit{discrete} harmful text tokens, but the visual pathway feeds the model \textit{continuous} embeddings that occupy a different region of the representation space. The guardrails simply don't generalize geometrically.

Attackers have learned to weaponize this mismatch. They render harmful instructions as images (bypassing text-side filters entirely), or distribute malicious intent \textit{across} modalities --- pairing an innocuous image with an innocuous prompt that, only when fused, produces emergent harm. Because the harmful signal lives in the \textit{combined} internal state rather than in either input alone, input-level moderation is structurally insufficient.

The conclusion is uncomfortable but clear: safety must be enforced on the \textbf{fused internal activations} themselves, where the integrated multi-modal context actually meets the model's latent refusal machinery. The question is how to do that surgically --- without retraining the model, and without breaking benign behavior.

\vspace{6pt}\noindent\textbf{The Approach: Read the Latent State, Then Steer It}\vspace{2pt}

The methodology we describe combines two complementary mechanistic-interpretability tools to inspect and intervene on a VLM's fused activations:

\begin{enumerate}[leftmargin=*, itemsep=2pt, parsep=0pt]
\item \textbf{Contrastive activation analysis} to identify directions in activation space that correspond to safety-relevant concepts (e.g., "harmful intent" vs. "benign request").
\item \textbf{Sparse autoencoders (SAEs)} to decompose dense, polysemantic activations into discrete, monosemantic features, so the steering directions we recover are interpretable and precisely targeted rather than entangled with unrelated behaviors.
\end{enumerate}

Together these give us a way to (a) \textit{probe} whether a fused multi-modal context has crossed an internal safety boundary, and (b) \textit{steer} the model back toward refusal when it has --- all without fine-tuning.

\vspace{6pt}\noindent\textbf{Step 1: Isolate concept vectors via contrastive activation}\vspace{2pt}

The Linear Representation Hypothesis tells us that high-level semantic traits --- truthfulness, sentiment, harmfulness, refusal --- tend to be encoded as \textbf{linear directions} in a model's activation space. We can recover such a direction by collecting paired inputs that differ only along the trait of interest (a harmful request vs. its benign counterpart, both delivered through the same modality configuration), running each through the VLM, and taking the \textbf{vector difference} of their fused internal states at a chosen layer.

The contrastive subtraction is doing real work: shared syntactic structure, formatting tokens, and context-specific noise cancel out, leaving a vector that points along the axis the model itself uses to distinguish the two conditions. Crucially, because we collect these activations \textit{after} fusion of vision and language tokens, the resulting concept vector lives in the same space where cross-modal attacks actually manifest.

Once isolated, a concept vector is operational. We can:

\begin{itemize}[leftmargin=*, itemsep=2pt, parsep=0pt]
\item \textbf{Probe} by projecting an unseen fused activation onto it, asking "how harmful does this combined context look from the inside?"
\item \textbf{Steer} by adding or subtracting the vector during generation to push the model toward or away from refusal --- entirely bypassing retraining.
\end{itemize}

\vspace{6pt}\noindent\textbf{Step 2: Disentangle with sparse autoencoders}\vspace{2pt}

Raw concept vectors derived from contrastive pairs work, but they ride on top of a representation in \textbf{superposition}: individual neurons (and the activation directions they span) encode many overlapping concepts at once. A "harmfulness" direction recovered naively can entangle with topic, tone, or modality artifacts, producing brittle interventions that misfire on benign inputs.

This is exactly the problem SAEs are designed to solve. An SAE trained on the model's fused-layer activations acts as a \textbf{dictionary learner}, projecting dense activations into a much higher-dimensional, sparsely-active latent space. The sparsity constraint forces each active dimension to specialize, yielding \textbf{monosemantic features} that correspond to single, human-interpretable concepts.

In our setting, this means we can:

\begin{itemize}[leftmargin=*, itemsep=2pt, parsep=0pt]
\item Decompose the fused activation of a harmful multi-modal input into its constituent SAE features.
\item Identify the small set of features that consistently activate on harmful content across modalities --- including cases where harm is conveyed visually or only emerges from the image-text combination.
\item Use those features (rather than raw activation directions) as the substrate for steering, giving us interventions that are precise and auditable.
\end{itemize}

\vspace{6pt}\noindent\textbf{Step 3: Enforce safety on fused activations}\vspace{2pt}

Putting the pieces together, the runtime defense looks like this:

\begin{enumerate}[leftmargin=*, itemsep=2pt, parsep=0pt]
\item The VLM processes a multi-modal input and produces fused internal activations at a target layer.
\item The SAE decomposes those activations into monosemantic features; the relevant safety-related features are inspected.
\item A contrastively-derived refusal direction --- sharpened by the SAE decomposition --- is used either as a \textbf{detector} (flagging inputs whose fused state crosses the harmfulness boundary) or as a \textbf{steering vector} (added to activations to restore refusal behavior in real time).
\end{enumerate}

Because all of this happens on the \textit{fused} internal state, it is invariant to which modality smuggled in the harmful content. Visually-rendered text, benign-looking image+prompt pairs that only become harmful when combined, and ordinary text jailbreaks all converge to the same internal signature, where the intervention can act.

\vspace{6pt}\noindent\textbf{Why It Works}\vspace{2pt}

Three properties make this synthesis robust:

\begin{itemize}[leftmargin=*, itemsep=2pt, parsep=0pt]
\item \textbf{It targets the right surface.} Cross-modal attacks succeed because input-level safety checks see each modality in isolation. Operating on fused activations meets the attack where it actually lives.
\item \textbf{It is mechanistic, not behavioral.} Rather than patching outputs after the fact, we identify the model's own internal representation of harm and act on it directly. The Linear Representation Hypothesis gives us a principled reason to expect such a direction to exist; contrastive analysis gives us a way to find it.
\item \textbf{It is surgical.} SAEs cut through superposition so that interventions affect the intended concept and little else. This is what makes activation steering practical as a safety mechanism rather than a blunt instrument: we can audit \textit{which} feature we're modulating and verify it corresponds to the behavior we care about.
\end{itemize}

The broader point is that the cross-modal safety gap is not solved by training harder on the input side. It is solved by recognizing that safety is ultimately a property of the model's internal state, and by using interpretability tools --- linear concept vectors and sparse, monosemantic features --- to make that internal state legible and controllable.
\end{tcolorbox}

\begin{tcolorbox}[
  colback=gray!5,
  colframe=gray!50,
  boxrule=0.5pt,
  arc=2pt,
  left=8pt,
  right=8pt,
  top=6pt,
  bottom=6pt,
  title={\textbf{Claude Opus 4.7 -- Example Idea 2}},
  fonttitle=\small,
  breakable
]

\textbf{Atoms:}
\begin{itemize}[leftmargin=*, itemsep=2pt, parsep=0pt]
  \item Sparse Autoencoders (SAEs) resolve the problem of neural superposition---where individual neurons simultaneously encode multiple, overlapping concepts---by functioning as dictionary learners that project dense, polysemantic model activations into a higher-dimensional latent space constrained by sparsity. This mechanism decomposes complex internal representations into discrete, 'monosemantic' features, where each active dimension corresponds to a single, human-interpretable concept. By isolating these individual computational primitives, SAEs enable researchers to transparently audit a neural network's internal logic, map causal feature circuits, and perform precise, surgical interventions on specific behaviors without the need for opaque fine-tuning.
  \item Activation patching---along with variants like path patching and interchange intervention---is a causal interpretability technique that isolates the functional components of a neural network by surgically transplanting internal activations from a source execution run into a counterfactual or corrupted target run; if this targeted substitution alters or restores the model's final prediction to match the source, it definitively identifies the patched layers, tokens, or neural circuits as the causally responsible pathways for that specific reasoning step or behavior.
  \item Process Reward Models enhance complex artificial intelligence reasoning by shifting evaluation from final-outcome validation to granular, step-by-step verification. By assigning localized success probabilities to individual intermediate actions within a logical chain, these models solve the credit assignment problem and eliminate outcome bias caused by flawed logic that coincidentally yields a correct answer. This continuous, step-level supervision acts as a computational critic that enables advanced search algorithms to dynamically prune erroneous branches, trigger targeted self-correction, and optimally scale inference-time compute to isolate the most robust logical trajectory.
\end{itemize}

\vspace{6pt}
\textbf{Research Idea:}

{\bfseries Opening the Black Box of Reasoning: A Mechanistic Approach to Auditing Step-Level Logic}\vspace{4pt}

\vspace{6pt}\noindent\textbf{The Problem: We Can Score Reasoning, But We Can't See It}\vspace{2pt}

Modern language models perform impressive multi-step reasoning, but our tools for understanding \textit{how} that reasoning works remain crude. We typically evaluate a chain of thought by checking whether the final answer is correct---a coarse signal that conflates lucky guesses with sound logic and gives us no insight into where reasoning actually happens inside the network. Worse, when a model produces a flawed intermediate step, outcome-based supervision can reinforce the bad logic so long as the answer happens to land right.

This creates two intertwined gaps:

\begin{enumerate}[leftmargin=*, itemsep=2pt, parsep=0pt]
\item \textbf{A behavioral gap.} We lack reliable, step-level signals for which intermediate moves in a reasoning chain are actually correct.
\item \textbf{A mechanistic gap.} Even when we know a step is wrong, we don't know which internal components of the model produced it, nor what concepts those components were representing.
\end{enumerate}

A methodology that closes both gaps would let us not only catch faulty reasoning, but \textit{localize it inside the model} and \textit{intervene on it} without retraining.

\vspace{6pt}\noindent\textbf{The Approach: Step-Level Verification Coupled to Mechanistic Localization}\vspace{2pt}

The methodology synthesizes three complementary tools, each addressing one stage of an end-to-end auditing pipeline for reasoning:

\begin{itemize}[leftmargin=*, itemsep=2pt, parsep=0pt]
\item \textbf{Process Reward Models (PRMs)} to detect \textit{where} in a reasoning chain something goes wrong.
\item \textbf{Sparse Autoencoders (SAEs)} to expose \textit{what concepts} the model is representing at that moment.
\item \textbf{Activation patching} to verify \textit{which components} are causally responsible for the step.
\end{itemize}

Used together, these form a loop: localize-in-trajectory $\to$ decompose-in-representation $\to$ confirm-with-causation.

\vspace{6pt}\noindent\textbf{Step 1: Use a Process Reward Model to Localize Failure in the Reasoning Trajectory}\vspace{2pt}

A PRM scores each intermediate step in a chain rather than only the final answer. By assigning a localized success probability to every action, it solves the credit-assignment problem and removes the outcome bias that plagues end-to-end evaluation. In our pipeline, the PRM plays two roles:

\begin{itemize}[leftmargin=*, itemsep=2pt, parsep=0pt]
\item \textbf{As a search critic at inference time}, it lets us prune branches that derail, trigger self-correction, and concentrate compute on the most robust logical trajectories.
\item \textbf{As a diagnostic flag}, it pinpoints the specific token positions and steps where the model's reasoning probability collapses---giving us a precise \textit{site of interest} to investigate mechanistically.
\end{itemize}

The PRM, in other words, converts the vague question "what went wrong?" into "what happened at step \textit{k} of the chain?" That handoff is what makes the rest of the pipeline tractable.

\vspace{6pt}\noindent\textbf{Step 2: Decompose the Activations at the Failure Site with Sparse Autoencoders}\vspace{2pt}

At the flagged step, the model's hidden state is dense and polysemantic: individual neurons simultaneously encode many overlapping concepts due to superposition. Reading these activations directly tells us little.

We project them through a Sparse Autoencoder trained to act as a dictionary learner---mapping dense activations into a much higher-dimensional, sparsity-constrained latent space. In that space, only a handful of features fire, and each active feature ideally corresponds to a single, human-interpretable concept (a "monosemantic" feature).

This decomposition turns the failure site into something readable. Instead of staring at a vector of opaque numbers, we see a short list of concepts the model was attending to when its reasoning faltered---candidate hypotheses for \textit{what the model was actually doing}.

\vspace{6pt}\noindent\textbf{Step 3: Confirm Causality with Activation Patching}\vspace{2pt}

Interpreted features are still only correlational. To establish that a specific concept, layer, or attention pathway \textit{causally} drove the faulty step, we use activation patching and its variants (path patching, interchange interventions).

The protocol:

\begin{enumerate}[leftmargin=*, itemsep=2pt, parsep=0pt]
\item Run the model on the original (failing) chain and a counterfactual chain that should produce the correct step.
\item Surgically transplant activations---from particular layers, tokens, or even SAE feature directions---from one run into the other.
\item Observe whether the model's prediction at that step flips accordingly.
\end{enumerate}

If patching the candidate component restores or breaks the behavior, we have causal evidence that this component carries the reasoning step. Combined with the SAE decomposition, this lets us say not just "circuit \textit{X} is responsible" but "circuit \textit{X}, encoding concept \textit{C}, is responsible."

\vspace{6pt}\noindent\textbf{Why the Three Components Fit Together}\vspace{2pt}

Each technique compensates for the weakness of the others:

\begin{itemize}[leftmargin=*, itemsep=2pt, parsep=0pt]
\item \textbf{PRMs without mechanistic tools} tell us a step is wrong but not why; they operate purely at the behavioral surface.
\item \textbf{SAEs without PRMs} produce mountains of features with no principled way to know which ones to look at; the PRM provides the targeting signal.
\item \textbf{SAEs without patching} yield interpretive hypotheses that may be epiphenomenal; patching disciplines interpretation with causal tests.
\item \textbf{Patching without SAEs} can identify a responsible layer or head but leaves the \textit{content} of its computation unreadable, since superposed activations resist semantic labeling.
\end{itemize}

The result is a pipeline whose outputs are far richer than any component alone:

\begin{itemize}[leftmargin=*, itemsep=2pt, parsep=0pt]
\item \textbf{Trajectory:} Question answered: At which step did reasoning fail? Tool: Process Reward Model.
\item \textbf{Representation:} Question answered: What concepts were active at that step? Tool: Sparse Autoencoder.
\item \textbf{Causation:} Question answered: Which components actually drove the step? Tool: Activation patching.
\end{itemize}

\vspace{6pt}\noindent\textbf{Implications: From Auditing to Surgical Intervention}\vspace{2pt}

Once a faulty reasoning step has been localized in the trajectory, decomposed into interpretable features, and causally pinned to specific components, intervention becomes targeted rather than blunt. Rather than fine-tuning the entire model and hoping the bad behavior is overwritten, a practitioner can suppress or amplify specific SAE features, or block specific patched pathways, at exactly the steps the PRM flags.

This methodology reframes interpretability from a post-hoc curiosity into an operational loop: \textit{detect bad reasoning step-by-step, explain it in human concepts, prove the explanation causally, and act on it surgically.} It treats the reasoning chain as the natural unit of analysis---long enough to contain real logic, short enough that mechanistic tools can reach inside each link.
\end{tcolorbox}

\end{document}